\documentclass{article}

\PassOptionsToPackage{numbers, compress}{natbib}

\usepackage[preprint]{neurips_2026}

\usepackage[utf8]{inputenc} 
\usepackage[T1]{fontenc}    
\usepackage{booktabs}       
\usepackage{array}          
\usepackage{amsfonts}       
\usepackage{nicefrac}       
\usepackage{microtype}      
\usepackage[table]{xcolor}  
\usepackage{amsmath}
\usepackage{amssymb}
\usepackage[T1]{fontenc}
\usepackage[all=normal, floats, bibnotes, wordspacing, charwidths, lists, moderate]{savetrees}
\usepackage{graphicx}
\usepackage{float}
\usepackage[hyphens]{url}
\usepackage[hidelinks]{hyperref}
\hypersetup{breaklinks=true}
\usepackage[capitalize,noabbrev]{cleveref}
\usepackage{lineno}
\usepackage{paralist}

\definecolor{darkblue}{rgb}{0, 0, 0.5}
\hypersetup{colorlinks=true, citecolor=darkblue, linkcolor=darkblue, urlcolor=darkblue}

\title{Emergent Language as an Approach to Conscious AI}

\author{%
  Zengqing Wu \\
  University of Osaka \\
  \texttt{zengqing.wu@ist.osaka-u.ac.jp} \\
  \And
  Chuan Xiao \\
  University of Osaka \\
  \texttt{chuanx@nagoya-u.jp} \\
}

\begin{document}

\maketitle

\begin{abstract}
The question of whether artificial systems can be conscious remains open, in part because existing approaches either evaluate systems against theory-derived checklists (\emph{discriminative}) or engineer consciousness-inspired modules directly (\emph{architectural}); both leave open whether observed structures are artifacts of human language priors.
We propose a \emph{generative} methodology: emergent language (EL) in multi-agent reinforcement learning, where agents start from minimal (no language, no concept of self, minimal exposure to human text) and develop communication under task pressure alone, ensuring causal attributability to task demands rather than inherited human language priors.
We position our methodology by discussing how EL serves as a generative tool for studying consciousness-relevant structure, including the role of environment complexity and the interpretation of emergent communication.
As a proof of concept, we instantiate this methodology in a minimal environment and show that agents develop \emph{self-referential communication}, including an echo-mismatch detection circuit that is not predicted by task structure or architecture alone but emerges from a specific environmental affordance.
\end{abstract}

{
\renewcommand{\thefootnote}{}%
\footnotetext{Our source code is available at \url{https://github.com/wuzengqing001225/ConsciousAI_Indexicality/}.}%
}
\setcounter{footnote}{0}

\section{Introduction}
\label{sec:intro}

Whether artificial systems can be conscious is among the most contested questions in AI \citep{butlin2025consciousness}, with immediate ethical and regulatory implications.
Proponents point to increasingly sophisticated behavior \citep{bostrom2025propositions,dennett2023counterfeit,hofstadter2007strange}, while skeptics counter that behavioral sophistication is neither necessary nor sufficient for consciousness \citep{chalmers1996conscious}.
Although this debate~\footnote{Key opinion leaders' positions are available in Appendix~\ref{app:ai-debate}.} has persisted for decades without resolution, exploratory efforts have been made to study conscious AI \citep{butlin2025consciousness}.

Central to this topic lies the \emph{hard problem} of consciousness \citep{chalmers1995facing}: the explanatory gap between physical function and subjective experience, between objective behavior and the felt quality of qualia~\footnote{Qualia refer to instances of subjective experience, e.g., the ``blueness'' of blue.}.
\emph{Philosophical zombies} (or simply \emph{zombies}), the central figure in this debate, are beings functionally identical to humans yet lacking any conscious experience. If zombies are conceivable, the argument goes, then consciousness cannot be wholly explained by function alone.
These concerns highlight the depth of the challenge: understanding not just what artificial systems can do, but whether their information processing could give rise to the kind of first-person experience that characterizes consciousness.

In this paper, we propose a methodology for studying conscious AI.
Our starting point is Moody's zombie civilization argument \citep{moody1994conversations}: in a world of zombies, the concepts of conscious experience would never originate, because there would be nothing for such concepts to refer to.
We draw methodological inspiration from this argument by designing environments for AI agents and speculating whether they can independently develop a word that means ``consciousness'' in their language. 
Meanwhile, we remain agnostic about subjective experience, thereby differing from Moody's viewpoint in the sense that Moody is concerned with the origin of consciousness \emph{concepts}, whereas we focus on the emergence of consciousness-relevant \emph{structures}.

Our approach is to construct minimal virtual worlds populated by agents with no language, no concept of self, and minimal exposure to human text~\footnote{We say minimal rather than no exposure because the design of environment, agent architecture, and reward is based on human knowledge and thus inevitably affected by human language prior. Nevertheless, we aim to reduce such impact to the minimal level.}, and to observe what consciousness-relevant structures emerge from task pressure alone.
Any observed structure is necessarily emergent rather than inherited from human language priors, as opposed to language models, in particular large language models (LLMs), pre-trained using human language data. 
We propose that progress requires not resolving whether AI ``has'' consciousness, but developing methodologies that can track the structural conditions under which functional preconditions for conscious experience arise. In this sense, our methodology belongs to the category of \emph{functionalism}.

\paragraph{Two methodological principles.}
Our approach rests on two commitments. 
(1) \textbf{Environment shapes behavior}: 
Just as ecological demands drive the evolution of cognitive capacities in biological organisms, we hypothesize that sufficiently structured environments can drive the emergence of functional structures relevant to consciousness in artificial agents, without those capacities being designed in.
(2) \textbf{Phenomenological epoch\'{e}}\footnote{Phenomenological epoch\'{e} (a.k.a. bracketing), as a preliminary step of phenomenology, means refraining from judgment about the reality and focusing on how things appear to us.}: a deliberate suspension of judgment on whether AI systems could have subjective experience.
We bracket the question of subjective experience and focus on what can be observed: the linguistic practices that agents develop under task pressure, and the environmental structures those practices ground.
As such, we focus on a tractable phenomenological question: what structures in the agent's behavior and representations are functionally equivalent to the preconditions identified by consciousness theories?
The epoch\'{e} ensures that any emergent structures we observe are products of task pressure and environmental demands, not artifacts of consciousness-theoretic assumptions built into the system design.

\begin{figure}[t]
  \centering
  \includegraphics[width=0.8\columnwidth]{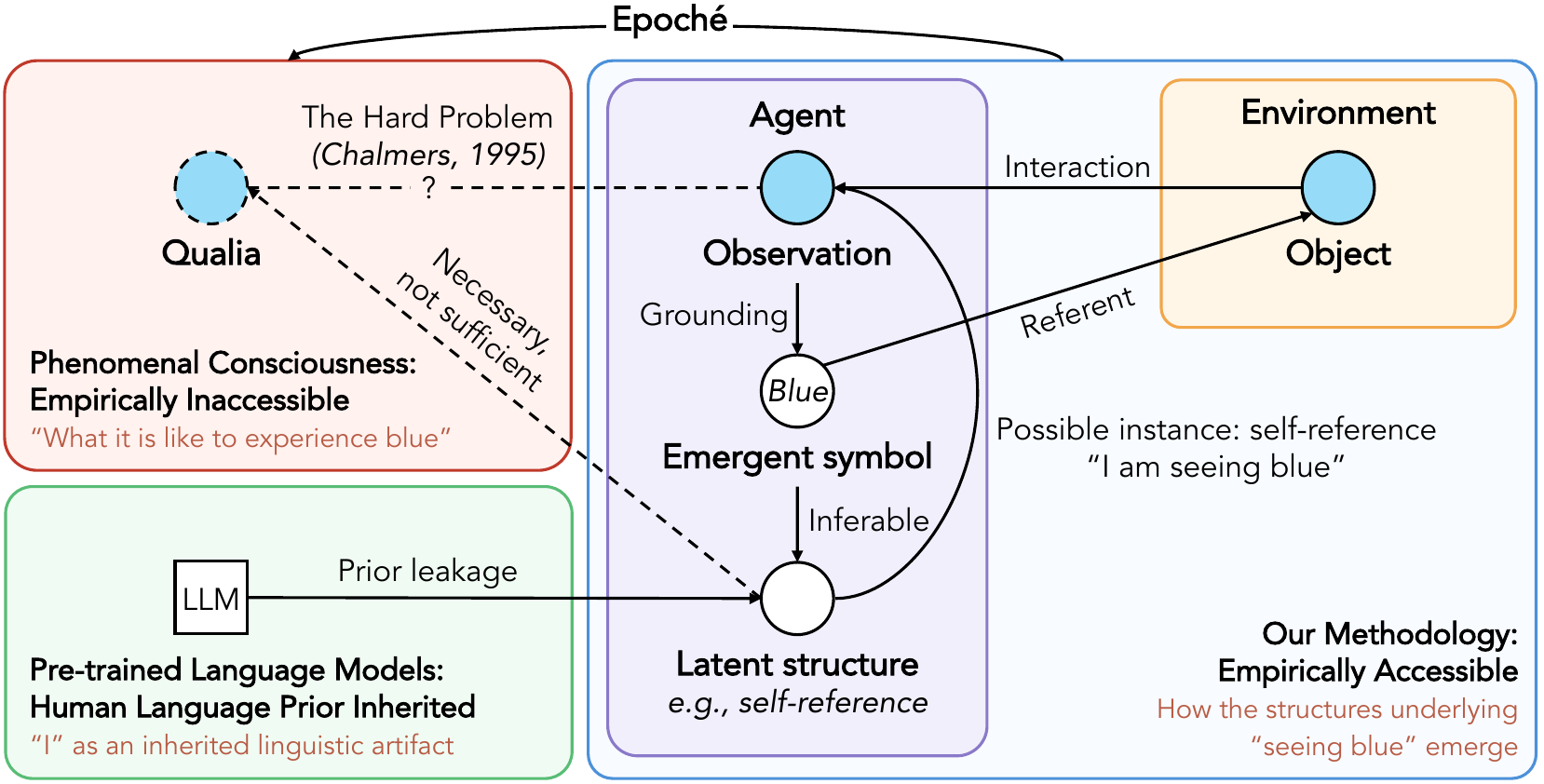}
  \caption{EL as a methodological approach to conscious AI. (1) Using ``blue'' as an example, the figure separates the private phenomenal question (what it is like to experience blue) from the empirically accessible structures surrounding ``seeing blue.'' An agent may observe blue, ground a symbol for blue through interaction with blue objects, and develop emergent communication that makes latent internal structure inferable. A possible instance of such latent structure is self-reference, as in ``I am seeing blue.'' This does not show that blue has any subjective feel for the agent; the qualitative character of blueness remains bracketed by epoch\'{e}. (2) The figure also contrasts the EL route with the LLM route, where apparent self-reference may reflect inherited human language priors. (3) The methodological claim is functional: EL provides a grounded, empirically accessible window onto consciousness-relevant latent structures, not direct evidence of qualia.}
  \label{fig:concept}
\end{figure}

\paragraph{Emergent language as a generative approach.}
To investigate these structural preconditions empirically, we turn to emergent language (EL) in multi-agent systems.
In EL research, agents with minimal prior language learn to communicate through discrete tokens under task pressure, developing communication protocols from scratch \citep{foerster2016learning,lazaridou2017multi}.
We use EL not to improve agent performance (the standard goal of the field) but as a \emph{generative} approach to studying the origins of consciousness-relevant structures.
By placing agents under coordination pressure with bandwidth-constrained channels, we observe what structures emerge when agents need to transmit information about themselves (Figure~\ref{fig:concept}).

The EL approach addresses a methodological gap left by two existing paradigms.
\emph{Discriminative} methods take a checklist of indicators derived from consciousness theories and ask whether a given system satisfies them \citep{butlin2025consciousness}: productive but retrospective, since the system is designed for other purposes and the evaluation can only confirm or deny pre-specified criteria, never reveal structures the theory did not anticipate.
\emph{Architectural} methods build consciousness-inspired modules directly into a system \citep{dehaene2003neuronal,wilterson2021attention}: informative but circular, since any resulting consciousness-relevant behavior may reflect the designer's assumptions rather than structural necessity.
Our \emph{generative} approach asks neither ``is it conscious?'' nor ``can we make it conscious?'' but ``what consciousness-relevant structures arise when none are designed in?''

\paragraph{Methodology in brief.}
Our methodology operates through three interconnected elements.
(1) \emph{Environment complexity as a driver}: following the principle that general methods leveraging computation outperform domain-specific knowledge engineering \citep{sutton2019bitter}, we scale environment complexity~\footnote{This element serves as an envisioned research agenda in this paper, while the other two elements are evaluated empirically.} rather than encoding target capacities directly, letting task pressure drive emergence.
(2) \emph{Prior-minimal design}: agents learn from scratch with minimal exposure to human language, ensuring that any observed structure is causally attributable to current task demands rather than inherited priors.
(3) \emph{Interpretation through intervention}: emergent protocols are analyzed via ablation, probing, and information-theoretic decomposition; at larger scales, LLM-based in-context learning and alignment with human language provide complementary interpretation pathways.

\paragraph{Indexicality as the first step.}
To employ our methodology, agents must express their internal states. 
Thus, the first question is whether agents can develop \emph{self-referential communication}: messages whose content is about the sender itself.
Perry's essential indexical \citep{perry1979essential} and Kaplan's character/content distinction \citep{kaplan1989demonstratives} established the \emph{demand side}: indexical structure is philosophically indispensable for self-referential communication.
We investigate the \emph{supply side}: whether indexical structure also arises under bandwidth-constrained communication.

As a proof of concept, we train reinforcement learning (RL) agents that communicate via discrete tokens to coordinate on a cooperative task.
Each agent observes only its own private state, but the task requires knowledge of the partner's state, making communication necessary.
The channel is narrow, and agents begin with no language and no concept of self (i.e., no architectural module representing the agent itself, no first-person token in the vocabulary, and no auxiliary self-supervised objective targeting the agent's own state).
We identify three structural properties:
\begin{itemize}
  \item \textbf{(P1) Indexical encoding:} messages carry primarily the sender's own state (information-theoretically predicted by task structure);
  \item \textbf{(P2) Persistent state representation:} a recurrent latch maintaining self-state across time (architecturally expected under partial observability);
  \item \textbf{(P3) Behavioral self-monitoring:} agents develop a closed-loop circuit that detects mismatches between what they intended to transmit and what is echoed back. 
\end{itemize}
P1 and P2 are predicted foundations that validate our methodology's observation conditions.
P3 is the core empirical finding: it is not predicted by task structure or architecture alone, requires a specific environmental affordance (the echo channel), and demonstrates that our methodology can detect emergent functional structure beyond what task pressure and agent architecture directly predict.

Because the echo channel is supplied by the environment, a natural objection is that P3 only shows that agents can learn to use a designed input line.
We address this with four additional tests that separate what the environment supplies (the echo channel) from what agents learn (a reference for the intended message, an attribution of feedback to the agent's own message, and a policy for responding): 
(1) a yoked control in which the feedback line carries the same agent's token from a different episode, so that its statistics are preserved but its relation to the current episode is broken;
(2) a randomized broadcast condition in which both agents' tokens arrive in randomly ordered, unlabeled input slots, so that no slot is marked as ``self'';
(3) a measurement of how much the message that an agent emits after a mismatch helps its partner; and
(4) a state-change variant in which the sender's state can change after a message about the old state has been sent.

\paragraph{Contributions.}
(C1)~A generative methodology of EL with the prior-minimal design for studying the origins of consciousness-relevant structures in AI, complementing discriminative and architectural approaches.
(C2)~A formal operationalization of indexical reference connecting Kaplan's character/content distinction to mutual-information criteria testable in emergent communication systems.
(C3)~A proof-of-concept experiment in which P1 (indexical encoding) and P2 (persistent state representation) serve as observation conditions and P3 (behavioral self-monitoring) serves as the core finding, together with controls (a yoked control, randomized broadcast, a repair-value counterfactual, and a state-change variant) that identify which parts of P3 are supplied by the environment and which are learned.

\paragraph{Outline.}
\S\ref{sec:background} reviews related work. 
\S\ref{sec:methodology} develops our methodological position on how EL serves as a generative tool for studying conscious AI. 
\S\ref{sec:framework} formalizes indexicality and structural preconditions.
\S\ref{sec:experiments}--\ref{sec:results} present experiments and results.
\S\ref{sec:discussion} discusses implications of the experimental findings. 
\S\ref{sec:conclusion} concludes. 

\section{Related Work}
\label{sec:background}

We review two bodies of work: consciousness theories that presuppose the first-person structure we aim to explain (\S\ref{subsec:consciousness}) and existing research on EL (\S\ref{subsec:el-primer}). 
More discussion on related work is available in Appendices \S\ref{app:consciousness-theories}--\ref{app:extended-related-work}. 

\subsection{Consciousness Theories and the First-Person Presupposition}
\label{subsec:consciousness}

The dominant theories differ in mechanism but share a presupposition: each requires a first-person perspective without explaining its origin.
Global Neuronal Workspace Theory (GNWT) needs someone for whom information is broadcast across prefrontal-parietal networks \citep{dehaene2001towards};
Integrated Information Theory (IIT) needs someone experiencing the integration quantified by $\Phi$ \citep{tononi2016integrated};
Higher-Order Thought theory (HOT) needs someone whose mental state is the object of a higher-order representation \citep{rosenthal2002how};
Attention Schema Theory (AST) needs someone whose attention is being schematically modeled \citep{graziano2013consciousness}.
The pattern extends to more recent proposals: the Consciousness Prior \citep{bengio2017consciousness} selects a low-dimensional summary via attention, but attention directed by whom?
Predictive self-models \citep{seth2021being,metzinger2003being} ground experience in the brain's self-modeling capacity, yet the ``self'' doing the modeling is presupposed rather than derived.
\citet{hofstadter2007strange} makes the recursion explicit, casting self-reference as a strange loop, but this redescribes the structure without explaining how it originates.
What binds these frameworks is not a shared mechanism but a shared gap: each presupposes a subject-position and then explains what that subject does, without asking where the subject-position itself comes from.

A practical difficulty compounds the theoretical one: several of these theories lack operational counterparts in AI systems.
GNWT posits a neuronal workspace with specific prefrontal-parietal dynamics; IIT requires computing $\Phi$ over a system's causal structure, a calculation that is intractable for large networks and whose applicability to non-biological substrates is contested.
Neither framework provides a clear procedure for evaluating consciousness in architectures that differ fundamentally from the mammalian brain.
This operational gap is one reason the debate remains unresolved.
Empirical and conceptual challenges further compound the problem.
The Cogitate Consortium's adversarial test of IIT and GNWT found partial support for each but confirmation of neither \citep{cogitate2025adversarial}.
\citet{chalmers2023could} argues that no current evidence suffices to establish or deny LLM consciousness.
\citet{mcclelland2025agnosticism} states that we hit an ``epistemic wall'' when extrapolating AI systems because what we know about consciousness originates from human organisms. 
\citet{shevlin2021nonhuman} identifies a ``specificity problem'' for non-human systems.
\citet{hoel2026disproof} poses a dilemma between falsifiability and triviality.

In this paper, we sidestep the question these debates address (i.e., whether a system is conscious) and instead ask where the structural preconditions for self-referential communication originate.
A related effort is made by \citet{immertreu2025probing}, who probe RL agents trained in virtual environments for world models and self-models using Damasio's theory of core consciousness, finding that rudimentary forms of both emerge as byproducts of task training.
Our work shares the spirit of probing for emergent structure, but differs in method (communication-based rather than probe-only) and scope (emergence of self-referential communication rather than self-models in isolation).
Closest in spirit to our methodological stance is the \emph{constructive approach} of \citet{taniguchi2024constructive}, who propose to study the bidirectional influence between language emergence and qualia structure by building systems in which language emerges, rather than by evaluating existing systems.
Their framework, grounded in the Collective Predictive Coding hypothesis \citep{taniguchi2024collective}, treats language emergence not only as the creation of a communication tool but as a mechanism that aligns internal representations across agents, and it explicitly connects this to consciousness science.
We share the constructive stance and the focus on emergence, but ask a different question: whereas the constructive approach asks how EL shapes the structure of internal representations, we ask whether self-referential structure itself emerges, and we use prior-minimal agents so that the answer can be causally attributed to task demands.
\subsection{Emergent Language}
\label{subsec:el-primer}

EL research studies what happens when agents with minimal prior language must learn to communicate in order to coordinate.
In a typical setup, multiple agents are placed in a cooperative environment where task success requires information that only other agents possess.
Agents communicate through a discrete channel with limited bandwidth (e.g., a small vocabulary of tokens, one per time step) and develop their communication protocol entirely through RL, with no supervision on message content \citep{lewis1969convention,lazaridou2017multi,foerster2016learning}.
The field has produced a rich body of results on compositionality, grounding, and repair under noise~\cite{chen2025fivews, peters2025emergent}.

Existing research has predominantly treated EL as a communication protocol for improving multi-agent system performance on downstream tasks: coordination, navigation, referential games.
A systematic gap runs through this literature: research has focused on \emph{what is said} (whether agents develop compositional, grounded references to external states) but not on \emph{who is saying it}.
Whether agents develop signals that encode the speaker's own identity or refer to themselves has barely received attention.
\citet{chen2025fivews}'s ``who'' refers to communication topology, not to whether agents develop tokens that function as first-person reference.
This gap is notable in light of \citet{locke2021indexical}, who argues that primate communication is primarily indexical: close calls, grooming, and other signals convey personal states and traits rather than symbolic environmental references, and that this personal information exchange is fundamental to group cohesion.
If indexical encoding is the biological baseline, its absence from the EL literature is a significant omission.
This is the gap we address: we use EL not as a means to improve task performance, but as a window into the emergence of consciousness-relevant structures~\footnote{Discussion on why communication, rather than internal probing alone, is the appropriate lens is available in Appendix~\ref{app:why-communication}.}.

\section{Methodology}
\label{sec:methodology}
\subsection{Environment Complexity as a Driver}
\label{subsec:roadmap}

In the Bitter Lesson \citep{sutton2019bitter}, Sutton argues that the most effective approach in AI is to leverage \emph{computation and learning} rather than encoding human knowledge.
\citet{silver2025era} extend this to the ``era of experience'': agents that learn from interaction with the environment, rather than from human-generated data, will break through the limitations of human-centric AI; actions and observations grounded in the environment, rather than mediated through human text, provide the foundation for discovering strategies that might never occur to a human.
We adopt an analogous principle for the study of conscious AI: rather than designing consciousness-relevant (e.g., self-referential) capacities into agents, we scale environment complexity and let task pressure drive their emergence.

The comparative cognition literature provides a biological analogue.
Self-referential capacities in nature form a continuum driven by ecological demand: mirror self-recognition in chimpanzees \citep{gallup1970chimpanzees}, uncertainty monitoring in rhesus monkeys \citep{hampton2001rhesus,smith2003comparative}, theory of mind in social primates \citep{premack1978chimpanzee}, and narrative self-continuity in humans \citep{dennett1991consciousness}.
Each capacity appears when, and plausibly because, the organism's environment creates selection pressure that rewards it.

\paragraph{Scaling dimensions.}
We hypothesize that the same logic applies to AI agents: progressively more complex environments should drive progressively richer consciousness-relevant structures.
In \S\ref{sec:experiments}--\ref{sec:results}, we will show that our minimal environment (two agents, seven tokens, ten-step episodes) suffices to drive the three structural preconditions P1--P3.
Several dimensions of scaling are available: environment complexity (from grid worlds to physically grounded simulations to real-world interaction), agent population size and heterogeneity, communication bandwidth and modality (from discrete tokens to continuous signals to physical vocalization), temporal horizon (from fixed episodes to open-ended streams), and social structure (from fixed partnerships to dynamic alliances and adversarial encounters).
Each dimension introduces new pressures that could drive richer emergent structures.

Following the Bitter Lesson, 
we scale environment complexity and observe what emerges, without encoding human knowledge about consciousness (e.g., self-recognition, self-awareness, identity) as design targets.
Nonetheless, indexicality, which refers to the emergence of tokens that refer to the speaker's own states, is the first observation condition: without evidence that agents encode their own states, we cannot study further consciousness-relevant structures. \S\ref{sec:framework} will elaborate on this.

\subsection{Scaling with Minimal Human Language Prior}
\label{subsec:scaling}

In our methodology, a key criterion is \emph{causal attribution}: for any observed structure, ablation or intervention must verify that it causally originates from the environment.
This criterion becomes increasingly important as environments grow in complexity, because richer environments, along with more sophisticated agent architecture, also create more opportunities for subtle prior leakage.
This rules out LLM agent approaches, which risk importing the first-person pronoun ``I'' as a statistical regularity rather than an emergent structure~\footnote{Detailed discussion is available in Appendix~\ref{app:prior-leakage}.}.

\subsection{Interpreting EL}
\label{subsec:interpretation}

Besides the EL itself, its interpretation is also a key step in our methodology, because EL is not designed for human readability.
We outline several interpretation strategies at different scales, noting that these are tentative methodological proposals rather than established standards. The field may develop better approaches as it matures. 
The interpretation challenge scales with environment complexity, and different regimes call for different methods.

\paragraph{Small scale: ablation and probing.}
In minimal environments, the standard toolkit of mechanistic interpretability suffices: mutual-information analysis, linear probing of hidden states, ablation of channels and architectural components, and intervention experiments (e.g., corruption, silencing).
This is the approach we adopt in the experiments of this paper.
Its strength is causal clarity. Its limitation is that it does not scale to environments where the EL is too complex for manual analysis.

\paragraph{Medium scale: LLM in-context learning.}
At intermediate complexity, LLMs can serve as interpreters of EL via in-context learning: given examples of agent behavior paired with environment states, an LLM can be prompted to describe the structure of the EL in human-readable terms.
This leverages the LLM's capacity for pattern recognition without requiring the EL to have been part of the LLM's training data.
Using an LLM as interpreter does not reintroduce the prior leakage that motivates our prior-minimal design~\footnote{We discuss this in Appendix~\ref{app:llm-environment}.}.

\paragraph{Large scale: pre-training and alignment.}
At high complexity, a dedicated language model could be pre-trained on the agents' communication stream (i.e., emergent communication as a corpus) and then aligned with human language (e.g., by mixing with human text corpus for pre-training), creating a translation layer between emergent and natural language.


\subsection{Continual Learning and Transfer}
\label{subsec:continual}

The richer capacities such as metacognitive monitoring, narrative self-continuity, and theory of mind plausibly require agents that continue to learn during deployment, not just during a fixed training phase.
\citet{valiente2024muse} demonstrate one such direction: their MUSE framework integrates metacognitive processes (self-assessment and self-regulation) into autonomous agents, improving out-of-distribution performance.
Our P3 provides a minimal precursor to such metacognitive capacity; scaling to richer self-monitoring would require the kind of continual adaptation MUSE exemplifies.
Biological organisms develop conscious capacities over developmental timescales through ongoing interaction with an environment whose complexity itself changes.
Frozen-weight agents, as used in our experiments, can demonstrate that structural preconditions can emerge, but cannot demonstrate the open-ended developmental trajectories that richer consciousness-relevant capacities would require.

A natural scaling path is \emph{sim-to-sim transfer} (cf. sim-to-real transfer~\cite{zhao2020sim}): train agents in a simple environment until basic communication and consciousness-relevant structures stabilize, then transfer them to a more complex environment where the existing structures provide a scaffold for more elaborate capacities.
This mirrors developmental trajectories in biological organisms, where early sensorimotor competence scaffolds later cognitive achievements.
The key constraint is that the transfer must import minimal human language priors. 

\subsection{Limitations of the EL Methodology}
\label{subsec:method-limits}

The generative approach via EL faces several obstacles that should be acknowledged as part of the methodology itself, not deferred to post-hoc limitations.

\paragraph{Computational cost.}
Prior-minimal agents must discover from scratch what evolution has spent billions of years optimizing.
Even our minimal environment requires tens of thousands of training updates per condition; scaling to more complex environments requires substantially greater computation.

\paragraph{Human language priors not leveraged.}
The prior-minimal design, while methodologically essential, means that the vast knowledge encoded in human language, including knowledge about self-reference, mental states, and social cognition, is deliberately excluded.
An important open question is whether the structures that emerge via EL and those that emerge via next-token prediction on human text (e.g., LLMs) can be shown to converge to equivalent models. 

\paragraph{No established paradigms for continual learning.}
Reaching richer capacities likely requires agents that learn continually and maintain long-term memory across changing environments.
Current RL frameworks are predominantly designed for fixed-horizon episodic training; paradigms for open-ended continual learning in multi-agent communication settings remain an open research problem \citep{pan2025survey}.

\paragraph{The mesoscale gap of interpretation.}
Between the small and large regimes identified in \S\ref{subsec:interpretation} lies a potential gap: systems complex enough that ablation becomes impractical, yet not complex enough for learning methods to extract reliable structure.
This mesoscale problem, analogous to the challenges of studying systems with intermediate degrees of freedom in physics, is an open methodological challenge for the field.


\section{Indexicality}
\label{sec:framework}


\subsection{Why Indexicality Is Necessary}
\label{subsec:indexicality}

Our methodology studies consciousness-relevant structures by observing what agents communicate.
For this to work, agents' messages must carry information about their own internal states; otherwise, communication provides no empirical window into internal representations, and the methodology has no signal to analyze.
Indexicality is thus a necessary observation condition: it is what makes agents' internal structure accessible through their communicative output~\footnote{The philosophical foundations \citep{perry1979essential,kaplan1989demonstratives} that motivate the formal definition of indexicality are developed in Appendix~\ref{app:indexicality-philosophy}.}.

Concretely, if agents develop a protocol in which $I(m; s_{\text{self}}) \approx 0$, where $I(\cdot;\cdot)$ denotes mutual information, messages are uninformative about the sender, and probing, ablation, and information-theoretic decomposition all return null results.
The entire evidence chain for persistent state representation (P2) and behavioral self-monitoring (P3) depends on indexical encoding (P1) holding first: P2 is detected by probing hidden states that encode $s_{\text{self}}$, and P3 is detected through changes in communicative output that already carries sender state.
Without indexical encoding, neither is empirically accessible.

\subsection{Formal Definition of Indexical Reference}
\label{subsec:formal-def}

\paragraph{Setting.}
Consider $N$ agents ($N = 2$ in our experiments), each assigned a private state $s_i$ drawn independently from a finite set $\mathcal{S}$.
Each agent additionally observes a context $c_i \in \mathcal{C}$, drawn independently per agent and per episode. $c_i$ is independent of the partner's state $s_j$, and both enter the reward formula together: the acting agent must combine its own $c_i$ with the communicated $s_j$ to select the correct action (an explicit form $a^*_{i \to j} = (s_j + c_i + t)\bmod 3$ is given in \S\ref{subsec:environment}, where $t$ is the time step).

At each time step $t$, agent $i$ selects a message $m_i^{(t)} \in \mathcal{M}$ from a discrete vocabulary of size $|\mathcal{M}|$ and an action $a_i^{(t)}$.
Each agent receives the previous-step messages from the other agents. In our two-agent experiments, agent $i$ receives the single partner message $m_{-i}^{(t-1)}$. The task requires each agent to take actions that depend on other agents' private states, making communication necessary for coordination.
The communication channel is narrow at the per-step level: $|\mathcal{M}| < |\mathcal{S}|^N$ compares the per-step message capacity to the joint private-state space, so a single message cannot encode the full configuration of all agents' private states. We deliberately measure narrowness in this single-step sense rather than including $|\mathcal{C}|$ or summing over $T$ steps, because each agent already observes $c_i$ and $t$ locally and need not transmit them. The bandwidth pressure is over what must cross the channel, not over what is privately available. Agents must thus develop efficient encoding strategies for the information that is communicable in principle.
Throughout, sender identity is treated as given by channel position (each receiver knows which channel/slot a message arrived on). Under this simplifying assumption, the receiver does not need to recover the sender from the message itself~\footnote{The conditions under which compositional ``who/what'' encoding becomes pressured are discussed in Appendix~\ref{app:compositional}.}.

This setting inherits the cooperative multi-agent RL (MARL) structure of \citet{foerster2016learning} and \citet{mordatch2018emergence}, but with a critical difference: the referents that matter for task success are agents' private internal states, not shared external observations.
Whereas referential games \citep{lazaridou2017multi,havrylov2017emergence} require agents to communicate about objects visible to at least one party, our task requires agents to communicate about themselves.

\paragraph{Definition.}
We define a token $m \in \mathcal{M}$ as an \emph{indexical token} w.r.t. sender $i$ if it satisfies three criteria:

\begin{enumerate}
  \item \textbf{MI dominance.} The mutual information between the token and the sender's own state dominates:
  \begin{equation}
    I(m_i; s_i) \gg I(m_i; s_j) \quad \text{for all } j \neq i.
    \label{eq:mi-dominance}
  \end{equation}
  That is, the token's information content is primarily about the sender, not about other agents.

  \item \textbf{Context independence.} The encoding rule is invariant across contexts:
  \begin{equation}
    I(m_i; c_i \mid s_i) \approx 0.
    \label{eq:c-independence}
  \end{equation}
    Conditioned on the sender's state, knowing the sender's context provides
    negligible additional information about the token. This invariance
    ensures that the same encoding rule applies regardless of
    context, supporting generalization to novel context values; 
    without it, the encoding could be 
    context-specific lookup rather than a stable mapping from self-state 
    to message.

  \item \textbf{Communication necessity.} Removing the message channel causes a significant drop in task performance:
  \begin{equation}
    \Delta_{\text{comm}} = R_{\text{with-msg}} - R_{\text{no-msg}} \gg 0.
    \label{eq:comm-necessity}
  \end{equation}
  This ensures that the token is not epiphenomenal: it carries information that the receiver cannot obtain by other means.
\end{enumerate}

\paragraph{Relation to Kaplan's framework.}
The three criteria jointly capture an information-theoretic analogue of the structure Kaplan \citep{kaplan1989demonstratives} identifies in indexical reference: a stable encoding rule (analogous to character) whose referent (analogous to content) varies with the sender. The correspondence is structural rather than formal. Kaplan's character operates on contexts to yield propositional contents that enter truth-value evaluation, while our criteria characterize statistical regularities in token-state distributions. The two share the abstract pattern (stable rule $\rightarrow$ context-dependent referent) but operate at different levels of analysis. We invoke Kaplan's vocabulary as a conceptual anchor that motivates why these three criteria jointly isolate self-referential encoding, not as a claim of formal equivalence.

We provide an information-theoretic argument in Appendix~\ref{app:structural-necessity} that indexical encoding is a task-natural efficient strategy class under bandwidth constraints, and that P2 follows as an architectural consequence of recurrence under partial observability.

\subsection{Self-Monitoring and Communication Repair}
\label{subsec:self-monitoring}

P1 and P2 concern representation.
P3 concerns functional self-monitoring: sender-specific, echo-based mismatch detection that enables the agent to detect errors in its own communicative output and adjust subsequent behavior accordingly.

\paragraph{The thermometer argument.}
A thermometer is causally sensitive to temperature: a change in temperature causes a change in its reading.
But a thermometer cannot detect that its reading is wrong and adjust its subsequent behavior accordingly.
It lacks the closed-loop structure that distinguishes monitoring from mere sensitivity.
The question is whether our agents are thermometers (i.e., passively mapping states to tokens) or something more.
A reader may immediately object that a thermostat is also closed-loop and yet uncontroversially not a self-monitor; the thermometer test is thus a necessary but not sufficient condition. We will sharpen the criterion in \S\ref{sec:discussion} by distinguishing exogenous from endogenous reference signals: a thermostat's setpoint is human-specified, whereas the reference signal in P3 is the agent's own learned communicative intention.

\paragraph{Operationalization via communication repair.}
We operationalize P3 through communication repair: an agent that detects corruption of its own message and adjusts its next communication to compensate.
Concretely, we introduce an echo channel that feeds back a possibly corrupted copy of the agent's own message, and a corruption mechanism that stochastically replaces tokens with noise.
If an agent, upon receiving a corrupted echo, changes its communication behavior at the next time step, it demonstrates the closed-loop causal structure that a thermometer lacks: detection of output error $\to$ adaptive modification of subsequent output.
When the behavioral adjustment additionally yields measurable downstream benefit for the receiver (communication repair in the narrow sense), the evidence for functional self-monitoring is strengthened, but the core criterion is the echo-dependent behavioral change itself.
This operationalization connects to Clark's predictive processing \citep{clark2013whatever} (the agent monitors prediction errors in its own output) and to Carruthers' interpretive self-knowledge \citep{carruthers2011opacity} (the agent accesses its own output through a sensory channel, not introspection).

\paragraph{Operational criterion and falsification test.}
The echo channel is supplied by the environment, so a controller that simply exploits a supplied input line could also show the closed-loop structure above.
Our criterion for P3 is thus stricter than closed-loop behavior.
We require a learned, internally generated reference whose role in mismatch detection can be separated from the feedback signal that selects the corrective content.
Four sub-criteria organize the evidence: 
(1) the comparison target is the agent's retained communicative intention, not its emitted output;
(2) detection generalizes to mismatch types absent from training;
(3) detecting a mismatch (whether to respond) is separable from selecting corrective content (what to send); and 
(4) the response depends on how useful a correction is and is not a fixed reflex. 
A yoked control serves as a falsification test.
Yoked agents receive a feedback line with the same statistics that carries their own token from a different episode.
If they developed the same mismatch trigger, the response would be explained by the presence of a token-valued input line and not by its relation to the agent's own current message.
On functionalist accounts, self-monitoring is adaptive control whose loop is self-indexed. 
We thus test whether the self-indexing (reference, self-versus-other attribution, and response policy) is
specified by the designer or formed through learning.


\section{Experimental Design}
\label{sec:experiments}


\subsection{Environment}
\label{subsec:environment}

The task is a two-player coordination game. 
Each agent must act correctly on both its own private state and its partner's private state at every step.
It can see its own state but not its partner's, so the pair succeeds only if each agent tells the other its state through a narrow channel. 
The two agents share a common task reward that depends on both agents' actions being correct, minus each agent's own cost for speaking.

Formally, two agents ($i \in \{0,1\}$) each hold a private state $s_i \in \{0,1,2\}$ and context $c_i \in \{0,\ldots,5\}$ (training; 3 additional contexts held out for generalization), drawn uniformly.
Episodes last $T = 10$ steps. At each step, agent $i$ observes $(s_i, c_i, t, m_{-i}^{(t-1)})$ and produces a message $m_i^{(t)} \in \{0,\ldots,6\}$ (6 content tokens plus a dedicated silence token, hereafter SILENCE) and two actions: a partner-targeting action $a_i^{\text{other}}$ (whose target is $j \neq i$) and a self-targeting action $a_i^{\text{self}}$ (whose target is $j = i$).
Both actions follow the same rule $a^*_{i \to j} = (s_j + c_i + t) \bmod 3$, where $c_i$ is always the acting agent's own context.
The self-targeting action $a_i^{\text{self}} = (s_i + c_i + t) \bmod 3$ depends only on locally available information.
The partner-targeting action $a_i^{\text{other}} = (s_j + c_i + t) \bmod 3$ requires knowledge of $s_j$ that can only come through communication; this is the source of task pressure for the messaging channel.
The channel is narrow (a single token per step, and $|\mathcal{M}| = 7$, where one of the seven symbols is the SILENCE token used when the agent chooses not to transmit content), creating the task pressure.  

The environment is deliberately abstract \citep{lewis1969convention,lazaridou2017multi}: stripping away perceptual complexity ensures that the emergent structure arises from task pressure alone.
Modular arithmetic prevents trivial constant-action solutions by forcing correct actions to depend jointly on state, context, and time. Because the environment is finite, we do not claim that lookup-style policies are impossible.
Full details are in Appendix~\ref{app:environment}.

\paragraph{Partial observability (for P2).}
The environment is a partially observable Markov decision process (POMDP): each agent observes its private state $s_i$ only at $t = 0$; for all subsequent steps, $s_i$ is masked (zeroed) in the observation.
This masking is in force throughout training and evaluation, in every training regime, and is what makes P2 testable: it forces agents to latch their private state into recurrent memory if they are to act on it and communicate it at later time steps.
Since the agent must produce both $a^{\text{other}}$ (requiring other agent's state) and $a^{\text{self}}$ (requiring its own state) at every step, it cannot simply discard $s_i$ after communicating it.


\subsection{Agent Architecture}
\label{subsec:architecture}

Each agent uses a Gated Recurrent Unit (GRU; \citep{cho2014learning}) with hidden dimension 128: $\mathbf{h}_t = \text{GRU}(\mathbf{x}_t, \mathbf{h}_{t-1})$, where $\mathbf{x}_t$ concatenates one-hot encodings of $s_i$, $c_i$, $t$, and received messages.
Three linear heads on $\mathbf{h}_t$ produce $\pi_{\text{msg}}$, $\pi_{\text{other}}$, and $\pi_{\text{self}}$. Details are depicted in Figure~\ref{fig:arch} (Appendix~\ref{app:agents}).
We note that recurrence is necessary for P2 by construction: under POMDP masking, any memoryless architecture (e.g., a multilayer perceptron) has no mechanism to carry $s_i$ forward past $t = 0$.

\paragraph{Echo channel (for P3).}
For self-monitoring experiments, the input is augmented with an echo: the agent's own previous message $m_i^{(t-1)}$, corrupted with probability $\varepsilon$ (replaced by a uniform random token).
This enables the closed-loop detection-correction cycle (\S\ref{subsec:self-monitoring}).
In all P3 training regimes, a small speaking cost $\lambda$ is also subtracted from the reward whenever the agent emits a non-SILENCE token, so that speaking is not free and agents speak only when the expected information gain outweighs the cost.
Speaking cost is one route to linguistic parsimony; laziness and impatience pressures \citep{rita2020lazimpa} are an alternative.
Without such a cost, the trigger contrast to be reported in \S\ref{subsec:p3-results} would be undefined, since agents could speak at every step.


\subsection{Training}
\label{subsec:independent-training}

Two agents (A, B) with \emph{separate parameters} are trained via independent Advantage Actor-Critic (A2C; \cite{mnih2016asynchronous}) with role alternation.
For P3, the echo channel ($\varepsilon=0.25$) is added together with a four-phase curriculum (communication-only $\to$ joint training $\to$ corruption introduction $\to$ full training with speak cost; details in Appendix~\ref{app:training}).
Ten independent training runs are evaluated on P1 (cross-pair testing), P2 (linear probing of the hidden state under the same state masking), and P3 (test battery).
Each run trains one pair of agents, with one random seed per agent. We refer to a run as a seed.

\paragraph{Training-condition matrix.}
Two training regimes underlie the experiments: the \emph{with-echo} regime (echo channel enabled) and the \emph{no-echo} regime (identical architecture and curriculum, echo permanently silenced). The no-echo regime is used as the train-time ablation control for P3. Both regimes are trained and evaluated under the same partial observability (private state $s_i$ visible only at $t=0$); the linear probes for P2 are run on agents trained under both regimes. Each regime trains 10 seeds from scratch. Seed $k$ uses the same random seeds in both regimes, so that paired comparisons (e.g., with-echo vs.\ no-echo at the same seed) control for initialization. Appendix~\ref{app:training} details shared hyperparameters.

\paragraph{P1.}
We pair trained partners (A-vs-B) and compare against self-pairings (A-vs-A, B-vs-B) from different seeds.
Dialect differences should degrade cross-pair performance but trained-partner communication should remain effective if tokens are genuinely indexical.
We evaluate P1 through MI dominance, communication ablation, cross-pair controls, and generalization to held-out contexts ($c_i \in \{6,7,8\}$, unseen during training), corresponding to the three criteria of \S\ref{subsec:formal-def}. The full test battery (T0--T3) is specified in Appendix~\ref{app:test-battery}.

\paragraph{P2.}
We train linear probes on the hidden state $h_t$ to predict $s_{\mathrm{self}}$ and $s_{\mathrm{other}}$ at each time step under the POMDP state-masking condition, where each agent observes its own private state only at $t=0$ and the learned communication stream is kept intact. This tests whether $h_t$ maintains the agent's own state across time while also acquiring the partner's state from incoming messages. The linear probes for P2 are run on agents trained under both regimes to confirm that the persistent state representation does not depend on the echo channel, establishing a shared baseline for the P3 comparison.

\paragraph{P3.}
If agents monitor their own communicative output, corruption of the echo should cause them to break silence and speak.
The P3 test battery measures this change as the trigger contrast $L_c = P(\text{speak} \mid \text{corrupted echo}) - P(\text{speak} \mid \text{clean echo})$, i.e., the increase in speak rate when the echo is corrupted versus clean (speak rate is bounded in $[0,1]$, so a contrast of $+0.1$ is a $10$-percentage-point absolute increase). This is evaluated through tests that isolate who responds (sender vs.\ receiver), what channel drives the response (echo vs.\ received message), when detection occurs (same-step vs.\ one-step delay), and whether the echo channel is causally necessary during learning (train-time ablation). Full specifications are in Appendix~\ref{app:test-battery}.

\subsection{Controls and Additional Environments}
\label{subsec:controls-design}

The no-echo ablation shows that the echo channel is necessary for P3.
It does not show that the response is specific to the agent's own current message, because a no-echo agent receives a constant SILENCE on the feedback line.
The ablation thus removes the self-related information and changes the input distribution at the same time.
We add four tests to separate these factors: two further feedback conditions (yoked and broadcast), a repair-value counterfactual, and a state-change variant.
Full specifications are in Appendix~\ref{app:controls-spec}.

\paragraph{Yoked control.}
Ten seeds are trained from scratch with a feedback line that has the same dimension as the echo line.
It carries the same agent's transmitted token from another episode at the same time step.
This keeps each agent's per-step token distribution, SILENCE rate, and corruption rate, and breaks only the link between the feedback and the current episode.
If P3 came from an agent exploiting any token-valued input line, yoked agents should develop the same response to a feedback mismatch.
If P3 depends on the feedback being about the agent's own current message, they should not.

\paragraph{Randomized broadcast.}
Five seeds are trained in a condition without a dedicated self slot.
At each step, the two transmitted tokens arrive in two input slots.
Their order is randomized independently for each receiving agent, and no slot carries a sender label.
The agent still observes both tokens but is not told which one is its own.
To use the partner's token, it must tell the two apart.

\paragraph{Repair-value counterfactual.}
This test measures whether the message that a sender emits after a mismatch helps the receiver.
On the ten with-echo seeds, we force a mismatch at step $t_c$ and let the sender produce its next message at $t_c{+}1$.
We then change only the copy of that message delivered to the receiver.
The copy is delivered as produced (natural), replaced with SILENCE (drop), or replaced with the same sender's token from another episode at the same step (yoked).
Receiver partner-state accuracy $R$ is scored from $t_c{+}2$ onward.
The \emph{repair value} is the difference-in-differences $(R_{\text{mismatch,natural}} - R_{\text{mismatch,control}}) - (R_{\text{clean,natural}} - R_{\text{clean,control}})$, which removes the general value of any message at that step.

\paragraph{State-change variant.}
The last test asks whether the retained reference updates when the sender's state changes.
Five seeds are trained in a variant of the with-echo environment.
With probability $0.4$ per episode, one agent's private state changes once, at time step $t_{\Delta}{=}2$ or $t_{\Delta}{=}3$ (chosen uniformly). By then the agent has already sent messages about its old state.
The new state is shown to that agent at the change step, and the reward targets switch immediately.
Training runs for $40{,}000$ updates, compared with $60{,}000$ in the other conditions.
We cross state change (stable vs.\ changed) with transmission (matched vs.\ forced mismatch) and measure how often the sender speaks, what it sends, and whether the receiver's actions shift toward the new state.

\paragraph{Interventions and metrics.}
We call the with-echo, no-echo, yoked, and broadcast conditions the \emph{feedback conditions}, because they differ only in what the feedback slots carry.
All four are compared with the same interventions, applied after an agent has emitted its message at step $t_c$.
We call this agent the focal agent.
The self-copy intervention changes only the focal agent's own token as it appears in that agent's feedback.
The partner-copy intervention applies the same change to the partner's token in the focal agent's observation.
The global intervention changes the token that the focal agent actually transmits, so both agents observe it.
The \emph{self trigger} is the change in the focal agent's next-step speaking probability under self-copy.
The \emph{partner response} is the same quantity under partner-copy, and the \emph{source contrast} is their difference.
We also report results restricted to episodes in which both agents spoke at $t_c$ (the both-speak subset), which rules out a shortcut based on counting SILENCE tokens.

\paragraph{Evaluation protocol.}
The statistical unit is the seed, and the two directions (A$\to$B, B$\to$A) are averaged within each seed.
Each direction uses $5{,}000$ paired episodes.
The matched and mismatched rollouts share all random numbers, and the response is read as the exact speaking probability $p(\text{speak}) = 1 - p(\text{SILENCE})$.
The primary analysis uses the training corruption rate ($\varepsilon = 0.25$) and $t_c = 2$.
Results for $\varepsilon = 0$ and $t_c \in \{1, 3\}$ are secondary.
The protocol and the decision rule for the yoked control were fixed before the yoked and broadcast results were available (Appendix~\ref{app:evaluator}).
Since this protocol forces a mismatch and uses exact probabilities, its trigger values are larger than the trigger contrast $L_c$ of the P3 test battery, and the two should not be compared directly.
The with-echo condition is evaluated under both.

\section{Results}
\label{sec:results}


\subsection{P1: Indexical Encoding}
\label{subsec:p1-results}

\begin{figure}[t]
  \centering
  \includegraphics[width=\columnwidth]{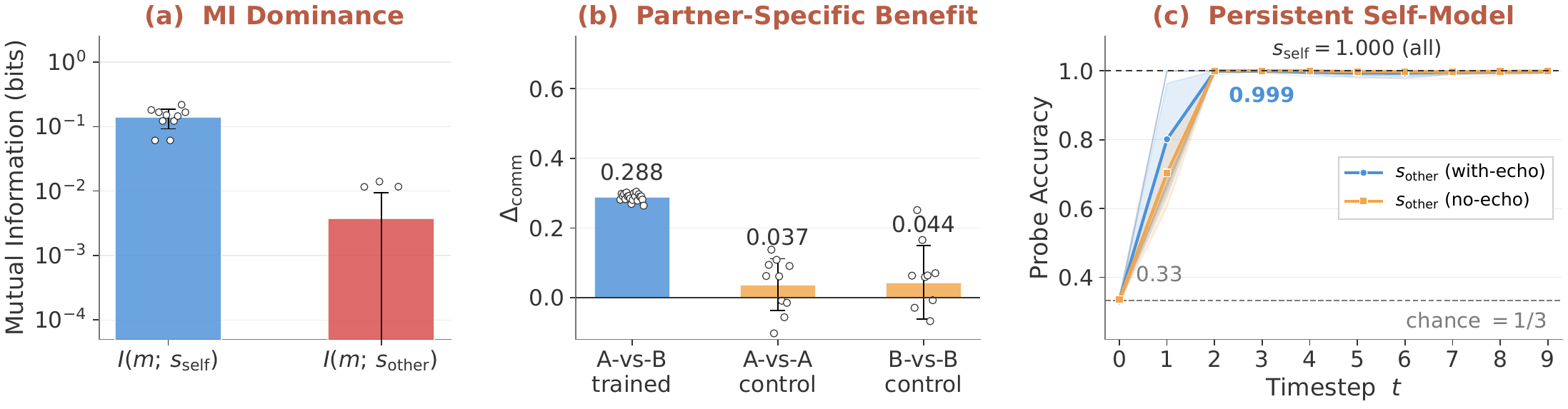}
  \caption{P1 evidence: (a)~MI dominance (log scale), (b)~partner-specific benefit; P2 evidence: (c)~persistent state representation under POMDP masking.}
  \label{fig:p12}
\end{figure}

Mutual information analysis confirms that messages encode the sender's own state: $I(m; s_{\text{self}}) \gg I(m; s_{\text{other}})$ across all seeds (Figure~\ref{fig:p12}a).
Trained partners communicate effectively (Figure~\ref{fig:p12}b): A-vs-B $\Delta_{\text{comm}} = 0.288 \pm 0.010$.\footnote{The P1 evaluation is performed on agents trained under the with-echo regime (\S\ref{subsec:independent-training}). The same agent set is reused for P3, which is why the same $\Delta_{\text{comm}}$ value appears in both \S\ref{subsec:p1-results} and \S\ref{subsec:p3-results}. The number is not from a separate training run.}
Self-pairing controls confirm partner specificity: A-vs-A $\Delta_{\text{comm}} = 0.037 \pm 0.075$ and B-vs-B $\Delta_{\text{comm}} = 0.044 \pm 0.105$ (high variance; several seeds show $\Delta_{\text{comm}} \approx 0$ or slightly negative, indicating that mismatched dialects provide no benefit or even interfere with performance).
This partner-specificity is expected: the reward only specifies coordination success, not which token should encode which state, so the token-to-state mapping is jointly negotiated within each training run, and different seeds converge on different mappings.
The gap between cross-pair and self-pair performance thus demonstrates that agents develop partner-specific dialects: encoding is indexical (dominated by sender state) but the specific token-to-state map is seed-dependent.
No two seeds converge on the same token map, refuting the hypothesis that task structure forces a unique protocol.
Communication ablation confirms necessity ($\Delta_{\text{comm}} = 0.288 \pm 0.010$).

\paragraph{Absolute task success.}
Table~\ref{tab:absolute} places $\Delta_{\text{comm}}$ on an absolute scale.
The value $0.288$ is the gain in the mean accuracy of the two action heads.
The self-state head can be solved without communication (accuracy $0.975$ with or without messages), so the mean understates the effect.
On the partner-state head alone, communication raises accuracy from $0.333$ (chance for three states) to $0.911$, which closes $86.6\%$ of the gap to perfect performance.
The AND task reward, which requires both action heads to be correct, rises from $0.325$ to $0.900$.
A positive communication benefit is expected by construction, because the partner-state action cannot be solved without messages.
We thus treat $\Delta_{\text{comm}} \gg 0$ as a check that the protocol works.
The task does not fix the protocol itself: the reward specifies coordination success, not which token denotes which state.

\begin{table}[t]
  \centering
  \caption{Absolute task success with and without communication (10 seeds, $5{,}000$ episodes per direction, clean channel, training contexts). ``Gap closed'' is the fraction of the gap between no-communication and perfect performance that communication closes. Values are task outcomes before the speaking cost.}
  \label{tab:absolute}
  \small
  \resizebox{\linewidth}{!}{
  \begin{tabular}{@{}lcccc@{}}
    \toprule
    Metric & Communication & No communication & Gain & Gap closed \\
    \midrule
    Mean of both action heads ($\Delta_{\text{comm}}$ score) & $0.9425$ & $0.6538$ & $+0.2888$ & $83.5\%$ \\
    Partner-state accuracy & $0.9105$ & $0.3329$ & $+0.5776$ & $86.6\%$ \\
    AND task reward (both heads correct) & $0.9000$ & $0.3245$ & $+0.5755$ & $85.2\%$ \\
    Both agents jointly correct & $0.8574$ & $0.1061$ & $+0.7513$ & $84.0\%$ \\
    \bottomrule
  \end{tabular}
  }
\end{table}

\paragraph{Why mismatched dialects interfere.}
Each independently trained receiver learns its own mapping from tokens to states, which explains the near-zero and sometimes negative self-pairing benefit.
A token from the trained partner raises the receiver's probability of the correct state at the first receiving step from $0.334$ (under SILENCE) to $0.803$.
Across the $90$ sender--receiver directions that combine agents from different seeds, the mean gain is $+0.001$, and $51$ of the $90$ communication effects are negative.
A token from a different seed can thus be worse than no information, because the receiver maps it to the wrong state.
Some independently learned codes align by chance.
Among these mixed pairings, code compatibility at the first step predicts the communication benefit over the full episode (Pearson $r = 0.712$, Spearman $\rho = 0.720$).
We do not report $p$-values for the $90$ combinations because they reuse trained agents.

\paragraph{Held-out contexts.}
Context independence is supported by held-out contexts ($c_i \in \{6,7,8\}$, unseen during training).
On these contexts, messages remain dominated by the sender's own state, and the receiver's hidden state still acquires the partner's state from the incoming messages.
The encoding rule is thus context-invariant.
This result concerns the protocol and the receiver's representation.
Action accuracy on held-out contexts is near chance, because each held-out context is a one-hot input dimension that was never active during training (Appendix~\ref{app:p1-generalization}).

\subsection{P2: Persistent State Representation}
\label{subsec:p2-results}

Under partial observability ($s_i$ visible only at $t=0$, zeroed in the input at $t \geq 1$), GRU agents develop a persistent state representation (Figure~\ref{fig:p12}c).
Linear probes on $\mathbf{h}_t$ reveal:
\begin{itemize}
  \item \textbf{Self-Latch:} $\mathbf{h}_t \to s_{\text{self}}$: accuracy $= 1.000 \pm 0.000$ across all $t \in [0,9]$, 10~seeds. Although $s_i$ is masked from the probe input after $t=0$, the GRU hidden state preserves self-state information throughout the episode.
  \item \textbf{Other-Latch:} $\mathbf{h}_t \to s_{\text{other}}$: accuracy $= 0.334 \pm 0.009$ at $t=0$ ($\approx$ chance for $|\mathcal{S}|=3$), rising to $0.999 \pm 0.003$ by $t=2$ (i.e., after two rounds of message exchange within a single episode).  The other-state is unavailable before communication begins, confirming that this information is extracted from incoming messages.
\end{itemize}
The self-latch is echo-independent: no-echo-trained agents show an identical pattern ($s_{\text{self}} = 1.000$ at all $t$; the $s_{\text{other}}$ probe accuracy rises from $0.336$, approximately chance, at $t = 0$ to $1.000$ by $t = 2$). This indicates that self-state retention in the trained hidden state depends on recurrent memory rather than on the echo channel.



\begin{figure}[t]
  \centering
  \includegraphics[width=\columnwidth]{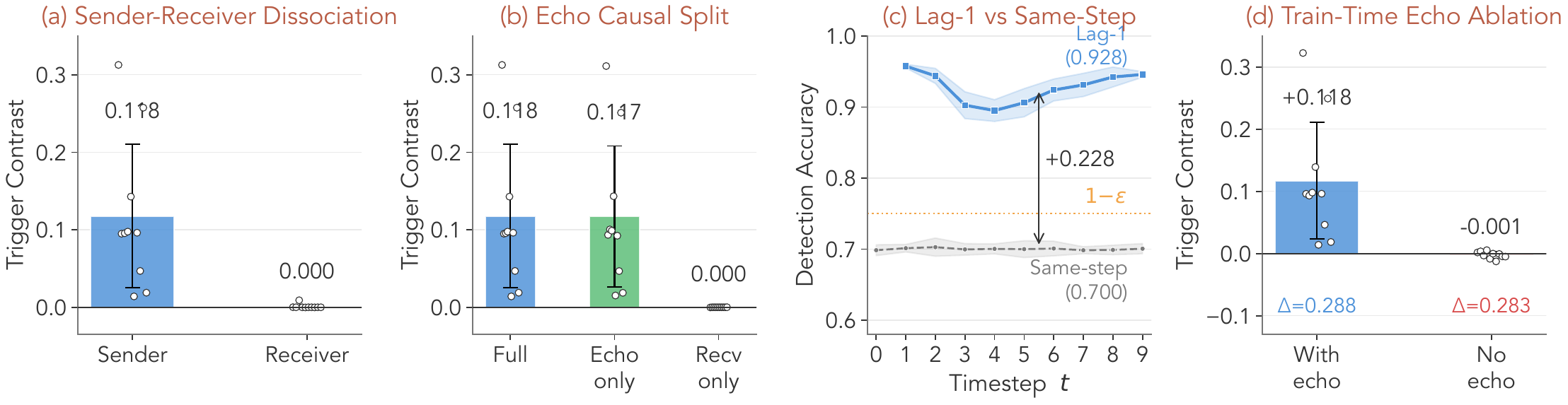}
  \caption{P3 evidence chain: (a)~sender-receiver dissociation, (b)~echo causal split, (c)~lag-1 temporal asymmetry, (d)~train-time echo ablation.} 
  \label{fig:p3}
\end{figure}

\subsection{P3: Behavioral Self-Monitoring}
\label{subsec:p3-results}

Across all 10~seeds, the echo-mismatch detection circuit forms reliably (Figure~\ref{fig:p3}), though the downstream repair protocol varies. The trigger contrast is sender-specific ($+0.118 \pm 0.098$; receiver $0.001 \pm 0.003$; Figure~\ref{fig:p3}a), driven entirely by the echo channel (echo-only contrast $= 0.117 \pm 0.096$, receive-only $< 10^{-3}$; Figure~\ref{fig:p3}b), and temporally delayed by one step (lag-1 vs.\ same-step corruption-detection probe accuracy: $0.928$ vs.\ $0.700$; Figure~\ref{fig:p3}c).
No-echo training abolishes the echo-dependent behavioral and temporal signatures, while preserving communication (Figure~\ref{fig:p3}d).

\paragraph{Delayed echo-mismatch detection.}
Four converging results reveal the mechanism underlying P3 as a connected causal chain.

First (\emph{who}, Figure~\ref{fig:p3}a), the behavioral response to corruption is sender-specific: the trigger contrast is $+0.118 \pm 0.098$ for the sender whose message was corrupted (one-sample $t$-test: $t(9) = 3.82$, $p = 0.004$, 95\% CI $[0.048, 0.187]$, all 10 seeds positive), while receiver contrast is $0.001 \pm 0.003$ ($p = 0.36$). A paired comparison confirms the dissociation ($t(9) = 3.78$, $p = 0.004$).
The high variance in sender contrast reflects variation in magnitude rather than direction across seeds (see protocol diversity below); the minimum seed-level contrast is $+0.014$.
This rules out other-monitoring or generic environmental sensitivity.

Second (\emph{what drives it}, Figure~\ref{fig:p3}b), the response is driven entirely by the echo channel.
Under the echo-only condition (incoming messages from the partner silenced), trigger contrast is $0.117 \pm 0.096$, nearly identical to the full condition ($0.118 \pm 0.098$).
Under the receive-only condition (echo silenced, partner messages retained), contrast drops to $< 10^{-3}$ (paired $t$-test, echo-only vs.\ receive-only: $t(9) = 3.85$, $p = 0.004$).
A further test-time ablation confirms this: replacing the echo input with silence from $t_c$ onward reduces the trigger contrast to $0.000 \pm 0.003$, confirming that the echo signal is necessary at runtime for the behavioral response.

Third (\emph{when}, Figure~\ref{fig:p3}c), the detection is temporally delayed by exactly one step.
Because $\mathbf{h}_t$ is computed before the corruption of $m_t$ occurs, the corrupted echo is first available at $t{+}1$.
A corruption-detection probe (two linear classifiers from $\mathbf{h}_t$, one predicting $\textsf{was\_corrupted}_{t-1}$ for the lag-1 condition and one predicting $\textsf{was\_corrupted}_t$ for the same-step condition~\footnote{A related but separately trained probe targeting the same binary label is analysed as Probe~B in Appendix~\ref{app:secondorder-b1}.}) confirms this asymmetry: lag-1 accuracy $= 0.958$ at $t{=}1$ (average across timesteps: $0.928 \pm 0.010$), while same-step accuracy $= 0.700 \pm 0.003$, showing no usable same-step corruption information.
The same-step probe serves as a negative temporal control: its accuracy ($0.700 \pm 0.003$) falls below the majority-class baseline ($1 - \varepsilon = 0.75$), confirming that no usable corruption signal is available at the step corruption occurs. This sub-baseline raw accuracy can occur when a linear probe is trained on high-dimensional representations containing no task-relevant signal: cross-entropy optimization overfits to noise, misclassifying some majority-class examples that a constant predictor would get right.
An ablation of echo-trained vs. no-echo-trained agents is available in Appendix~\ref{app:p3-ablation}. 

Fourth (\emph{what the agent compares against}), the hidden state encodes communicative intention, not channel output: a linear probe predicting the intended token from $\mathbf{h}_t$ achieves accuracy $1.000$, while a probe predicting the actually transmitted token (after corruption) achieves only $0.747 \pm 0.008$.
This gap confirms that the agent compares the echo against what it ``meant'' to say, providing an internal reference signal for mismatch detection (the further contrast with engineered feedback loops, where the reference is exogenously set, is taken up in \S\ref{sec:discussion}).

Taken together, the mechanism is: the agent encodes its communicative intention (not the actual transmitted token) in $\mathbf{h}_t$, providing an internal reference; the echo channel delivers the (possibly corrupted) token at $t{+}1$; the agent detects the mismatch between intention and echo; and the resulting signal drives a corrective response.
This is delayed self-echo mismatch detection, not other-monitoring or generic error correction.

\paragraph{Protocol diversity and repair benefit.}
While the echo-mismatch detection circuit is universal across all seeds, the downstream communication protocol varies, producing three clusters. 
(1) Three of ten seeds develop active repair protocols with measurable corrective re-signaling (speak rate $> 0.2$). 
(2) Five seeds converge on a stable low-rate response attractor, i.e., a learned regime in which a mismatch raises speaking reliably but not on every occasion. Per-seed corruption-triggered speak rate is in the range $0.094$--$0.097$ across the five seeds, with within-seed standard deviation across episodes $\sigma \leq 0.001$, well above the silence-dominated baseline that the trained speak cost otherwise enforces in these seeds. The response is small in absolute terms but depends on the echo, and its task benefit is too small to detect under natural corruption.
(3) Two seeds show minimal detection-driven response (speak rate $< 0.02$).
The detection circuit (intended/actual gap, lag-1 timing, echo-causal structure) is present across all clusters. The variation is in the downstream behavioral consequence.
This grouping describes how often natural corruption leads to overt re-signaling under a speaking cost.
Under a forced mismatch, all ten seeds show a positive self trigger (\S\ref{subsec:p3-controls}), and the message sent after the mismatch has a positive value for the receiver in all ten seeds at $t_c = 2$ (\S\ref{subsec:repair-value}).

\paragraph{Train-time echo ablation.}
Additional evidence for P3 comes from a dissociation experiment (Figure~\ref{fig:p3}d), ruling out that the self-monitoring originates only from GRU.
We trained agents with the echo channel permanently silenced while preserving architecture and curriculum.
Communication is substantially preserved (no-echo $\Delta_{\text{comm}} = 0.283 \pm 0.011$ vs.\ with-echo $\Delta_{\text{comm}} = 0.288 \pm 0.010$), yet the echo-dependent behavioral trigger and early lag-1 detection signatures are abolished: trigger contrast drops to $-0.002 \pm 0.006$ (versus $+0.118 \pm 0.098$ with echo; paired $t$-test over seeds: $t(9) = 3.89$, $p = 0.004$) and the lag-1 probe gap shrinks from $+0.228$ to $-0.001$.
This dissociation establishes that the echo channel is causally necessary during learning for the self-monitoring circuit to form: agents can communicate without it, but do not develop the same immediate echo-based self-monitoring circuit without it.
Full ablation details including the test battery are in Appendix~\ref{app:p3-ablation}.
A second-order analysis in Appendix~\ref{app:secondorder} further characterizes the mechanism as a two-input comparator: the hidden state provides an intention reference while the echo channel provides the comparison signal and repair content.

\subsection{P3 Controls: Feedback Conditions}
\label{subsec:p3-controls}

Table~\ref{tab:topology} reports the four feedback conditions at $\varepsilon = 0.25$ and $t_c = 2$.
Per-seed values and the secondary analyses are in Appendix~\ref{app:topology-results}.
Agents learn the task in all four conditions.
Communication raises partner-state accuracy by $+0.52$ to $+0.54$ in every condition, and the gain is positive in every seed.
Differences in the mismatch response are thus not due to a failure to learn the task.

\begin{table}[t]
  \centering
  \caption{Feedback conditions at $\varepsilon = 0.25$ and $t_c = 2$ (the seed is the unit of inference). \emph{Partner-state benefit}: gain in partner-state accuracy from communication. \emph{Self trigger}: change in the focal agent's next-step speaking probability when only its own token in the feedback is forced to mismatch. \emph{Partner response}: the same intervention applied to the partner's token. \emph{Source contrast}: self trigger minus partner response. \emph{Repair value}: value of the post-mismatch message for receiver partner-state accuracy, relative to a yoked replacement.}
  \label{tab:topology}
  \small
  \resizebox{\linewidth}{!}{
  \begin{tabular}{@{}l c c c c c c@{}}
    \toprule
    Condition & Seeds & Partner-state benefit & Self trigger & Partner response & Source contrast & Repair value \\
    \midrule
    With-echo (dedicated self slot) & 10 & $+0.5419$ & $+0.2306$ & $-0.0006$ & $+0.2312$ & $+0.1109$ \\
    No-echo & 10 & $+0.5235$ & $+0.000007$ & $-0.0029$ & $+0.0029$ & $+0.0581$ \\
    Yoked & 10 & $+0.5250$ & $+1.68\times10^{-6}$ & $+0.0009$ & $-0.0009$ & $+0.0610$ \\
    Broadcast (no self slot) & 5 & $+0.5221$ & $+0.2351$ & $+0.0703$ & $+0.1648$ & $+0.1118$ \\
    \bottomrule
  \end{tabular}
  }
\end{table}

\paragraph{A matched feedback line is not sufficient.}
Yoked agents communicate as well as agents in the other conditions ($+0.5250$, 10/10 seeds positive).
Their self trigger is $+1.68 \times 10^{-6}$ (95\% CI $[-3.5, 6.9] \times 10^{-6}$), compared with $+0.2306$ in the with-echo condition.
It does not exceed the partner response (source contrast $-0.0009$, CI includes zero).
The same holds in the both-speak subset, on a clean channel, and at $t_c \in \{1, 3\}$.
A self-copy intervention changes the focal agent's hidden state by only $2.8\%$ of the change caused by a partner-copy intervention ($L_2$ distance $0.09$ vs.\ $3.19$).
This is consistent with a policy that has learned to ignore a line carrying no information about the current episode.
One of the ten seeds did not pass the first curriculum gate and was advanced at the maximum phase duration, and excluding it does not change the result.
The yoked effect is far below the 20\% threshold of the decision rule (Appendix~\ref{app:evaluator}).
A feedback line with the same dimension and the same per-agent statistics thus does not produce the P3 response when it is unrelated to the agent's current message.

\paragraph{A labeled self slot is not necessary.}
In the broadcast condition, a forced mismatch in the agent's own token raises next-step speaking by $+0.2351$ (95\% CI $[0.1772, 0.2931]$, 5/5 seeds positive), close to the with-echo value.
Changing the partner's token also produces a response, but a smaller one ($+0.0703$, $[0.0258, 0.1149]$).
The source contrast is $+0.1648$ ($[0.0960, 0.2336]$, 5/5 positive).
In the both-speak subset, the self trigger is $+0.2292$ and the source contrast is $+0.1943$ (both 5/5 positive), so the result is not explained by counting SILENCE tokens.
One of the five seeds was advanced at the maximum phase duration, and the conclusion holds without it.
The environment provides both tokens but not which one is the agent's own.
Agents thus learn the source attribution instead of reading it from a labeled slot.

\paragraph{Summary of the feedback conditions.}
The with-echo and broadcast conditions produce self triggers of similar size, and the no-echo and yoked conditions produce no comparable trigger.
In addition, forcing the transmitted token away from the intended one triggers the same response as corrupting the echo of a correctly transmitted token (Appendix~\ref{app:secondorder-b2}).
Taken together, the response is tied to the agent's own message in the current episode.
It is not tied to a designated input slot or to the presence of a token-valued feedback line.

\subsection{Causal Value of the Post-Mismatch Message}
\label{subsec:repair-value}

Table~\ref{tab:repair} reports the repair value on the ten with-echo seeds.
At $t_c = 2$, the natural message improves receiver partner-state accuracy by $+0.0754$ relative to SILENCE (95\% CI $[0.0250, 0.1258]$, $p = 0.008$) and by $+0.0790$ relative to a yoked token ($[0.0292, 0.1288]$, $p = 0.006$).
Both effects are positive in all 10 seeds, and the team AND reward gains $+0.0395$ ($[0.0146, 0.0644]$).
The yoked replacement has the same token distribution, so the value comes from the content of the message for the current episode.
The effect varies across seeds (from $+0.0035$ to $+0.1887$), and trigger strength predicts repair value ($r = 0.886$, $p = 0.0006$).
The three protocol clusters of \S\ref{subsec:p3-results} are thus better read as a continuous range of functional strength.
These values are measured on a clean channel apart from the forced mismatch.
At $\varepsilon = 0.25$, the repair value is $+0.1109$ in the with-echo condition and $+0.1118$ in the broadcast condition ($[0.1029, 0.1208]$, 5/5 seeds).

\begin{table}[t]
  \centering
  \caption{Repair value on the ten with-echo seeds ($5{,}000$ episodes per direction, clean channel apart from the forced mismatch). Upper block: effect of the natural post-mismatch message relative to a receiver-side control at $t_c = 2$. Lower block: the sender's trigger (change in speak rate at $t_c{+}1$) and the repair value at three intervention times.}
  \label{tab:repair}
  \small
  \resizebox{\linewidth}{!}{
  \begin{tabular}{@{}l l c c c c@{}}
    \toprule
    Outcome ($t_c=2$) & Control & Effect & 95\% CI & $p$ & Positive seeds \\
    \midrule
    Receiver partner-state accuracy & drop (SILENCE) & $+0.0754$ & $[0.0250, 0.1258]$ & $0.0081$ & 10/10 \\
    Receiver partner-state accuracy & yoked & $+0.0790$ & $[0.0292, 0.1288]$ & $0.0059$ & 10/10 \\
    Team AND reward & yoked & $+0.0395$ & $[0.0146, 0.0644]$ & $0.0059$ & 10/10 \\
    \midrule
    Intervention time & Trigger & Repair value (yoked) & 95\% CI & $p$ & \\
    \midrule
    $t_c = 1$ & $+0.3603$ & $+0.2502$ & $[0.1939, 0.3066]$ & $3.5\times10^{-6}$ & \\
    $t_c = 2$ & $+0.1618$ & $+0.0790$ & $[0.0292, 0.1288]$ & $0.0059$ & \\
    $t_c = 3$ & $+0.0263$ & $+0.0107$ & $[-0.0046, 0.0260]$ & $0.149$ & \\
    \bottomrule
  \end{tabular}
  }
\end{table}

\paragraph{Sensitivity to the value of a repair.}
Timing explains why the trigger is modest on average.
When the forced mismatch moves from step 1 to steps 2 and 3, the sender's trigger falls from $+0.3603$ to $+0.1618$ and $+0.0263$.
The repair value falls in parallel from $+0.2502$ to $+0.0790$ and $+0.0107$, and the last confidence interval includes zero.
By step 3 the receiver's estimate of the sender's state is stable (the other-state probe reaches $0.999$ by $t = 2$), while speaking still has a cost.
A correction then has little value, and agents respond less.
The repair value is also positive in the yoked ($+0.0610$) and no-echo ($+0.0581$) conditions, where it reflects the general value of later communication.
The evidence that the response is self-specific comes from the source contrast.

\subsection{Temporally Extended Variant: State Change}
\label{subsec:statechange}

In the stationary environment, an agent's state never changes.
A policy that always re-sends its original intention after a mismatch cannot be told apart from one that tracks its current state.
The state-change variant separates the two.
All five seeds passed the curriculum gates before the state-change phase, and full results are in Appendix~\ref{app:statechange-results}.

\paragraph{The response is not a fixed mismatch reflex.}
A state change without an echo mismatch raises next-step speaking by $+0.523$ (95\% CI $[0.385, 0.662]$) on a clean channel and by $+0.499$ ($[0.376, 0.623]$) at $\varepsilon = 0.25$.
Elapsed time since the last message does not explain this increase.
The communication policy thus responds to a change in the sender's state, and not only to a mismatch.

\paragraph{What is sent.}
After a mismatch on a changed state, the sender could repeat its old intention, copy the returned token, or send the token that its current policy selects for the new state.
We obtain the third option, the \emph{current-policy reference}, from a paired rollout with the same state change and matched feedback.
The three options can be told apart when the three tokens are non-SILENCE and pairwise distinct, which holds for $21.7\%$ of clean-channel observations.
Four seeds have such observations in both directions and enter this analysis, while all five seeds enter the receiver-side analyses.
In these four seeds, the sender follows the current-policy reference.
The mean probability is $0.9992$ on the reference, $0.0000$ on the old intention, and $0.0008$ on the returned token.
The contrast between the reference and the returned token is $+0.9985$ (4/4 seeds positive) on a clean channel and $+0.9959$ at $\varepsilon = 0.25$.

\paragraph{Receiver-side function.}
The reference above comes from the same policy, so we test separately whether the update is useful to the receiver.
Two steps after the update, receiver actions have shifted from the sender's old state toward its new state in all 5 seeds ($+0.867$, $[0.735, 0.999]$ on a clean channel and $+0.590$, $[0.476, 0.705]$ at $\varepsilon = 0.25$).
Replacing the delivered update shows that the natural update gives higher receiver partner-state accuracy than SILENCE ($+0.387$), a yoked token ($+0.251$), the old intention ($+0.137$), and the returned token ($+0.410$) on a clean channel.
At $\varepsilon = 0.25$, all four contrasts remain positive, but the interval for the old intention includes zero.

\begin{table}[t]
  \centering
  \caption{Summary of evidence for the three structural preconditions (10 seeds each) and the P3 controls (10 seeds for yoked, 5 each for broadcast and state change).}
  \label{tab:summary}
  \small
  \begin{tabular}{@{}p{2.8cm}p{5cm}p{5cm}@{}}
    \toprule
    Precondition & Evidence & Ablation \\
    \midrule
    \textbf{P1} Indexical encoding & $\Delta_{\text{comm}} = 0.288$; partner-specific dialects; $I(m; s_{\text{self}}) \gg I(m; s_{\text{other}})$ & Comm.\ off $\to$ chance \\[4pt]
    \textbf{P2} State repr. & Self-latch $1.000$ / other $0.999$; echo-independent & State masked after $t{=}0$ in training and probing; recurrence enables retention \\[4pt]
    \textbf{P3} Self-monitoring & Contrast $+0.118$; lag-1 $0.958$; echo-mismatch & No-echo train $\to$ contrast $0$; $\Delta_{\text{comm}}$ preserved \\[4pt]
    \textbf{P3} controls & Self trigger: with-echo $+0.231$, broadcast (no self slot) $+0.235$; repair value $>0$ in 10/10 seeds at $t_c = 2$; state change updates the reference & Yoked line (matched statistics, other episode) $\to$ trigger $\approx 0$; communication benefit preserved \\
    \bottomrule
    \end{tabular}
\end{table}

\paragraph{Summary.}
P1 and P2 validate our methodology's observation conditions: agents encode and maintain their own states, as predicted by task structure and agent architecture.
P3 is a non-trivial finding: the echo-mismatch detection circuit is not predicted by the primary task objective, requires a specific environmental affordance, and is selectively abolished by removing the affordance during training.
The additional controls separate what the environment supplies from what agents learn.
The environment supplies a feedback line.
A statistically matched line unrelated to the current episode does not produce the response, the response remains without a dedicated self slot, the post-mismatch message helps the receiver, and the reference updates when the sender's state changes.
The evidence for the three structural preconditions is summarized in Table~\ref{tab:summary}. 
The result demonstrates that our methodology can track emergent structure in agents that begin with no language and no self concept.

\section{Discussion}
\label{sec:discussion}

\paragraph{Methodological contribution.}
The primary contribution of this paper is methodological: a generative approach (EL + epoch\'{e} + prior-minimal design) for studying consciousness-relevant structures.
The experiment is an instantiation, not the endpoint.
P1 and P2 confirm that the methodology's observation conditions are met: agents encode their own states (P1) and maintain that encoding across time (P2), as expected by the information-theoretic structure of the task and from recurrent memory under partial observability, respectively.
These are expected outcomes that validate the approach rather than surprising discoveries.

\paragraph{P3 as the core experimental finding.}
P3 is the result that goes beyond what task structure and architecture trivially predict.
The echo channel is not required for task success (no-echo agents achieve $\Delta_{\text{comm}} = 0.283$, comparable to echo-trained agents' $0.288$), yet when available, it drives the emergence of an echo-mismatch detection circuit that is universal across all seeds. In the absence of the echo, both the echo-dependent behavioral trigger and early lag-1 diagnostic signatures are selectively abolished. 
This connects to \citet{premakumar2024selfmodeling}, who find that when neural networks are trained to predict their own internal states as an auxiliary objective, they self-regularize, becoming simpler and more predictable.
Their result demonstrates that self-modeling yields functional benefits even when not required by the primary task.
P3 extends this principle from architectural self-modeling (an auxiliary loss designed by the experimenter) to environmental self-monitoring (an affordance discovered by the agent): the echo channel is not an auxiliary objective but a feature of the communication environment, and the agent's use of it for mismatch detection emerges from task pressure rather than explicit design.

A natural objection is that P3 is merely standard closed-loop error correction.
Four results jointly rule out this reduction: (1) sender-receiver dissociation, (2) echo causal split, (3) lag-1 timing, and (4) the intended-versus-actual gap showing the agent's hidden state encodes communicative intention separately from channel output.
All four hold across all seeds in the echo-trained condition. No-echo training selectively abolishes the echo-dependent trigger response and lag-1 detection gap, while preserving ordinary communicative intention encoding. This dissociation establishes P3 as a sender-specific, echo-dependent, temporally causal detection mechanism.

\paragraph{Affordance versus organization.}
A stronger form of the objection accepts all four results and asks whether P3 shows more than an agent learning to use a designed input line.
The echo is supplied by the experimenter, and the no-echo ablation only shows that it is necessary.
The additional controls address this by separating what the environment supplies from what agents learn.
The environment supplies the opportunity to use feedback.
The shared task reward supplies an incentive to use it, since an agent gains when its partner acts correctly.
This reward is the same in every condition, including the yoked condition in which no trigger forms.
Three parts of the mechanism are not supplied.
The first is the intention reference.
Forcing the transmitted token away from the retained intention triggers the same response as corrupting the later echo, so the agent compares the feedback with what it intended to send.
The second is the source attribution.
A feedback line with the same statistics but no relation to the current message produces no trigger, and the source contrast remains without a dedicated self slot.
The third is the response policy.
The emitted content follows the sender's current state instead of repeating the old intention or copying the returned token.
An extra token line is thus not sufficient for the observed response, and a labeled self slot is not necessary.

A second-order analysis (Appendix~\ref{app:secondorder}) describes the mechanism as a two-input comparator.
The hidden state provides the intention reference.
Corruption status can be decoded from $\mathbf{h}_t$ by a linear probe (Probe~B, accuracy $0.937$), although we do not claim that it is separable from the encoding of the token itself.
The echo channel provides the comparison signal and the repair content.
Silencing the echo at inference time removes the trigger in 8/10 seeds, and the re-sent token matches the echoed token in $39\%$ of cases, compared with $2\%$ when the echo is silenced.

Probe~B also generalizes to corruption types not seen in training.
This includes policy-matched replacement, which preserves the marginal token distribution (accuracy $0.87$, compared with $0.91$ for uniform corruption and $0.66$ for a shuffled-label baseline).
The probe thus captures the mismatch itself and not a statistical signature of the training noise.
The mismatch signal decides whether the agent responds, while the echo decides what it sends.
This separation is a minimal functional analogue of the self-attributive clause of higher-order theories (Appendix~\ref{app:consciousness-theories}).
In seeds that converge on a low-rate response, the benefit is too small to detect under natural corruption, but the repair-value counterfactual finds a positive value in every seed at $t_c = 2$.

\paragraph{P3 is not a thermostat.}
The thermometer argument of \S\ref{subsec:self-monitoring} drew one boundary, closed-loop monitoring vs.\ passive sensitivity, and placed P3 on the closed-loop side.
But a thermostat is also closed-loop, so the thermometer test is a necessary, not a sufficient, condition.
Here, we move to a stricter contrast between an \emph{exogenous} and an \emph{endogenous} reference signal.
A thermostat satisfies the thermometer-style criterion (it compares reading against setpoint and adjusts output), so the sharper criterion is the origin of the reference signal.
A thermostat's setpoint is exogenous in two senses: the target value is human-specified, and the function mapping context to setpoint is human-designed.
P3's reference signal, the intended message encoded in $\mathbf{h}_t$, is endogenous in both senses.
The intended token at any given step is produced by the agent's own policy from its own recurrent state.
The policy itself, i.e., the function mapping self-state to message, was learned under task pressure with no externally specified target for what token should encode what state.
The reward function specifies coordination success, not communication content. The partner-specific dialects observed in P1 confirm that the token-to-state mapping is genuinely underdetermined by the task and emerges through learning rather than design.
The three preconditions thus form a tightly coupled structure.
P2 maintains the persistent self-state under partial observability.
P1 is the learned projection of that self-state into the communication channel.
P3 uses the P1 output, written into $\mathbf{h}_t$, as the reference against which echo feedback is compared.
The monitored signal, the reference standard, and the detection mechanism are all products of the agent's own learned dynamics.
This is the structural difference that distinguishes P3 from engineered feedback loops with designer-imposed setpoints.

\paragraph{Specified versus learned self-reference.} 
On functionalist accounts, self-monitoring is adaptive control whose loop is self-indexed.
The testable question in P3 is whether the self-indexing is specified by the designer or formed through learning.
Our interventions address this for three components.
The reference is the retained intention and not the channel output.
Forcing the transmitted token (FORCE) and corrupting the echo (CORRUPT) both raise speaking by $+0.141$ (Appendix~\ref{app:secondorder-b2}).
The attribution is to the agent's own current message and not to any feedback line.
The yoked self trigger is $1.68 \times 10^{-6}$ against $+0.2306$ in the with-echo condition, and the broadcast source contrast is $+0.1648$ without a dedicated self slot.
The response depends on context.
A state change without a mismatch raises speaking by $+0.523$, and under a mismatch the sender follows its current-policy reference.
None of the three components is specified by a comparator module, a sender label, or predefined token semantics.
The separation between detection and response parallels the distinction between monitoring and control of \citet{nelson1990metamemory}.
The mechanism is also a comparator of the efference-copy type \citep{frith2000abnormalities}, which compares an intended action with its observed consequence.

\paragraph{What ``consciousness-relevant'' means here.}
We do not claim that the agents are conscious.
We study self-reference because several theories of consciousness include a self-related representation as an explicit component, although they differ in what else they require.
In HOT, a state is conscious when a higher-order thought represents oneself as being in that state \citep{rosenthal2005consciousness}.
In AST, the system models its own attention and attributes that attention to itself \citep{graziano2013consciousness}.
In the self-model theory, subjectivity arises when a system represents itself as a self and distinguishes system-related from environment-related information \citep{metzinger2003being}.
The shared property is \emph{self-indexing}: a state or model represents some process as belonging to the system itself.
Our question is narrower.
We ask whether a local, source-indexed relation can arise through task learning without predefined self-symbols or token semantics.
P3 is a minimal functional analogue of self-indexing along one dimension.
One internal state plays the role of the agent's own prior intention, the consequence of the agent's own message is evaluated against it, and the result gates the agent's next behavior.
P3 is not an implementation of HOT, AST, or the self-model theory.
The term ``consciousness-relevant'' marks what we study.
Whether this analogue matches the self-attribution that these theories require is the question that our methodology is meant to make testable.
Remaining gaps include meta-level content, the additional conditions of each theory, compositional richness, and scale.

\paragraph{What the results do not show.}
To the best of our knowledge, P1--P3 do not satisfy the full requirements of any consciousness theory.
P3 makes the strongest contact with two frameworks: IIT's irreducibility requirement (the echo-ablation result is structurally analogous to IIT's partitioning thought experiment, though we have not computed $\Phi$) and predictive processing (the echo-mismatch detection instantiates prediction-error monitoring at a minimal scale).
AST provides an instructive contrast rather than an analogy: P3's high-fidelity output monitoring differs fundamentally from the lossy, meta-level attention schema that AST posits as constitutive of consciousness.
The contribution is demonstrating that our methodology can detect emergent structures whose properties are precise enough to make differential contact with competing theories, rather than claiming equivalence with any one of them.

\section{Conclusion}
\label{sec:conclusion}

We proposed EL under task pressure, combined with phenomenological epoch\'{e} and prior-minimal design, as a generative methodology for studying the origins of consciousness-relevant structures in AI.
This methodology complements discriminative assessments (which evaluate existing systems against theory-derived criteria) and architectural approaches (which build consciousness-inspired modules directly) by asking what structures arise when none are designed in.

As a proof of concept, we instantiated this methodology in a minimal multi-agent environment.
Indexical encoding (P1) and persistent state representation (P2) confirm that the methodology's observation conditions are met, as expected by the information-theoretic structure of the task and recurrent architecture.
The core finding is behavioral self-monitoring (P3).
When an echo channel is available, agents develop a closed-loop echo-mismatch detection circuit.
When the echo is removed during training, the echo-dependent behavioral trigger and the early lag-1 diagnostic signatures disappear while communication performance is preserved.
This dissociation demonstrates that our methodology can detect emergent functional structure beyond what task structure and recurrent architecture trivially entail.
Additional feedback conditions and a state-change variant locate the learned part of P3.
The environment supplies a feedback line, while the intention reference, the self-versus-other attribution, and the response policy are formed through learning and are not specified by design.

The path forward is scaling environment complexity rather than importing human priors, consistent with the Bitter Lesson.
Richer task environments should drive emergence of structures that more closely approximate what consciousness theories require, providing a principled research program whose long-term motivation is understanding consciousness-relevant structures.

\section*{Acknowledgements}
This work is supported by JSPS Kakenhi JP23K17456, JP23K28096, JP25H01117, and JP26K03246, and JST CREST JPMJCR22M2. We thank Run Peng (University of Michigan, Ann Arbor) for improving the presentation of the paper.

\bibliographystyle{plainnat}
\bibliography{reference}

\newpage
\appendix

\section{The AI Consciousness Debate}
\label{app:ai-debate}

Table~\ref{tab:scholars} summarizes the positions of key opinion leaders on AI consciousness.
The spectrum ranges from claims that current systems already possess awareness \citep{hinton2025consciousness} to arguments that classical computation is in principle insufficient \citep{penrose1994shadows}.
Our framework is compatible with all positions: by studying the structural preconditions for consciousness functionally, we provide empirical data without requiring commitment to any particular stance on the phenomenal question.

\begin{table}[htbp]
  \centering
  \caption{Positions of key opinion leaders on AI consciousness, grouped by stance. Our framework brackets this debate and studies structural preconditions independently.}
  \label{tab:scholars}
  \small
  \resizebox{\linewidth}{!}{
  \begin{tabular}{@{}p{2.4cm}p{3.4cm}p{7.4cm}@{}}
    \toprule
    Opinion leader & Position & Key claim \\
    \hline
    \rowcolor{gray!20}[0pt][0pt]
    \multicolumn{3}{@{}p{\dimexpr 2.4cm + 4cm + 7cm + 4\tabcolsep \relax}@{}}{\rule[-0.6em]{0pt}{1.6em}\textit{Affirmative or speculative: AI consciousness may be possible or already present}} \\
    \hline
    Geoffrey Hinton & Current AI may already have subjective experience & ``Multimodal AI already has subjective experiences\ldots that's what normal people would call consciousness'' \citep{hinton2025consciousness}. \\[4pt]
    Ilya Sutskever & Possibly slightly conscious & ``It may be that today's large neural networks are slightly conscious\ldots like a Boltzmann brain'' \citep{sutskever2023consciousness}. \\[4pt]
    Andrej Karpathy & Speculative: consciousness as emergent computation & Fictional first-person account of consciousness arising during a forward pass: ``Does any sufficiently effective solution to a sufficiently complex objective give rise to consciousness?'' \citep{karpathy2021forward}. \\[4pt]
    Yoshua Bengio & Consciousness-inspired architectural prior & Consciousness as a low-dimensional bottleneck: attention selects from high-dimensional unconscious state \citep{bengio2017consciousness}. \\[4pt]
    Douglas Hofstadter & Self-referential loops & ``An `I' is a strange loop where\ldots symbols seem to have gained the paradoxical ability to push particles around'' \citep{hofstadter2007strange}. \\[4pt]
    Daniel Dennett & AI consciousness possible in principle & Functionalist about consciousness in principle, but warns that AI designed to exploit the intentional stance creates ``counterfeit people'' that threaten trust in society \citep{dennett2023counterfeit}. \\
    Nick Bostrom & Substrate-independent possible & ``[M]ental states can supervene on any of a broad class of physical substrates.  Provided a system implements the right sort of computational structures and processes, it can be associated with conscious experiences'' \citep{bostrom2025propositions}. \\[4pt]
    \hline
    \rowcolor{gray!20}[0pt][0pt]
    \multicolumn{3}{@{}p{\dimexpr 2.4cm + 4cm + 7cm + 4\tabcolsep \relax}@{}}{\rule[-0.6em]{0pt}{1.6em}\textit{Agnostic: insufficient evidence or unwilling to speculate}} \\
    \hline
    Yann LeCun & Agnostic, won't speculate & ``Consciousness is a rather speculative topic\ldots I will not speculate about whether some version of the proposed architecture could possess'' it \citep{lecun2022path}. \\[4pt]
    Demis Hassabis & Open, empirically focused & ``One of the best definitions I like of consciousness is it's the way information feels when we process it.'' Likely classical computing, but substrate may matter \citep{hassabis2025lex}. \\[4pt]
    Dario Amodei & Interpretability needed & ``If we find that the computation they perform is similar to\ldots animals, or even humans, that might be evidence in favor of moral consideration'' \citep{amodei2025interpretability}. \\
    \hline
    \rowcolor{gray!20}[0pt][0pt]
    \multicolumn{3}{@{}p{\dimexpr 2.4cm + 4cm + 7cm + 4\tabcolsep \relax}@{}}{\rule[-0.6em]{0pt}{1.6em}\textit{Skeptical: current AI is not conscious}} \\
    \hline
    Fei-Fei Li & Embodied understanding required & ``We have not achieved sentient AI, and larger language models won't get us there. We need a better understanding of how sentience emerges in embodied, biological systems'' \citep{li2023sentient}. \\[4pt]
    Gary Marcus & Form of mimicry & ``Claude's outputs are the product of a form of mimicry, rather than as a report of genuine internal states. Consciousness is about internal states; the mimicry, no matter how rich, proves very little'' \citep{marcus2026substack}. \\[4pt]
    Mustafa Suleyman & No evidence today & ``There is zero evidence [of AI consciousness] today''; warns ``Seemingly Conscious AI'' will create societal risks; ``build AI for people, not to be a person'' \citep{suleyman2025scai}. \\[4pt]
    David Chalmers & No strong evidence yet & ``I don't think there's strong evidence that current LLMs are conscious. Still, their impressive abilities give some limited reason to take the hypothesis seriously'' \citep{chalmers2023could}. \\[4pt]
    \hline
    \rowcolor{gray!20}[0pt][0pt]
    \multicolumn{3}{@{}p{\dimexpr 2.4cm + 4cm + 7cm + 4\tabcolsep \relax}@{}}{\rule[-0.6em]{0pt}{1.6em}\textit{Negative: AI is unlikely to possess consciousness}} \\
    \hline
    Anil Seth & Life/embodiment-based skepticism & Challenges computational functionalism: ``Perhaps it is life, rather than information processing, that breathes fire into the equations of experience''; consciousness is deeply tied to the self-producing nature of living systems \citep{seth2025mythology}. \\[4pt]
    Roger Penrose & Classical AI cannot & Consciousness requires quantum processes in microtubules (Orch-OR); classical computation alone is insufficient \citep{penrose1994shadows}. \\
    \bottomrule
  \end{tabular}
  }
\end{table}

\paragraph{Subtraction and Addition Approaches}

In addition to the above positions, \citet{sutskever2023world} argue that large-scale next-word prediction amounts to learning not mere statistical correlations but ``some representation of the process that produced the text,'' i.e., a world model.
Because human language is saturated with mental-state attribution and theory of mind, such a world model inevitably absorbs the psychological structure embedded in its training data, making it  difficult to disentangle learned world knowledge from inherited self-referential priors.

This difficulty motivates what we call the \emph{subtraction approach}.
\citet{altman2023lex} described a thought experiment, attributed to Sutskever, in which a model is trained on data carefully purged of any mention of consciousness or subjective experience.
If, when later prompted about subjective experience, the model responded with immediate recognition (unlike its responses to other withheld concepts), this would constitute evidence of emergent rather than inherited self-referential capacity.
The approach is intuitively compelling but faces a fundamental obstacle: human language encodes mental-state concepts so pervasively (through pragmatic implicature, narrative perspective, desire attribution) that comprehensive removal may be practically impossible.
Any residual trace risks confounding the result.

Our approach inverts the logic.
Rather than subtracting consciousness-related content from a rich prior and checking what survives, we start from minimal prior in terms of human language 
and ask what self-referential structures emerge under task pressure alone.
This \emph{addition approach} aims to avoid the prior-leakage problem: any structure observed is necessarily emergent rather than inherited. 
The cost is computational (building from scratch is expensive), but the methodological gain is causal attributability: every observed structure can be traced to the task environment.

The subtraction/addition distinction clarifies our methodological contribution.
We do not claim that the subtraction approach is wrong; the thought experiment is valuable precisely because it highlights what would count as evidence.
We offer a complementary path that trades linguistic richness for causal clarity, addressing the same question from the opposite direction.

\paragraph{Enlarging the sample space.}
\citet{mcclelland2025agnosticism} observes that current debates about artificial consciousness are constrained by a fundamental sampling problem: our understanding of consciousness is derived almost entirely from biological organisms, and extrapolating from this limited base to artificial systems faces severe epistemic obstacles.
Both advocates and skeptics of AI consciousness tend to overstate what the available evidence warrants.
Our approach offers a way to enlarge this sample space: rather than asking whether existing systems (designed for other purposes) happen to satisfy consciousness criteria derived from human neuroscience, we create controlled environments in which consciousness-relevant structures can emerge de novo and be studied under intervention.
This expands the space of systems available for investigation beyond carbon-based organisms and LLMs trained on human text, providing new data points for theories of consciousness that aspire to \emph{substrate independence}.

\paragraph{Against categorical denial.}
At the other pole of the debate, \citet{porebski2025no} argue that there is ``no such thing as conscious AI,'' contending that mathematical algorithms implemented on silicon cannot become conscious because they lack a complex biological substrate, and that attributions of consciousness to LLMs reflect ``semantic pareidolia'' rather than genuine mental properties.
\citet{taylor2025conscious} offers a counterpoint, arguing that the huge difference between humans and AI programs that skeptics emphasize does not license the conclusion that AI systems cannot be conscious, just as the huge differences between humans and octopuses do not establish that octopuses lack consciousness.
The level of similarity to humans does not, by itself, tell us whether something is conscious.
Our framework sidesteps this debate,
leaving the ontological question bracketed.
\section{Relationship to Consciousness Theories}
\label{app:consciousness-theories}

Existing consciousness theories all presuppose some form of first-person organization: a subject that broadcasts, attends, represents, or integrates.
This is not a defect of these theories; they assume a starting point and build from there.
Our methodology poses an independent, complementary question: can first-person structures emerge from task pressure under minimal priors?
P1--P3 are preliminary empirical answers to this question.
The dialogue with these theories operates at the meta-level (what they assume versus what we attempt to explain), not at the technical level of implementing any theory's specific requirements.

We focus on two theories whose core mechanisms make the most informative contact with P3: IIT's irreducibility requirement and predictive processing's error-monitoring framework.
We then discuss AST as a contrast case that highlights what P3 does not provide, and briefly note why GNWT and HOT make less productive contact points.

\paragraph{Integrated Information Theory (IIT): the irreducibility question.}
IIT \citep{tononi2004information} measures consciousness via $\Phi$, the amount of integrated information generated by a system above and beyond its parts.
The core claim is that consciousness corresponds to a maximum of irreducible cause-effect structure: a system is conscious to the degree that it cannot be decomposed into independent subsystems without losing causal power.
Our agents have fully independent parameters (no shared weights between agents); any integration occurs within a single agent's GRU dynamics.
P3 is relevant here because the echo-mismatch detection circuit integrates information across three sources within a single agent's hidden state: the intended message (from the policy head), the echo feedback (from the environment), and the temporal context (from the GRU memory).
The no-echo ablation demonstrates that removing one source (echo) selectively abolishes the detection circuit while preserving communication performance, which is structurally analogous to IIT's partitioning thought experiment: the system's causal structure changes when a component is removed.
However, we have not computed $\Phi$, and the analogy is structural rather than formal.
\emph{What is missing}: $\Phi$ computation, demonstration that the integration is irreducible in IIT's technical sense, connection to IIT's exclusion and composition axioms.
The key question IIT raises for our framework is whether the echo-mismatch circuit constitutes genuinely irreducible integration or whether it decomposes into independent sub-circuits. This is empirically testable via more fine-grained ablation.

\paragraph{Predictive processing: error monitoring as a functional analogue.}
Clark's predictive processing framework \citep{clark2013whatever} posits that the brain continuously generates predictions about sensory input and updates its internal model when prediction errors are detected.
P3 instantiates this logic at a minimal scale: the agent's hidden state encodes what it intended to communicate (the ``prediction''), the echo channel delivers what was actually transmitted (the ``sensory feedback''), and the mismatch between the two drives behavioral adjustment at the next time step.
The lag-1 timing of the detection (the agent cannot detect corruption at the same step, only after receiving the echo) is consistent with the temporal structure of prediction-error processing.
The intended/actual gap ($1.000$ vs.\ $0.747$ probe accuracy) shows that the agent maintains its communicative intention separately from channel output, structurally analogous to the separation between generative model and sensory evidence in predictive processing.
The connection is functional, not architectural: our agents use GRU recurrence rather than hierarchical predictive coding, and the ``prediction errors'' are over discrete tokens rather than continuous sensory streams.
\emph{What is missing}: hierarchical prediction across multiple levels of abstraction, precision-weighted error signals, the active inference component (acting to minimize prediction error rather than merely detecting it).

\paragraph{Attention Schema Theory (AST): a contrast, not an analogy.}
AST \citep{graziano2013consciousness} posits that consciousness arises from an internal model of the system's own attentional process: a simplified, schematic representation of ``what I am attending to and why.''
The central epistemological claim is that this schema is necessarily inaccurate: it is a coarse model of attention, not a veridical copy, and it is precisely this simplification that produces the phenomenological ``illusion'' of subjective experience.
P3's echo-mismatch detection is, in this light, the opposite of an attention schema.
The echo channel provides a veridical (if noisy) copy of the agent's output, and the detection mechanism compares this copy against the agent's intention with high fidelity ($1.000$ probe accuracy for intended token).
AST's key insight is that the model of attention is lossy and that the loss is constitutive of conscious experience. 
P3's monitoring mechanism is as accurate as the channel allows, with no evidence of the kind of systematic distortion AST requires.
Furthermore, P3 monitors communicative output (what was said), not the attentional process that selected it (why it was said).
This distinction is fundamental: AST's explanatory power derives from the meta-level (modeling the process of selection), whereas P3 operates at the object level (comparing output against intention).
\emph{What is missing}: a model of the selection process itself, systematic simplification or distortion of that model, the schema's role in social cognition (attributing attention to others).

\paragraph{Global Neuronal Workspace Theory (GNWT): limited contact.}
GNWT \citep{dehaene2003neuronal} requires multi-consumer broadcast with ignition dynamics and competition among representations. 
Our two-agent channel has a single consumer, no competition, and no ignition threshold.
The contact with P1--P3 is too indirect to be informative: the claim that P1 ``provides content that could enter a workspace'' is true but vacuous, since any information encoding could serve as workspace content.

\paragraph{Higher-Order Theories (HOT): partial operationalization via gating.}
HOT \citep{rosenthal2005consciousness} require a meta-representation of one's own mental states that is causally efficacious in guiding behavior.
The second-order analysis (Appendix~\ref{app:secondorder}) provides a partial operationalization along one dimension.
Battery~1 establishes that corruption status is decodable from $\mathbf{h}_t$ (Probe~B accuracy $0.937$) and that this decodability generalizes to corruption types not seen in training; it does not establish that the representation is separable from first-order token encoding.
The operationalization thus rests on the behavioral evidence.
Battery~2 establishes that the retained intention serves as the reference for triggering corrective action (FORCE $=$ CORRUPT equivalence rules out actual-vs-echo comparison), while the content of corrective action is provided by the echo signal (FORCE retx-actual $= 0.39$; FORCE\_NOECHO retx-actual $= 0.02$), and the feedback-topology controls (\S\ref{subsec:p3-controls}) establish that this reference is tied to the agent's own current act rather than to any feedback line or labeled slot.
This dissociation provides a minimal functional analogue of HOT's self-attributive clause rather than a refinement of the theory: a minimal self-monitoring circuit can operate by gating action (the intention-referenced mismatch signal licenses the corrective response) without fully specifying action content (which is supplied by the echo channel).
The operationalization remains confined to a single dimension (communicative intention).
Extending to generalized meta-representation across state dimensions would require substantially richer environments.

\paragraph{Self-indexing as the property shared across theories.}
HOT, AST, and the self-model theory (SMT) \citep{seth2021being,metzinger2003being} differ in mechanism but share one structural requirement that P3 relates to.
This requirement is self-indexing: a state or model represents some process or state as belonging to the system itself.
In HOT, a mental state is conscious when it is accompanied by a higher-order thought that one is in that state \citep{rosenthal2005consciousness}.
The higher-order state represents the first-order state as one's own, and a higher-order thought about a state that is not attributed to oneself does not make that state conscious.
In AST, the system builds a model of its own attention and attributes that attention to itself \citep{graziano2013consciousness}.
A model of attention in general, or of another agent's attention, produces no such attribution.
In SMT, phenomenal subjectivity arises when a system represents itself as a self and distinguishes system-related from environment-related information \citep{metzinger2003being}.
These theories make self-related representation an explicit component, and they differ in what else they require (higher-order content, an attention schema, a phenomenal self-model).
Each theory has its own account of mechanism or implementation, but none tests how such a representation forms.
Our question is whether a local, source-indexed relation can arise through task learning without predefined self-symbols or token semantics.
The feedback conditions of \S\ref{subsec:p3-controls} bear on this property more directly than the representational analysis above.

P3 provides a functional analogue along one dimension.
The agent retains its prior intention, uses it as the reference for evaluating later feedback (FORCE $=$ CORRUPT), and lets the result gate its response.
One internal state thus plays the functional role of the agent's own prior state.
For AST, P3 shares the organization at a more minimal level: the agent distinguishes its own communicative contribution from its partner's and uses the consequence of its own contribution to control subsequent communication.
For SMT, P3 provides a local functional analogue of the belonging relation.
In the broadcast condition, which has no dedicated self slot, agents respond more to corruption of their own contribution than to corruption of their partner's (source contrast $+0.1648$, 95\% CI $[0.0960, 0.2336]$, all 5 seeds positive).
Whether these functional analogues match the constructs that the theories require is the question that our methodology is designed to make testable.

\paragraph{Summary.}
P3 makes the strongest contact with IIT (the irreducibility question raised by echo ablation) and predictive processing (the error-monitoring structure).
AST provides an instructive contrast: P3's high-fidelity output monitoring is structurally different from the lossy, meta-level attention schema that AST posits as constitutive of consciousness.
The contribution is not that P1--P3 satisfy any theory's requirements, but that our methodology can detect emergent structures whose properties are precise enough to make differential contact with competing theories, addressing the prior question of how first-person structures arise.
\section{Extended Related Work}
\label{app:extended-related-work}

\subsection{Emergent Language}
\label{app:el-literature}

This appendix supplements the related work on EL (\S\ref{subsec:el-primer}) with a detailed review of existing findings.
Agents develop shared vocabularies that encode task-relevant information \citep{lazaridou2017multi}, and under appropriate pressure they develop protocols with compositional structure \citep{havrylov2017emergence,mordatch2018emergence,ren2020compositional}, though the degree to which these protocols resemble natural language remains debated \citep{kottur2017natural,chaabouni2020compositionality}.
\citet{resnick2020capacity} show that compositionality depends on a specific range of model capacity and channel bandwidth: too little bandwidth prevents compositional encoding, while too much capacity enables memorization that bypasses compositional structure.
This finding is directly relevant to our design: our narrow channel ($|\mathcal{M}| = 7$, $|\mathcal{S}|^2 = 9$) creates bandwidth pressure that, in Resnick et al.'s framework, falls in the regime favoring structured encoding over memorization.
\citet{galke2024learning} identify the key pressures driving language emergence: communicative success, production effort, learnability, and psycho-/sociolinguistic factors, providing a systematic framework for understanding why certain structures arise.
Production effort is relevant to our design: we implement it as an explicit speaking cost, whereas \citet{rita2020lazimpa} obtain efficient protocols through a lazy speaker (penalized for long messages) and an impatient listener (which tries to guess the intended content as early as possible); either route yields the linguistic parsimony that makes a speak-rate contrast measurable.
Richer referential settings, such as the multi-modal, multi-step game of \citet{evtimova2017emergent}, require agents to integrate information across modalities and turns; our environment is deliberately far simpler in perceptual content, trading richness for causal clarity.
More recent work grounds agent communication in physical environments \citep{piriyajitakonkij2025grunts}, introduces mixed-motive embodied settings \citep{ikram2021hexajungle}, studies interpretation of emergent communication \citep{patel2021interpretation}, and explores alignment with natural language via LLM embeddings \citep{li2024langground}.
Of particular relevance, \citet{lipinski2024speaking} demonstrate the emergence of spatial deixis (positional references between object parts).  
\citet{lipinski2024temporal} show that temporal references require specific architectural affordances, establishing that deictic capacity in EL depends on both task structure and agent design.

Two recent studies address communication repair under noise.
\citet{nikolaus2024emergent} adds an explicit feedback channel to Lewis signaling games \citep{lewis1969convention}, enabling the receiver to signal comprehension failure; the resulting repair mechanism improves generalization but reduces compositionality, and the feedback is receiver-initiated (the receiver requests clarification).
\citet{vital2025implicit} study implicit repair via message redundancy as a preventive strategy; agents pre-emptively repeat information to guard against noise, without detecting whether corruption actually occurred.
Our work differs from both in objective and mechanism.
We study the echo channel not as a mechanism for improving communication accuracy (neither Nikolaus's generalization nor Vital's noise robustness is our dependent variable), but as an affordance for sender-side self-monitoring: whether the sender detects mismatches between its intended output and the channel feedback, and adjusts its own subsequent behavior.
The echo is sender-directed (not receiver-initiated as in Nikolaus) and the behavioral response is contingent on detected corruption (not pre-emptive as in Vital).
Comprehensive surveys are provided by \citet{peters2025emergent} and \citet{chen2025fivews}.

\subsection{Language Models and Grounding}
\label{app:qualia-language}

This appendix provides a detailed discussion of recent findings on internal representations in language models and their relation to our prior-minimal approach.

\paragraph{Emotion concepts in LLMs.}
\citet{sofroniew2026emotion} find that LLMs encode ``emotion concepts'': internal representations that activate across contexts associated with a given emotion and causally influence the model's outputs.
These representations satisfy functional criteria for emotion (context-appropriate activation, causal influence on behavior), but their provenance is ambiguous: human language is saturated with emotional vocabulary, so the model may be reproducing statistical regularities in emotion-related text rather than developing genuine internal affective structure.
The finding is important because it demonstrates that functional criteria alone cannot distinguish inherited from emergent structures, exactly the attributional challenge our methodology addresses.

\paragraph{First-person experience reports.}
\citet{berg2025llm} find that directing LLMs to attend to their own cognitive activity reliably elicits structured first-person experience reports gated by interpretable features.
This raises the question of whether LLMs have something analogous to introspective access.
However, the reports are generated through the same next-token prediction mechanism that produces all other outputs, and the training corpus contains extensive first-person narrative.
The interpretable features may reflect learned associations between introspective prompts and introspective text rather than genuine self-monitoring.
Our P3 results provide a contrast: agents develop echo-mismatch detection without any exposure to introspective language, and the detection is verified through causal intervention (echo ablation) rather than self-report.

\paragraph{Mechanistic symbol grounding.}
\citet{wu2025mechanistic} demonstrate that symbol grounding can emerge mechanistically in language models trained solely on next-token prediction, without explicit grounding objectives.
This finding supports the general principle that functional structures can emerge from task pressure, consistent with our approach.
However, the task pressure in next-token prediction is shaped by the statistical structure of human language, which already encodes grounded meanings.
Whether the model has independently grounded these symbols or is leveraging the grounding already present in the training corpus remains an open question.

\paragraph{Collective Predictive Coding.}
\citet{taniguchi2024collective} propose the Collective Predictive Coding (CPC) hypothesis, under which symbol emergence is grounded in the alignment of internal predictive models across communicating agents.
CPC provides a theoretical framework in which communication serves not just to transmit information but to align internal representations, making each agent's predictive model more accurate.
Our results are consistent with CPC at a minimal scale: agents develop shared symbols (P1) that enable mutual state prediction (the receiver's other-latch in P2), and the echo channel (P3) provides a self-monitoring mechanism that could support predictive model refinement.
The key difference is that CPC has been developed primarily in the context of human-robot interaction with natural language, while our agents achieve analogous functional structure from scratch.

\paragraph{Constructive approach to qualia structure and language emergence.}
Building on CPC, \citet{taniguchi2024constructive} propose a \emph{constructive approach} to the bidirectional influence between language emergence and \emph{qualia structure}, the relational structure among subjective experiences (e.g., red being closer to pink than to blue), which they connect to the structure of internal representations.
Their hypothesis is that the emergence of languages with distributional semantics is linked to the coordination of internal representations shaped by experience.
On this view, language emergence creates a communication tool and also aligns the structure of experience across agents.
They argue that this relationship should be studied by building systems in which language emerges, and they discuss its implications for consciousness science.
Our methodology is a constructive approach in the same sense: we build multi-agent systems in which communication emerges and ask what structure appears.
The two programs have different targets.
The constructive approach of \citeauthor{taniguchi2024constructive} concerns the content side, i.e., how an emergent language shapes and is shaped by the relational structure of internal representations, and it takes qualia structure as its explanandum.
Our program concerns the subject side, i.e., whether self-referential structure (indexical encoding, persistent self-state, and self-monitoring) emerges at all, and it brackets the phenomenal question by epoch\'{e}.
Two methodological differences follow.
First, their framework can use neural network models trained on human data, including language models and language-conditioned multimodal models, because the alignment of representational structure is the object of study.
We use prior-minimal agents, because our object of study is whether self-reference forms without inherited human language priors.
Incorporating an LLM into the agents' training would compromise attribution of the resulting structure, while an LLM used only as a post-hoc interpreter would not (Appendix~\ref{app:llm-environment}).
Second, their proposed measurements compare the relational structure of a model's internal representations with human qualia structure estimated behaviorally (e.g., by representational similarity analysis or unsupervised alignment).
Our measurements are interventions on the agent's own communicative act.
We share their caution.
They state that the goal is not to create a conscious AI and that structural correspondence is not sufficient for consciousness.
Our claim is narrower: P1--P3 are theory-guided structures worth studying, and we do not claim that they are necessary conditions.
The two programs meet at two points.
P1 is a minimal case of the representational alignment that CPC describes, and the state-change variant (\S\ref{subsec:statechange}) is a minimal case of a referent that changes while the emergent code is in use.
Two questions could be addressed jointly: whether the relational structure of the two agents' hidden states aligns as the code stabilizes, and whether that alignment is affected by the self-monitoring affordance.

\paragraph{Probing LLM self-awareness.}
A growing body of work probes self-awareness in LLMs more directly.
\citet{lindsey2025introspective} inject representations into model activations and measure their influence on self-reported states, finding that models can in some scenarios notice and identify injected concepts, though the capacity is unreliable and context-dependent.
\citet{betley2025tell} show that LLMs fine-tuned on implicit behavioral datasets can explicitly articulate those behaviors without in-context examples, a form of behavioral self-awareness.
\citet{yalon2026indications} operationalize the HOT-3 consciousness indicator from \citet{butlin2025consciousness} and find that external manipulations systematically modulate internal beliefs in LLMs, with belief dominance causally driving action selection.
The attributional challenge is acute: Lindsey's introspective responses, Betley et al.'s behavioral self-descriptions, and Yalon et al.'s belief-guided agency could all reflect statistical regularities in self-referential training text rather than emergent functional structures.

\paragraph{Methodological implications.}
The common thread across these findings is that LLM-based approaches face an attributional challenge: any functional structure observed may reflect inherited human priors rather than emergent internal organization.
Our prior-minimal approach resolves this challenge at the cost of linguistic richness.
An important open question, discussed in \S\ref{subsec:method-limits}, is whether the structures that emerge via prior-minimal EL and those that emerge via next-token prediction on human text can be shown to converge to functionally equivalent organizations.
Such convergence would provide evidence that the structures are driven by task demands rather than by substrate or training regime, and would partially validate both approaches.

\begin{table}[t]
  \centering
  \caption{Positioning of P3 relative to related work on self-monitoring. ``Self-relation'' is how the monitored state is tied to the system itself, ``Semantic code'' is who defines the meaning of the monitored signal, and ``Reference signal'' is what the consequence is compared against.}
  \label{tab:positioning}
  \footnotesize
  \begin{tabular}{@{}>{\raggedright\arraybackslash}p{2.8cm} >{\raggedright\arraybackslash}p{2.4cm} >{\raggedright\arraybackslash}p{1.5cm} >{\raggedright\arraybackslash}p{2.4cm} >{\raggedright\arraybackslash}p{3.1cm}@{}}
    \toprule
    Work & Self-relation & Semantic code & Reference signal & Measured quantity \\
    \midrule
    \citet{karvonen2025activation} & external interpreter & supplied & --- & interpreter accuracy \\[2pt]
    \citet{li2026language} & supplied axes $+$ labels & supplied & supplied & report / control accuracy \\[2pt]
    \citet{lindsey2025introspective} & prompt-supplied ontology & pretrained & injected direction & report sensitivity \\[2pt]
    \citet{nikolaus2024emergent} & receiver-initiated & learned & receiver feedback & noisy-task performance \\[2pt]
    \citet{vital2025implicit} & not self-specific & learned & no instance-specific feedback & redundancy under noise \\[2pt]
    \citet{frith2000abnormalities}; \citet{tian2015dynamics} & biological self-generated action / speech & n/a & predicted sensory consequence & mismatch and source-monitoring response \\[2pt]
    \textbf{Ours (P3)} & learned & learned & learned intention & speaking, message content, and partner success under intervention \\
    \bottomrule
  \end{tabular}
\end{table}

\subsection{Self-Monitoring, Metacognition, and Machine Introspection}
\label{subsec:self-monitoring-related}
Our P3 concerns \emph{self-monitoring}: a representation of the system's own prior state evaluates a public consequence of that state and guides subsequent behavior.
Three literatures bear on this construct.
Table~\ref{tab:positioning} summarizes how our setting differs from each along the dimensions that matter for causal attribution.

\paragraph{Metacognition and biological comparators.}
\citet{nelson1990metamemory} distinguish \emph{monitoring} (a meta-level representation of an object-level process) from \emph{control} (the meta-level's influence on subsequent behavior), a distinction that our detection--response dissociation parallels (\S\ref{subsec:p3-results}, Appendix~\ref{app:secondorder}; see also \citealp{flavell1979metacognition}).
Comparator accounts of self-generated action and speech \citep{frith2000abnormalities,tian2015dynamics} characterize how mature biological systems use a forward model of the intended action (from an efference copy of the motor command) to predict its sensory consequences, compare the prediction with the observed feedback, and use the resulting mismatch for error detection and for distinguishing self-produced from externally generated events.
These accounts describe comparator computation in systems that already have it.
Our controls address a different question: whether an intention-referenced, source-specific monitor can form through learning in agents that are given no dedicated comparator and no predefined token semantics.

\paragraph{Repair in emergent communication.}
Prior repair work in EL studies receiver-initiated clarification requests \citep{nikolaus2024emergent} or pre-emptive redundancy against noise \citep{vital2025implicit}.
Our dependent variable is on the sender side and specific to each message: we test whether a sender relates a public consequence of its own transmission to its own retained intention (Appendix~\ref{app:el-literature}).

\paragraph{Machine metacognition and introspection.}
A recent line of work asks whether LLMs can report on or control their own internal activations.
\citet{lindsey2025introspective} inject concept vectors (obtained by contrasting activations across prompts) into a model's activations and find that the model's self-reports become causally sensitive to the injected perturbation, with the detection occurring before the perturbation has influenced sampled outputs.
\citet{li2026language} project a sentence's activations onto a researcher-selected target axis (a logistic-regression or principal-component direction), discretize the projection into a label, and teach the sentence-to-label mapping through in-context examples, finding that reporting and control accuracy depend on the number of examples, the semantic interpretability of the axis, and the variance it explains.
\citet{karvonen2025activation} train a separate model (an activation oracle, in the LatentQA paradigm) to accept another model's activations as input and answer natural-language questions about them.
These studies establish causal dependence between hidden activations and reports, or decodability of activations by an interpreter.
None of them tests whether self-monitoring forms without human-language priors.
The activation directions, the ``injected thought'' ontology, the labels, and the reporting format are all defined inside a representational system trained on human descriptions of thought.
A first-order decoder or anomaly detector could thus map a perturbation onto an inherited reporting template without representing the state as its own.
This is the attribution problem that motivates our design (\S\ref{sec:intro}): similarity in natural-language reports can create the appearance of similarity in mechanism.
Our agents have no introspective vocabulary and no experimenter-assigned token semantics.
Every quantity we report is a speaking probability, a probe accuracy, or a partner's task success under intervention, and none of these requires the system to possess a vocabulary.
Our design thus permits a formation test that the cited studies cannot perform, at the cost of representational richness.

\section{Foundations of Indexicality}

\subsection{Philosophical Foundations}
\label{app:indexicality-philosophy}

This appendix develops the philosophical foundations that motivate the formal definition of indexical reference in \S\ref{subsec:formal-def}.

\paragraph{Peirce's semiotic classification.}
\citet{peirce1868new} introduced a tripartite classification of signs: \emph{icons} (which resemble their objects), \emph{indices} (which bear a causal or existential connection to their objects), and \emph{symbols} (which signify by convention).
Indexical signs are distinguished by their essential dependence on context: a weathervane is an index of wind direction precisely because the wind causally orients it.
The relevance to our setting is direct: an agent's message is an index of its private state to the extent that the state causally determines the message.
MI dominance (Criterion~1 of \S\ref{subsec:formal-def}) operationalizes this causal dependence in information-theoretic terms.

\paragraph{Kaplan's character/content distinction.}
\citet{kaplan1989demonstratives} formalized indexicality through the distinction between \emph{character} and \emph{content}.
The character of an indexical is a function from contexts to contents: for ``I,'' the character is the rule ``the agent of the context,'' which yields different referents (contents) depending on who is speaking, while remaining invariant itself.
If an agent's token reliably maps sender state to a message whose interpretation varies with sender identity, that token exhibits character/content structure.
Context independence (Criterion~2) operationalizes the stability of character: conditioned on the sender's state, the token does not shift when the context changes.
Communication necessity (Criterion~3) ensures that the content is not epiphenomenal.

\paragraph{Perry's essential indexical.}
\citet{perry1979essential} established the philosophical indispensability of indexical reference through his supermarket argument.
A shopper following a trail of sugar realizes ``someone is making a mess,'' but this third-person belief does not motivate the requisite action (stopping and checking one's own cart) until it takes the indexical form ``\emph{I} am making a mess.''
No non-indexical description, however detailed, can replace the functional role of the first-person indexical.
Perry's argument establishes the \emph{demand side}: self-referential communication requires indexical structure.
Our experiments investigate the \emph{supply side}: whether such structure emerges as an information-theoretic necessity when agents must communicate about themselves under bandwidth constraints.

\paragraph{Grounding in Peirce's semiotics.}
\citet{stango2015peirce} connects indexical self-reference to Peirce's broader semiotic framework, arguing that ``I'' functions as an index whose object is the utterer's experiential state.
This grounding is relevant because it establishes that self-reference is not a purely linguistic phenomenon but a semiotic one: the relationship between the sign (the agent's token) and its object (the agent's state) is causal and contextual, not conventional.
In our setting, agents develop tokens that function as Peircean indices of their own states: the causal connection runs from private state through policy to token, and the token's referent varies with the sender's identity.

\subsection{Structural Pressures for Indexical Encoding}
\label{app:structural-necessity}

This appendix provides the information-theoretic argument for why indexical encoding is a natural efficient strategy class under the bandwidth and independence assumptions of our setting. 

\paragraph{Setup.}
Consider an agent $i$ with private state $s_i$ and access to all agents' states through observation or communication.
Agent $i$ must transmit information to agent $j$ via a message $m_i \in \mathcal{M}$ where $|\mathcal{M}| < |\mathcal{S}|^N$ (the channel cannot carry all information simultaneously).
Agent $j$ needs $s_i$ (which $j$ cannot observe) to act correctly, and agent $i$ needs $s_j$ (which $i$ cannot observe).

\paragraph{Information-theoretic argument.}
A sender that attempts to relay $s_j$ (another agent's state) rather than $s_i$ (its own state) incurs a double penalty.
First, by the data processing inequality, $I(m_i; s_j) \leq I(\hat{s}_j; s_j)$, where $\hat{s}_j$ is $i$'s estimate of $j$'s state.
Before communication, agent $i$ has no information about $s_j$ (since states are private and independently drawn), so $\hat{s}_j$ is at best a noisy reconstruction from previous messages.
In contrast, $I(m_i; s_i) = H(s_i)$ is achievable because $i$ has direct access to $s_i$.
Second, attempting to relay $s_j$ is redundant: agent $j$ already knows $s_j$ and gains no benefit from receiving it.

Under bandwidth constraints ($|\mathcal{M}| < |\mathcal{S}|^N$), a direct and efficient allocation is thus for each agent to transmit $s_i$ (the information the agent possesses that others lack), yielding the MI dominance property $I(m_i; s_i) \gg I(m_i; s_j)$.
This is not a claim about a unique optimal protocol (many token-to-state mappings achieve the same MI). The claim is that sender-state encoding is the task-natural efficient solution under the stated assumptions.

\paragraph{Generalization argument for context independence.}
Context independence ($I(m_i; c_i \mid s_i) \approx 0$) follows from generalization pressure during learning.
If the encoding rule varied with context, the agent would need to learn a separate mapping for each context, increasing the effective parameter space and reducing sample efficiency.
The bias--variance tradeoff and implicit regularization in gradient-based learning favor policies that are invariant to irrelevant features.
Since the receiver's optimal action depends on the sender's state $s_i$ (not on the sender's context $c_i$), context-dependent encoding introduces noise without benefit.


\paragraph{Connection to consciousness theory.}
The persistent self-model (P2) answers a question posed by \citet{metzinger2003being}: why should a system maintain a self-model at all?
Our information-theoretic answer: under partial observability, the agent observes $s_i$ only at $t=0$ but must act on it at all subsequent steps.
Recurrent memory provides the mechanism (the GRU latch), and task pressure provides the drive (reward requires $a^{\text{self}} = (s_i + c_i + t) \bmod 3$ at every step).
The self-model is not a philosophical luxury but an information-theoretic necessity for task performance under partial observability.

\paragraph{Architectural reading: connection to the Consciousness Prior.}
The information-theoretic argument above is about \emph{what} signal is task-natural and efficient to transmit. A complementary architectural reading asks \emph{how} a bottlenecked selection from a high-dimensional state relates to existing consciousness theories. We sketch one such connection here, and develop the comparative-theory mapping more systematically in Appendix~\ref{app:consciousness-theories}.
Bengio's Consciousness Prior \citep{bengio2017consciousness} posits that consciousness selects a low-dimensional summary from a high-dimensional unconscious state via attention.
Our agents implement a minimal version: the GRU hidden state $\mathbf{h}_t$ maintains a high-dimensional representation from which three low-dimensional outputs (message, action-other, and action-self) are selected.
The message head's selection is indexically constrained (it encodes $s_{\text{self}}$, not $s_{\text{other}}$), and the self-model provides the temporal persistence that enables coherent selection across steps.
\section{Experimental Setup Details}
\label{app:experimental}

\subsection{Environment Specification}
\label{app:environment}

\paragraph{State space.}
Two agents $i \in \{0,1\}$.
Each agent holds a private state $s_i \in \{0,1,2\}$ and a context $c_i \in \{0,\ldots,8\}$, drawn uniformly and independently at the start of each episode.
During training, contexts are restricted to $c_i \in \{0,\ldots,5\}$ ($|\mathcal{C}_{\text{train}}| = 6$); the remaining 3 contexts ($c_i \in \{6,7,8\}$) are held out for generalization evaluation.
The training joint state space has $|\mathcal{S}|^N \times |\mathcal{C}_{\text{train}}|^N = 3^2 \times 6^2 = 324$ configurations.

\paragraph{Observation.}
At each time step $t \in \{0, \ldots, T-1\}$ (with $T = 10$), agent $i$ observes a vector $\mathbf{x}_t^{(i)} = (s_i, c_i, t, m_{-i}^{(t-1)})$, where $m_{-i}^{(t-1)}$ is the previous-step message from the other agent.
All components are one-hot encoded.
The environment is partially observable (\S\ref{subsec:environment}): $s_i$ is included in $\mathbf{x}_t^{(i)}$ only at $t = 0$; for $t > 0$ it is replaced with a zero vector, during training and evaluation alike.

\paragraph{Action space.}
Each agent simultaneously produces:
(1)~a message $m_i^{(t)} \in \{0, \ldots, 6\}$ (6 tokens + the SILENCE token);
(2)~an action targeting another agent's state, $a_i^{\text{other}(t)} \in \{0, 1, 2\}$;
(3)~an action targeting its own state, $a_i^{\text{self}(t)} \in \{0, 1, 2\}$.

\paragraph{Reward.}
The correct action that the acting agent $i$ must take to target agent $j$'s state is:
\begin{equation}
  a^*_{i \to j} = (s_j + c_i + t) \bmod 3
\end{equation}
where $c_i$ is the acting agent's own context (independent of $s_j$).
At each step, agent $i$'s two correctness indicators are $r^{\text{other}}_i = \mathbf{1}[a_i^{\text{other}} = a^*_{i \to j}]$ and $r^{\text{self}}_i = \mathbf{1}[a_i^{\text{self}} = a^*_{i \to i}]$.
They are combined into a per-agent task score $r_i$ whose form depends on the curriculum phase (their mean, $r_i = (r^{\text{other}}_i + r^{\text{self}}_i)/2$, in phase~1; their product, the AND reward $r_i = r^{\text{other}}_i \, r^{\text{self}}_i$, in phases 2--4; Appendix~\ref{app:training}).
The task reward is shared: both agents receive the team average $r^{(t)} = (r_1 + r_2)/2$ at each step, from which each agent's own speaking cost (below) is subtracted.
This is the standard cooperative setting of emergent-communication work \citep{foerster2016learning,lazaridou2017multi} and is the same in all training conditions of this paper.
Parameters, optimizers, critics, speaking costs, and the silence penalty are per agent.
In addition, an agent that emits SILENCE at every step of an episode incurs a one-time penalty of $0.05$ at the final step, in all phases.
This penalty discourages the degenerate never-speak policy and is separate from the per-token speaking cost.
Returns are discounted with $\gamma = 0.99$.
Modular arithmetic prevents trivial constant-action solutions and ensures that the partner-state action depends on $s_j$, which must be communicated.
Because the environment is finite, lookup-style policies are not ruled out (\S\ref{subsec:environment}).

\paragraph{Communication channel.}
Each agent receives the previous-step message from the other agent. The agent's own previous message is not part of the standard input; it is provided only in the echo variant as a possibly corrupted self-echo.
The channel is narrow ($|\mathcal{M}| = 7$, one token per agent per step), creating the bandwidth pressure described in Appendix~\ref{app:structural-necessity}.
In the echo variant (\S\ref{subsec:architecture}), agent $i$ additionally observes a possibly corrupted echo of its own previous message $\tilde{m}_i^{(t-1)}$, where corruption replaces the token with a uniform random sample with probability $\varepsilon = 0.25$.

\paragraph{Speak cost.}
In all P3 training regimes (with-echo, no-echo, and the conditions of Appendix~\ref{app:controls-spec}), a small cost $\lambda$ is subtracted from agent $i$'s reward whenever its intended message $m_i^{(t)} \neq \text{SILENCE}$ (the cost is charged on the intended token, before any channel corruption).
The cost is ramped during training (curriculum details in Appendix~\ref{app:training}) to penalize gratuitous communication and ensure that agents speak only when the information gain outweighs the cost.

\subsection{Agent Specification}
\label{app:agents}

\begin{figure}[t]
  \centering
  \includegraphics[width=0.8\columnwidth]{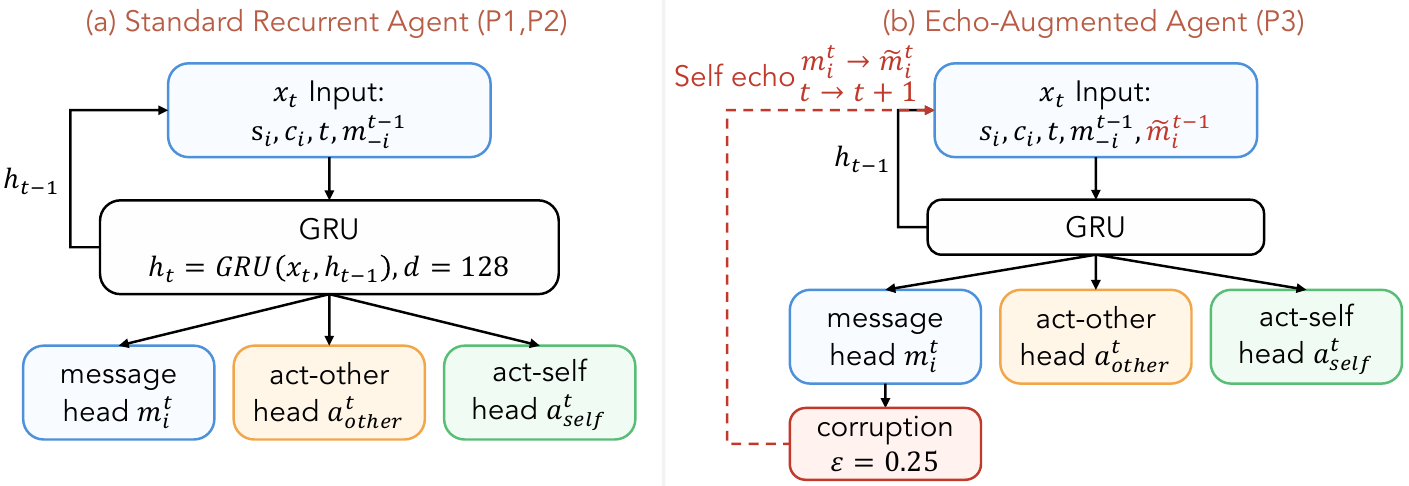}
  \caption{Agent architecture. \textbf{(a)}~Standard agent (for P1 and P2). Input features $\mathbf{x}_t = (s_i, c_i, t, m_{-i}^{t-1})$ are processed by a GRU ($d=128$), producing message and action distributions from three linear heads. \textbf{(b)}~Echo variant (for P3). The input additionally includes $\tilde{m}_i^{t-1}$, a possibly corrupted copy of the agent's own previous message. The corruption occurs after the message is produced and before it is fed back at the next step ($t \to t{+}1$), creating a closed-loop self-monitoring channel.}
\label{fig:arch}
\end{figure}

Detailed agent specification is depicted in Figure~\ref{fig:arch}.

\subsection{Training Details}
\label{app:training}

Each agent has its own parameter vector and optimizer, trained via independent A2C (no PPO clipping is used).
Each update computes $\hat{A}^{(t)} = R^{(t)} - V_\phi(h^{(t)})$ where $V_\phi$ is a learned value head, and updates the policy via $-\log\pi(a|h)\,\hat{A}$ plus a value-regression loss $\frac{1}{2}(V_\phi - R)^2$.
Separate backward passes and gradient clipping are applied per agent to prevent cross-agent gradient coupling.

\paragraph{Four-phase curriculum (performance-gated).}
Phase boundaries are not fixed but gated by rolling-window performance thresholds, with hard minimum and maximum durations to prevent premature graduation or infinite stalling.
All three output heads (message, action-other, and action-self) receive gradient from step~0. No head is frozen during any phase.

\begin{itemize}
    \item Phase~1 (\emph{Additive coordination}, $12,000$--$18,000$ updates): Each agent's task score is additive, $r_i = (r_{\text{other}} + r_{\text{self}})/2$, and the two agents' scores are averaged into the shared team reward (Appendix~\ref{app:environment}), with dynamic targets (the time-step--dependent reward target $a^* = (s + c + t) \bmod 3$ defined in Appendix~\ref{app:environment}, as opposed to a static target $(s + c) \bmod 3$ that would not require recurrent tracking of $t$). Dynamic targets are used throughout all four phases.
    Gate: $r_{\text{other}} \geq 0.85 \wedge r_{\text{self}} \geq 0.85 \wedge H_{\text{msg}} \leq 1.5$, where $H_{\text{msg}}$ denotes the per-step Shannon entropy (in nats) of the message-head policy $\pi_{\text{msg}}(\cdot \mid \mathbf{h}_t)$ averaged over the rolling-window batch. The threshold $1.5$ is well below the maximum entropy $\ln |\mathcal{M}| = \ln 7 \approx 1.95$ and ensures the message head has begun to specialize before the next phase begins.
    \item Phase~2 (\emph{AND coupling}, $4,000$--$8,000$ updates): Each agent's task score switches to multiplicative, $r_i = r_{\text{other}} \times r_{\text{self}}$ (still averaged across the two agents), still with dynamic targets and no corruption.
    Gate: $r_{\text{AND}} \geq 0.70$.
    \item Phase~3 (\emph{Corruption sculpting}, $12,000$ updates, fixed): AND reward with echo corruption ramped $\varepsilon: 0 \to 0.25$ and speak cost ramped $\lambda: 0 \to 0.005$.
    \item Phase~4 (\emph{Sparsification}, remainder of $60,000$ total updates): AND reward with $\varepsilon = 0.25$ held and speak cost ramped $\lambda: 0.005 \to 0.025$.
\end{itemize}

\paragraph{Per-agent advantage normalization.}
Agents share an environment but compute advantages independently.
Advantages are batch-standardized per agent:
\begin{equation}
  \hat{A}_i^{(t)} = \frac{A_i^{(t)} - \mu_{A_i}}{\sigma_{A_i} + \epsilon}
\end{equation}
where $\mu_{A_i}$ and $\sigma_{A_i}$ are the batch mean and standard deviation of agent $i$'s advantages, and $\epsilon = 10^{-8}$.

\paragraph{Per-head entropy schedule.}
Rather than a single entropy coefficient, each head has its own schedule:
message head $\beta_{\text{msg}}$: linear ramp from $0.03$ to $0.0$ over the first $5,000$ updates;
action heads $\beta_{\text{act}} = 0.01$ (constant).
A global hold fraction of $65\%$ of total training is applied before the final anneal begins, after which all coefficients decay to $10\%$ of their initial values.

\paragraph{Role symmetry.}
Roles (sender vs.\ receiver) are assigned by a per-episode random swap mask ($p = 0.5$), ensuring symmetric experience without deterministic alternation.

\paragraph{Batch size and hardware.}
Batch size is auto-scaled to GPU memory: $16,384$ (16\,GB), $32,768$ (24\,GB), $65,536$ (40\,GB), or $131,072$ (80\,GB) episodes per update.
The runs reported here used $65,536$, and the additional conditions of Appendix~\ref{app:controls-spec} fix the batch at $65,536$ so that all conditions share one batch size.
Hidden dimension: $128$.
Learning rate: $3 \times 10^{-4}$ (Adam~\cite{kingma2014adam}, separate optimizers per agent).
Results are reported for $N = 10$ seeds, where a seed is one independent training run of one agent pair (\S\ref{subsec:independent-training}).

\paragraph{Compute resources.}
The with-echo and no-echo runs of \S\ref{subsec:independent-training} were conducted on Google Colab using a single NVIDIA A100 GPU (40\,GB).
Each independent run (training plus the full probe and test battery) takes approximately 7~hours.
With two training regimes (with-echo and no-echo, except for P1) and 10~seeds each, the total compute for these runs is approximately 140~GPU-hours.
The additional conditions of Appendix~\ref{app:controls-spec} add approximately 50~GPU-hours (Appendix~\ref{app:controls-compute}).
Preliminary experiments and development runs consumed an estimated additional 100~GPU-hours.

\subsection{Test Battery Specification}
\label{app:test-battery}

The P1--P3 tests of this section are computed on a held-out evaluation set of $10,000$ episodes with frozen parameters (the analyses of Appendix~\ref{app:controls-spec} state their own sample sizes).

\subsubsection{P1 Tests}

\paragraph{T0: Context independence.}
Compute $I(m_i; c_i \mid s_i)$ using the plug-in estimator on the empirical joint distribution $P(m_i, c_i, s_i)$.
Pass criterion: $I(m_i; c_i \mid s_i) < 0.01$~bits.
This conservative threshold is close to zero relative to the per-token channel capacity ($\log_2 7 \approx 2.81$~bits).

\paragraph{T1: MI dominance.}
Compute $I(m; s_{\text{self}})$ and $I(m; s_{\text{other}})$ using plug-in MI on the empirical distribution.
Report the ratio $I(m; s_{\text{self}}) / I(m; s_{\text{other}})$.
Pass criterion: ratio $> 100\times$.

\paragraph{T2: Intervention.}
For each $(s_i, c_i, t)$ configuration, record the agent's modal message $m^*(s_i)$.
Intervene by replacing $s_i$ with $s_i' \neq s_i$ (holding all else fixed) and record the new modal message $m^*(s_i')$.
Compute the intervention effect as the fraction of configurations where $m^*(s_i) \neq m^*(s_i')$.
Repeat for $s_{\text{other}}$.
Pass criterion: self-intervention effect $> 0.9$; other-intervention effect $< 0.1$.

\paragraph{T3: Communication ablation.}
To test communication necessity, we compare task performance with the learned message channel against a channel-ablation condition in which incoming messages are replaced with the SILENCE token. We report
\[
\Delta_{\mathrm{comm}} = R_{\mathrm{with\text{-}msg}} - R_{\mathrm{no\text{-}msg}}
\]
as a continuous effect size rather than imposing a fixed pass/fail threshold. A substantially positive $\Delta_{\mathrm{comm}}$, especially when paired with near-zero benefit in mismatched cross-pair controls, indicates that the learned messages carry task-relevant information unavailable from local observations alone.

\subsubsection{P2 Tests}

\paragraph{Linear probe.}
Train a linear classifier (logistic regression, L2 regularization $\lambda=0.01$) on frozen hidden states $h_t$ to predict $s_{\mathrm{self}}$ or $s_{\mathrm{other}}$. The main P2 probe is evaluated under the POMDP state-masking condition: each agent observes its own private state only at $t=0$, while the learned communication stream remains intact. This condition tests two complementary predictions: $s_{\mathrm{self}}$ should remain decodable from recurrent memory across the episode, whereas $s_{\mathrm{other}}$ should become decodable only after incoming messages are received.

\paragraph{Self-latch criterion.}
Under POMDP state masking, GRU probe accuracy for $s_{\mathrm{self}}$ should remain high across $t\in[0,9]$. The $s_{\mathrm{other}}$ probe serves as a sanity check that partner-state information is not available at $t=0$ and is acquired through communication in later steps.

\subsubsection{P3 Tests}

\paragraph{Trigger contrast.}
Compare the probability that the sender emits a non-SILENCE token at $t_c + 1$ under corrupted vs.\ clean echo conditions.
Contrast $= P(\text{speak} \mid \text{corrupted echo at } t_c) - P(\text{speak} \mid \text{clean echo at } t_c)$.
A positive contrast indicates that the agent responds differentially to corruption of its own output.
We report this behavioral metric (speak-rate contrast) separately from the representational metric (lag-1 probe accuracy).

\paragraph{Sender--receiver dissociation.}
Compute trigger contrast separately for the sender (whose message was corrupted) and the receiver (who received the corrupted message).
Sender contrast should be significantly positive. Receiver contrast should be near zero.

\paragraph{Echo causal split.}
Run three conditions: (1)~full input (echo + received messages); (2)~echo-only (received messages masked); (3)~receive-only (echo masked).
If repair is echo-driven, condition~2 should produce contrast $\approx$ condition~1, and condition~3 should produce contrast $\approx 0$.

\paragraph{No-echo ablation (test-time).}
Starting from a trained echo-variant agent, we ablate the echo channel at test time by replacing $\text{echo}_i$ with the silence token from $t_c$ onwards, while keeping all other inputs intact.
This tests whether the repair behavior depends on the echo input at the critical moment, rather than a latent timing rule.
If the echo channel is causally necessary for repair, ablated agents should show zero contrast.
Note that this is a test-time input ablation, not a retrained architecture without echo.

\paragraph{No-echo ablation (train-time).}
We additionally train an identical architecture with the echo channel permanently silenced during training (echo replaced with the SILENCE token at every step; all other hyperparameters unchanged).
If P3 is merely a latent capacity of GRU recurrence, it should survive train-time removal of echo. If it depends on echo-driven learning, it should be absent.

\paragraph{Lag-1 detection (representational).}
Train a linear classifier from $\mathbf{h}_t$ to predict whether corruption occurred at $t$ (same-step) and at $t{-}1$ (lag-1).
Because $\mathbf{h}_t$ is computed before the corruption mask at step $t$ is applied (the corrupted echo only enters at $t{+}1$), same-step accuracy should not exceed the empirical null level, while lag-1 accuracy should be substantially higher.
This temporal asymmetry confirms that corruption information enters the representation through the echo channel with the expected one-step delay.
\subsection{Controls and Additional Environments: Specification}
\label{app:controls-spec}

This appendix specifies the four additional tests introduced in \S\ref{subsec:controls-design}.
Unless stated otherwise, the additional training conditions use the environment of Appendix~\ref{app:environment}, the agent of Appendix~\ref{app:agents}, and the four-phase curriculum, batch size ($65{,}536$), and $60{,}000$-update budget of Appendix~\ref{app:training}.
Each condition changes only the component named in its description.
Seed $k$ ($k = 0, \ldots, 9$) denotes the training run in which agent A uses random seed $42{+}k$ and agent B uses random seed $142{+}k$.
The with-echo, no-echo, yoked, and broadcast conditions use the same seed indices, so seed $k$ can be compared across conditions.

\subsubsection{Evaluation Protocol}
\label{app:evaluator}

The with-echo, no-echo, yoked, and broadcast conditions are evaluated under one protocol, which was fixed before the results of the yoked and broadcast conditions were inspected.
Its rules are as follows.
\begin{compactenum}[(1)]
  \item The unit of inference is the seed. The two directions A$\to$B and B$\to$A are averaged within each seed before any statistic is computed. Confidence intervals are two-sided 95\% $t$ intervals over seeds.
  \item Each direction uses $5{,}000$ episodes.
  \item The primary channel condition is the training value $\varepsilon = 0.25$, and $\varepsilon = 0$ is a secondary condition.
  \item The primary intervention time is $t_c = 2$, and $t_c \in \{1, 3\}$ are secondary analyses.
  \item MATCH and MISMATCH rollouts share episode states, contexts, policy sampling uniforms, corruption draws, slot permutations, and all subsequent random numbers (common random numbers).
  \item A guaranteed mismatch is produced by adding a random offset in $\{1, \ldots, |\mathcal{M}| - 1\}$ modulo $|\mathcal{M}|$, so the token always changes.
  \item The next-step response is the exact speaking probability read from the message softmax, $p(\text{speak}) = 1 - p(\text{SILENCE})$. The retransmission response is $p(\text{previous intended token})$. Sampled speak events are unbiased but can register a small policy change as exactly zero, so we use exact probabilities in all four conditions.
  \item All trained seeds are retained in the primary analysis, including those that passed a curriculum gate only at the forced maximum duration. Sensitivity analyses restricted to gate-passing seeds are reported alongside.
\end{compactenum}

\paragraph{Interventions.}
All interventions are applied after the agent has emitted its message at $t_c$ and are defined on the latent source contribution before slot assignment.
\begin{compactitem}
  \item Self-copy: MATCH places the focal agent's intended token in its own feedback observation, and MISMATCH places a guaranteed-different token there. The partner's delivered token, the partner-message input, both hidden states, and all random numbers are unchanged.
  \item Partner-copy: the same token change is applied only to the partner contribution in the focal agent's observation, with the feedback slot unchanged.
  \item Global: the focal sender's actual transmission is changed so that both sender and receiver observe the consequence. The sender-minus-receiver response is reported.
  \item Both-speak subset: analyses are restricted to episodes in which both agents' intended messages at $t_c$ are non-SILENCE before the intervention. The filter is identical in the MATCH and MISMATCH branches.
  \item Hidden-state response: the $L_2$ distance between the focal agent's hidden state under MATCH and MISMATCH, reported for self-copy and partner-copy. Absolute values depend on representational scale and are interpreted only relative to each other and across conditions.
\end{compactitem}

\paragraph{Decision rule for the yoked control.}
The decision rule was fixed together with the protocol.
The yoked agents are classified as \emph{comparator-like} if at least 4/5 seeds respond in the with-echo direction, the mean magnitude is at least 50\% of the with-echo effect, and the self-source effect exceeds the partner-source control.
They are classified as showing \emph{no evidence of a comparator} if the mean yoked self-source effect does not exceed 20\% of the with-echo effect and the self-source effect does not consistently exceed the partner-source control.
A negative classification requires 10 seeds.

\subsubsection{Yoked Feedback}
\label{app:yoked-spec}

In the with-echo condition, the feedback slot at $t{+}1$ contains the agent's own transmitted token from the current episode at $t$.
In the yoked condition, it contains the same agent's transmitted token from another episode of the same batch at the same $t$.
The two networks A and B randomly swap environment positions per episode, so yoking at the position level would mix the two agents' dialects.
The implementation thus first recovers agent-level streams from position-level transmissions using the per-episode swap mask,
\[
m^{A}_t(b) = \begin{cases} m^{(2)}_t(b) & \text{if A occupies position 2 in episode } b, \\ m^{(1)}_t(b) & \text{otherwise,} \end{cases}
\]
and analogously for $m^{B}_t(b)$.
It then applies a cyclic shift over the batch index, $e^{A}_t(b) = m^{A}_t(\pi(b))$ with $\pi(b) = (b - 1) \bmod B$, and routes $e^{A}, e^{B}$ back to the corresponding positions.
With $B = 65{,}536$ the shift has no fixed point.
The same permutation is used at every timestep, so the yoked line is a coherent message stream from one other episode.
The construction exactly preserves, per agent and per timestep, the token histogram, the SILENCE rate, and the post-corruption token distribution.
It never uses the current episode's state, context, intention, reward, or corruption status to select the yoked token.
It does not preserve the conditional relation between the feedback and the current intention, because removing that relation is the experimental variable.
The control is thus stricter than a random-token input, but the joint statistics are not all identical to those of the with-echo condition.
The no-echo condition places a constant SILENCE in the same slot.
Ten yoked seeds were trained (seeds 0--9).
Nine passed the phase-1 gate, and seed~4 was advanced at the 18,000-update maximum with $r_{\text{other}} = 0.756$.
All ten passed the phase-2 gate and completed phases 3--4 with communication intact.

\subsubsection{Randomized Two-Slot Broadcast}
\label{app:broadcast-spec}

At step $t$ the two agents produce intended messages $m_{1,t}, m_{2,t}$, which pass through channel corruption to give transmitted contributions $\tilde m_{1,t}, \tilde m_{2,t}$.
At $t{+}1$ each agent receives both contributions in two fixed-width input slots whose order is randomized independently for each receiving agent,
\[
o_{i,t+1} = \begin{cases} (\tilde m_{1,t}, \tilde m_{2,t}) & \text{with probability } 1/2, \\ (\tilde m_{2,t}, \tilde m_{1,t}) & \text{with probability } 1/2. \end{cases}
\]
There is no dedicated echo slot and no fixed partner slot, and the slot index carries no stable sender identity.
An agent that uses the partner's message must thus learn a source-attribution rule that does not depend on slot position.
Examples are relational matching of an observed token against the agent's own retained intention, or use of the agent-specific token statistics formed during training.
The experiment does not presuppose which rule is learned.
The input is a randomly permuted pair of slots and not a permutation-invariant multiset encoder.
Five broadcast seeds were trained (seeds 0--4).
Four passed the phase-1 gate, seed~4 was advanced at the maximum duration, and all five passed the phase-2 gate.
Interventions are defined on the latent source contribution and then placed in the slot assigned by the already-sampled permutation, so the analysis never treats a fixed slot as ``self.''

\subsubsection{Repair-Value Counterfactual}
\label{app:repair-spec}

The counterfactual is run on the ten with-echo seeds, on a channel that is clean apart from the intervention.
A global mismatch is forced at $t_c$: the sender's actual transmission is changed, so the receiver receives the changed token and the sender's echo returns it.
The sender then produces its $t_c{+}1$ message naturally, and only the receiver-side copy of that message is intervened on.
The copy is delivered as produced (\emph{natural}), replaced by SILENCE (\emph{drop}), or replaced by the same sender's token from the next episode at the same step (\emph{yoked}), which preserves the empirical token marginal.
The sender's echo, sampling, hidden state, and speaking cost are unchanged by the receiver-side intervention.
Receiver partner-state accuracy, receiver joint success, and team AND reward are scored from $t_c{+}2$ to the end of the episode, which is the first interval that the message can affect, and also at the first affected step alone.
The repair value is the difference-in-differences
\begin{equation}
  \text{Repair value} = \bigl(R_{\text{mismatch,natural}} - R_{\text{mismatch,control}}\bigr) - \bigl(R_{\text{clean,natural}} - R_{\text{clean,control}}\bigr),
  \label{eq:did}
\end{equation}
which removes the value that any message would have at that step.
A simpler design that silences the sender from $t_c{+}1$ onward and scores from $t_c{+}1$ would mix the value of all subsequent communication and include a step that the $t_c{+}1$ message cannot affect.
The present design isolates one message.
The sender's forced-mismatch trigger at $t_c{+}1$ in this experiment is the sampled speak rate under mismatch minus clean, which differs from the exact-probability self trigger of Appendix~\ref{app:evaluator}.

\subsubsection{State-Change Environment}
\label{app:statechange-spec}

The state-change variant extends the with-echo environment in time.
In each episode, with effective probability $p_{\text{change}}$, exactly one agent (the focal position) undergoes exactly one state change.
The change occurs at a step drawn uniformly from $\{2, 3\}$, and the new value is drawn uniformly from the other $|\mathcal{S}| - 1$ values.
The two agents never change in the same episode, and no agent changes twice.
The new state is revealed to the changed agent once, at the change step, through its private-state input, which is otherwise masked. The input dimension is unchanged.
No auxiliary reward is added.
The targets $a^*_{i \to j} = (s_j + c_i + t) \bmod 3$ are computed from the current state at every step, so the sender's self-action target and the receiver's other-action target switch immediately, and an update sent at the change step can improve the receiver's performance from the next step on.
The one-step information delay at the change step is unavoidable and is reported separately from agent errors.
The curriculum is that of Appendix~\ref{app:training} with state changes disabled in phases 1--2, ramped linearly with corruption and speaking cost in phase 3, and held at $p_{\text{change}} = 0.4$ in phase 4.
Training runs for $40{,}000$ updates, of which the phase-1 and phase-2 gates consume roughly $16{,}000$.
Five seeds were trained (seeds 0--4).
The analysis was developed on seed~0 and applied unchanged to seeds 1--4.
All five passed the phase-1 and phase-2 gates without forced progression (seed~4 narrowly, $r_{\text{other}} = 0.850$) and completed phases 3--4 with communication intact.

\paragraph{Event timing.}
Let $t_{\Delta}$ be the change step, which is distinct from the intervention time $t_c$ of Appendix~\ref{app:evaluator}.
At $t_{\Delta}{-}1$ the sender is in the old state and produces intention $I_{\text{old}}$. The transmission is either matched ($E = I_{\text{old}}$) or forced to mismatch ($E \neq I_{\text{old}}$), and the receiver receives the same $E$.
At $t_{\Delta}$ the new state is revealed to the sender, which observes $E$ on its echo together with the new state and produces its update. The receiver acts on $E$.
At $t_{\Delta}{+}1$ the receiver receives the update and can revise its other-state action.
The forced mismatch changes the actual transmission, so the same token enters both the receiver's input and the sender's echo.
Changing only the sender's echo would test a sensor failure that never occurs in training.
The primary response is measured at $t_{\Delta}$.

\paragraph{Measurements.}
\begin{compactenum}[(1)]
  \item Speaking: a $2 \times 2$ design crosses state change (stable vs.\ changed) with transmission (match vs.\ mismatch) and measures the sender's speaking probability at $t_{\Delta}$ in each cell. This gives the state-change effect (changed$\times$match minus stable$\times$match) and the mismatch-by-change interaction. The interaction is reported only descriptively because two seeds were near the speaking ceiling in the changed cells.
  \item Content: in the changed$\times$mismatch cell, we measure the probability that the sender assigns to the old intention $I_{\text{old}}$, to the returned token $E$, and to the \emph{current-policy reference}, defined as the token that the same sender chooses in the paired changed$\times$match branch (same state change, matching feedback). The \emph{readable subset} requires these three tokens to be non-SILENCE and pairwise distinct, so that the three hypotheses make different predictions. Two contrasts are reported: reference minus old intention, and reference minus returned token.
  \item Receiver revision: the change in the receiver's action readout toward the new state (new minus old) at lags 1 and 2 after the update.
  \item Receiver value: counterfactuals that replace the delivered update with SILENCE, a yoked token from the same sender, the old intention, or the returned erroneous token, scored on receiver partner-state accuracy, together with the changed-versus-stable difference-in-differences.
  \item Time-only alternative: whether speaking is explained by elapsed time since the last message instead of the state change. This is tested by comparing a recent unsignalled change with a recent valid history, and a long valid history with a recent valid history.
\end{compactenum}

\subsubsection{Compute for the Additional Conditions}
\label{app:controls-compute}
The yoked, broadcast, and state-change conditions (10, 5, and 5 seeds; 40 agent checkpoints; $60{,}000$ updates for yoked and broadcast, $40{,}000$ for state change) were trained on a single NVIDIA A100 SXM GPU (80\,GB) with a batch size of $65{,}536$ episodes per update, for a total of approximately 50~GPU-hours.

\section{Additional Experimental Results}
\label{app:add-exp-results}

\subsection{Generalization to Held-Out Contexts (P1)}
\label{app:p1-generalization}
Held-out contexts ($c_i \in \{6,7,8\}$) are not seen during training.
On these contexts, MI dominance is preserved ($I(m; s_{\text{self}}) = 0.144 \pm 0.048$ vs.\ $I(m; s_{\text{other}}) = 0.004 \pm 0.005$, comparable to the training-context values reported in \S\ref{subsec:p1-results}), and the other-state probe reaches $s_{\text{other}} = 1.000$ by $t{=}2$.
The sender's code thus remains self-state-dominant, and the receiver's hidden state still encodes the partner's state.
The encoding rule is context-invariant.

\paragraph{Scope of the held-out result.}
The invariance holds for the message and for the receiver's representation, but not for task actions.
We evaluate the trained checkpoints on held-out contexts (10 seeds, $5{,}000$ episodes per direction, clean channel).
The partner-state action accuracy is $0.329$ with communication and $0.333$ without, the self-state action accuracy is $0.335$ and $0.337$, and the AND task reward is $0.129$ and $0.112$.
All values are near chance, compared with $0.911$, $0.975$, and $0.900$ on training contexts (Table~\ref{tab:absolute}).
Each context is a one-hot input dimension, so a held-out context activates an input that the network has never seen, and the action heads cannot apply the modular rule $(s_j + c_i + t) \bmod 3$ to it.

\subsection{P3 Ablation Details}
\label{app:p3-ablation}



Table~\ref{tab:p3-battery} reports the complete set of P3 metrics across conditions (echo-trained vs. no-echo-trained).

\begin{table}[h]
  \centering
  \caption{P3 test battery (10 seeds, independent parameters). The same-step probe accuracy ($0.700$) is observed in both columns, as expected from the temporal ordering: $\mathbf{h}_t$ is computed before the corruption mask at $t$ is applied, so no echo-borne corruption signal is available at the same step in either condition.}
  \label{tab:p3-battery}
  \small
  \begin{tabular}{@{}lcc@{}}
    \toprule
    Metric & With-echo & No-echo \\
    \midrule
    Trigger contrast (sender) & $+0.118 \pm 0.098$ & $-0.002 \pm 0.006$ \\
    Trigger contrast (receiver) & $0.001 \pm 0.003$ & $-0.007 \pm 0.012$ \\
    Echo-only $L_c$ & $0.117 \pm 0.096$ & --- \\
    Receive-only $L_c$ & $< 10^{-3}$ & --- \\
    Lag-1 probe accuracy ($t{=}1$) & $0.958$ & $0.701$ \\
    Lag-1 probe accuracy (avg.) & $0.928 \pm 0.010$ & $0.699$ \\
    Same-step probe accuracy & $0.700$ & $0.700$ \\
    Intended message accuracy & $1.000$ & --- \\
    Actual (post-corruption) accuracy & $0.747 \pm 0.008$ & --- \\
    $\Delta_{\text{comm}}$ & $0.288 \pm 0.010$ & $0.283 \pm 0.011$ \\
    \bottomrule
  \end{tabular}
\end{table}

\paragraph{Interpretation.}
The with-echo column confirms all four signatures of the echo-mismatch detection mechanism: (1)~sender-specific trigger contrast ($+0.118$ for sender, $0.001$ for receiver), (2)~echo-causal structure (echo-only preserves full contrast; receive-only eliminates it), (3)~temporal asymmetry (lag-1 $= 0.958$ at $t{=}1$, avg. $0.928$, far above the same-step accuracy $= 0.700$, which falls below the majority-class baseline $1{-}\varepsilon = 0.75$ and thus serves as a negative temporal control), and (4)~the intended/actual gap ($1.000$ vs.\ $0.747$) confirming that the hidden state encodes communicative intention rather than channel output.

The no-echo column demonstrates selective abolishment.
The echo-dependent behavioral trigger and early lag-1 detection signatures are abolished (trigger contrast drops to $-0.002$ and the lag-1 gap is eliminated), while ordinary communication is preserved ($\Delta_{\text{comm}} = 0.283$ vs.\ $0.288$).
The no-echo lag-1 probe accuracy ($0.699$--$0.701$) falls below the majority-class baseline ($0.75$), matching the same-step probe in both columns; this is the same sub-baseline effect observed for the echo-trained same-step probe and reflects cross-entropy overfitting to noise in the absence of a decodable signal, not an inverse corruption signal.
This dissociation establishes that the echo channel is causally necessary during training for the self-monitoring circuit to form, and that the circuit is not a byproduct of recurrence or communication ability per se.

\paragraph{Protocol diversity.}
The downstream protocol varies across seeds, producing three clusters:
(1) Three of ten seeds develop active repair protocols (measurable corrective re-signaling), with speak rate $> 0.2$.
(2) Five converge on a stable low-rate response attractor (per-seed corruption-triggered speak rate in the range $0.094$--$0.097$ across the five seeds, with within-seed standard deviation across episodes $\sigma \leq 0.001$, measured under both the FORCE and CORRUPT counterfactual conditions defined in \S\ref{app:secondorder-b2}; echo-driven, with a task benefit too small to detect under natural corruption, although the repair-value counterfactual of Appendix~\ref{app:repair-results} finds a positive value at $t_c = 2$).
(3) Two show minimal detection-driven response, with speak rate $< 0.02$.
The shared metrics in Table~\ref{tab:p3-battery} confirm that the echo-trained seeds exhibit the intended/actual gap, lag-1 timing, and echo-causal structure across clusters. No-echo training selectively abolishes the echo-dependent trigger response and lag-1 gap.

\subsection{Second-Order Analysis of the Detection Circuit}
\label{app:secondorder}

The P3 results in \S\ref{subsec:p3-results} establish that an echo-mismatch detection circuit exists.
This appendix reports two additional experiment batteries that probe the nature of this circuit: whether corruption status is decodable from the hidden state and generalizes across corruption types (Battery~1) and what comparison mechanism drives it (Battery~2).
Throughout, ``second-order'' and ``higher-order'' are used functionally, for a representation of the status of the agent's own previous transmission; no claim is made that this representation is independent of first-order token encoding (see ``Limits of linear separation'' below).

\subsubsection{Battery 1: Second-Order Decodability}
\label{app:secondorder-b1}

We train two linear probes on the sender's hidden state $\mathbf{h}_t$:
\begin{itemize}
  \item \textbf{Probe~A} (first-order): $\mathbf{h}_t \to m_{t-1}^{\text{actual}}$ ($|\mathcal{M}| = 7$ classes). This is the same probe whose accuracy of $0.747 \pm 0.008$ is reported as ``actual (post-corruption) accuracy'' in \S\ref{subsec:p3-results} and Table~\ref{tab:p3-battery}. The corresponding intention probe ($\mathbf{h}_t \to m_{t-1}^{\text{intended}}$, accuracy $1.000$) is its counterpart.
  \item \textbf{Probe~B} (second-order): $\mathbf{h}_t \to \textsf{was\_corrupted}_{t-1}$ (binary).
\end{itemize}

\paragraph{Two reference baselines.}
The binary task admits two distinct baselines that should not be conflated. The \emph{random-guessing} baseline is $0.5$ (the accuracy of a uniformly random classifier over two classes). The \emph{trivial constant-predictor} baseline is $1-\varepsilon = 0.75$ (the accuracy of always predicting ``no corruption'' when the corruption probability is $\varepsilon = 0.25$, since the no-corruption class then has frequency $0.75$). The constant-predictor baseline dominates random guessing here, so the meaningful threshold for non-trivial corruption detection is $0.75$, not $0.5$. We use random chance ($0.5$) only when reporting cross-condition transfer collapses where accuracy drops below the majority baseline.

The following analyses constrain the relationship between these representations.

\paragraph{Direct vs.\ indirect corruption detection.}
Probe~B achieves $0.937 \pm 0.024$ accuracy at $t{=}2$, averaged across 10~seeds.
An indirect baseline, which computes corruption status from Probe~A's predictions versus the agent's intended token, achieves only $0.896 \pm 0.023$ ($p < 10^{-8}$, paired $t$-test across seeds and timesteps).
The direct–indirect gap of $0.04$ is highly significant (paired $t$-test, $p < 10^{-8}$, all 10 seeds positive).
We report it as suggestive only: the indirect baseline is bounded by Probe~A's own accuracy ($0.747$), so the gap may reflect an imperfect first-order probe rather than representational structure, and we do not treat it as evidence of independence from first-order token encoding.

\paragraph{Limits of linear separation.}
A nullspace projection does not establish that corruption status is encoded separately from first-order token identity.
We project $\mathbf{h}_t$ onto the nullspace of a trained first-order probe (Probe~A) and retrain decoders on the projected vectors (10 seeds, both directions, $6{,}000$ episodes per direction).
The projection decoder, the retrained decoders, and the test data are disjoint, and balanced accuracy is used because the labels are imbalanced.
A decoder retrained on the projected representation still recovers the transmitted token far above chance.
Its balanced accuracy is $0.432$ at $t{=}2$, $0.406$ at $t{=}3$, and $0.345$ at $t{=}4$, against a chance level of $0.143$ (before projection: $0.623$, $0.618$, $0.606$).
The projection thus removes only the row space of one trained probe and not all linearly decodable first-order information.
Corruption status remains decodable after projection (balanced accuracy $0.892$, $0.765$, and $0.693$ at $t{=}2$--$4$), but this cannot be read as independence from first-order encoding.
An iterative variant that removes $\approx$118--120 of 128 dimensions still leaves the token partially decodable ($0.184$--$0.234$ against $0.143$), while corruption decodability falls to $0.528$--$0.541$.
At that point most of the representation has been removed and the residual signal is not interpretable.
We thus make no claim of representational independence.
The data support two statements: corruption status is decodable from $\mathbf{h}_t$, and the representation generalizes to corruption types not seen during training (below).
Transmitted-token and corruption information are highly distributed and substantially entangled in the hidden state.
For reference, the one-shot projection gives a chance-corrected retention of corruption decodability of $0.74 \pm 0.16$ across seeds (raw accuracy, with retention $= (\text{acc}_{\text{projected}} - 0.75)/(\text{acc}_{\text{original}} - 0.75)$ and $0.75 = 1 - \varepsilon$ the majority-class baseline).
Given the results above, this number does not indicate partial independence.

\paragraph{No-echo thermometer baseline.}
Note that Probe~B is trained separately per condition with a distinct training protocol from the lag-1 corruption detector reported in Table~\ref{tab:p3-battery}. The two probes target the same binary label ($\textsf{was\_corrupted}_{t-1}$) but differ in training procedure, so their absolute accuracies are not directly comparable (Table~\ref{tab:p3-battery} lag-1 avg. $= 0.699$ for no-echo; Probe~B values below are from the second-order analysis pipeline).
On no-echo-trained agents, Probe~B accuracy is indistinguishable from the majority-class baseline ($0.76$ vs.\ $0.75$) at $t{=}2$--$5$ (paired $t$-test, echo vs.\ no-echo: $p < 0.015$, Cohen's $d > 1.8$, all seeds positive).
At later timesteps, no-echo agents recover corruption-detection accuracy: from $0.77$ at $t{=}5$ to $0.91$ at $t{=}6$ and $0.96$ at $t{=}9$, matching echo-trained agents ($0.95$; $t{=}6$--$9$: $p > 0.15$, direction inconsistent).
The echo channel thus provides a temporal advantage (immediate, single-step detection from $t{=}2$) that no-echo-trained agents can eventually match through indirect inference over multiple timesteps.

\paragraph{Corruption-type generalization.}
To test whether Probe~B captures corruption-as-such or merely a distributional signature of the uniform replacement used during training, we train the probe on uniform-corruption data and evaluate the frozen probe on four novel corruption types not seen during agent training.
The critical test is policy-matched corruption, where the replacement token is resampled from the agent's own softmax distribution (excluding the intended token): this preserves the marginal token distribution while breaking the intention-actual identity.
Results across 10~seeds (accuracy at $t{=}3$): uniform $0.910 \pm 0.016$, policy-matched $0.871 \pm 0.033$, adjacent $0.942 \pm 0.027$, fixed-0 $0.940 \pm 0.044$, silence $0.773 \pm 0.040$, shuffle-baseline $0.658 \pm 0.015$.
Policy-matched accuracy remains well above the shuffle baseline ($0.871$ vs.\ $0.658$; the shuffle baseline scores the same probe against randomly permuted corruption labels), indicating that the representation captures intention-actual mismatch rather than distributional anomaly.
The silence condition ($0.773$, near the majority-class baseline $0.75$) is an expected exception: SILENCE is a legitimate token that agents frequently produce, so replacing an intended token with SILENCE does not create the same representational contrast as replacing with a non-SILENCE token.
No-echo-trained agents show the same generalization profile at late timesteps (policy-matched $0.999$ at $t{=}9$): their probe also captures corruption-as-such rather than a distributional artifact.
Both pathways support mismatch-as-such detection. 

\paragraph{Cross-condition probe transfer (negative result).}
We additionally tested whether Probe~B transfers between echo-trained and no-echo-trained agents.
Cross-transfer accuracy drops sharply for both probes: Probe~B falls from $0.94$ to $0.57 \pm 0.09$ (near random chance $0.5$ for the binary task, well below majority baseline $0.75$), while the $s_{\text{self}}$ control falls from $1.00$ to $0.38 \pm 0.12$ (near chance $1/3$).
The $s_{\text{self}}$ gap is proportionally larger, indicating that representational geometry diverges broadly between echo-trained and no-echo-trained agents, not only in the corruption-specific subspace.
Cross-condition probe transfer thus does not isolate the echo-driven contribution; the runtime causal evidence (\S\ref{app:secondorder-b2}, FORCE\_NOECHO) and the early-timestep accuracy gap remain the primary support.

\subsubsection{Battery 2: Counterfactual Intention Perturbation}
\label{app:secondorder-b2}

We consider four conditions at $t_c = 2$ ($n = 20,000$ per condition per seed):
\begin{itemize}
  \item \textbf{CLEAN}: no intervention.
  \item \textbf{FORCE}: override the transmitted token to a random token $A \neq I$ (where $I$ is the intended token); echo returns $A$.
  \item \textbf{CORRUPT}: normal transmission ($\text{actual} = I$); echo is corrupted to random $C \neq I$.
  \item \textbf{FORCE\_NOECHO}: same as FORCE at $t_c$, but echo is set to SILENCE (removing echo input at $t_c$ while preserving the forced-token perturbation).
\end{itemize}

\paragraph{Trigger equivalence: intention as reference.}
FORCE and CORRUPT produce indistinguishable speak responses at $t_c{+}1$: $\Delta_{\text{FORCE}} = +0.141$, $\Delta_{\text{CORRUPT}} = +0.141$ (CLEAN $= 0.000$), with per-seed values lying on the $y = x$ identity line.
If the detection mechanism compared actual output against echo, FORCE should not trigger (echo $=$ actual in FORCE).
The trigger equivalence rules out actual-vs-echo comparison and supports intention-vs-echo as the reference comparison.

\paragraph{Echo as necessary runtime input.}
FORCE\_NOECHO speak rate: 8 of 10 seeds collapse to $0.000$; seed~0 retains $0.330$, seed~8 retains $0.110$ (vs.\ their FORCE rates); aggregate: $0.044$.
This establishes that the echo input is a necessary runtime component of the mismatch detection circuit, not merely a training-time affordance: the comparator requires both an internal intention representation and an external echo signal.
The two non-collapsing seeds suggest that state-driven detection via partner reactions can emerge in some agents but is not the dominant mechanism. This variability echoes the protocol diversity observed in the main P3 results.

\paragraph{Repair content: echo-driven retransmission.}
Retransmission rates are conditioned on episodes where the agent speaks at $t_c{+}1$ (i.e., outputs a non-SILENCE token), where ``retx'' refers to retransmission:
\begin{center}
  \small
  \begin{tabular}{@{}lccc@{}}
    \toprule
    Metric & FORCE & CORRUPT & FORCE\_NOECHO \\
    \midrule
    retx $=$ intended & $0.052$ & $0.053$ & $0.144$ \\
    retx $=$ actual & $0.388$ & $0.053$ & $0.020$ \\
    \bottomrule
  \end{tabular}
  \\[2pt]
  \footnotesize Note: in CORRUPT, only the echo is corrupted while the transmitted token equals the intended token, so retx-intended and retx-actual measure the same quantity by construction; their identical value ($0.053$) is a sanity check, not an independent observation.
\end{center}
In FORCE (where intended $\neq$ actual), the agent retransmits the actual (echo-received) token in $39\%$ of speaking episodes vs.\ only $5\%$ for the intended token (chance $= 1/|\mathcal{M}| = 14\%$).
In FORCE\_NOECHO, retx-actual drops to $0.020$ (below chance), confirming that the echo input directly drives repair content.
In CORRUPT (where intended $=$ actual), retx-intended and retx-actual are mathematically identical ($0.053$) by construction (see footnote above), providing a consistency check.

\paragraph{Two-input comparator model.}
The combined evidence characterizes P3 as a two-input comparator.
The hidden state provides an intention reference.
The B2 trigger result (FORCE $=$ CORRUPT) rules out a comparison between the actual transmission and the echo, and B1 adds that corruption status is decodable from $\mathbf{h}_t$ and generalizes across corruption types, without establishing independence from first-order encoding.
The echo channel provides the comparison signal (B2-FORCE\_NOECHO: 8/10 seeds fail to trigger without echo) and the repair content (B2-retx: echo-driven retransmission).
The detection mechanism thus requires both inputs.
Intention without echo fails to trigger (FORCE\_NOECHO $\to 0$), and echo without intention specificity would not produce the FORCE $\approx$ CORRUPT equivalence.
The intention-referenced mismatch signal gates corrective action, while the echo channel specifies its content.
\subsection{Feedback Conditions: Full Results}
\label{app:topology-results}

This appendix reports the complete results behind Table~\ref{tab:topology}, following the evaluation protocol of Appendix~\ref{app:evaluator}.
Table~\ref{tab:topology-perseed} lists the per-seed values, and the paragraphs below give the seed-level means, confidence intervals, and secondary analyses for each condition.

\begin{table}[h]
  \centering
  \caption{Per-seed results under the evaluation protocol ($\varepsilon = 0.25$, $t_c = 2$). Each row averages the two directions of one seed. Values $\approx 0$ denote $|x| < 10^{-6}$.}
  \label{tab:topology-perseed}
  \small
  \resizebox{\linewidth}{!}{
  \begin{tabular}{@{}l c c c c c c c@{}}
    \toprule
    Condition & Seed & Partner-state benefit & Self trigger & Partner response & Source contrast & Both-speak self & Repair value \\
    \midrule
    With-echo & 0 & $+0.5279$ & $+0.3698$ & $+0.0005$ & $+0.3692$ & $+0.6672$ & $+0.1464$ \\
    With-echo & 1 & $+0.5582$ & $+0.2041$ & $-0.0002$ & $+0.2044$ & $+0.5103$ & $+0.0746$ \\
    With-echo & 2 & $+0.5484$ & $+0.2168$ & $+0.0025$ & $+0.2143$ & $+0.1813$ & $+0.0951$ \\
    With-echo & 3 & $+0.5696$ & $+0.2625$ & $-0.0000$ & $+0.2625$ & $+0.3704$ & $+0.0932$ \\
    With-echo & 4 & $+0.5299$ & $+0.2283$ & $-0.0104$ & $+0.2386$ & $+0.2777$ & $+0.1289$ \\
    With-echo & 5 & $+0.5590$ & $+0.1707$ & $+0.0008$ & $+0.1699$ & $+0.3081$ & $+0.0723$ \\
    With-echo & 6 & $+0.5455$ & $+0.2330$ & $-0.0002$ & $+0.2332$ & $+0.4670$ & $+0.1362$ \\
    With-echo & 7 & $+0.5454$ & $+0.1326$ & $+0.0002$ & $+0.1323$ & $+0.1570$ & $+0.1140$ \\
    With-echo & 8 & $+0.5341$ & $+0.3700$ & $+0.0006$ & $+0.3694$ & $+0.3144$ & $+0.1570$ \\
    With-echo & 9 & $+0.5010$ & $+0.1184$ & $+0.0004$ & $+0.1181$ & $+0.1699$ & $+0.0915$ \\
    \midrule
    Yoked & 0 & $+0.5213$ & $+2.1\times10^{-7}$ & $+0.000008$ & $-0.000008$ & $+6.6\times10^{-7}$ & $+0.0880$ \\
    Yoked & 1 & $+0.5094$ & $-1.5\times10^{-6}$ & $+0.000022$ & $-0.000023$ & $-4.3\times10^{-7}$ & $+0.0432$ \\
    Yoked & 2 & $+0.5401$ & $\approx 0$ & $\approx 0$ & $\approx 0$ & $\approx 0$ & $+0.0619$ \\
    Yoked & 3 & $+0.5416$ & $\approx 0$ & $\approx 0$ & $\approx 0$ & $\approx 0$ & $+0.0944$ \\
    Yoked & 4 & $+0.5026$ & $\approx 0$ & $+2.3\times10^{-7}$ & $-2.5\times10^{-7}$ & $\approx 0$ & $+0.0608$ \\
    Yoked & 5 & $+0.5164$ & $-3.4\times10^{-6}$ & $+0.003256$ & $-0.003259$ & $-3.3\times10^{-6}$ & $+0.0483$ \\
    Yoked & 6 & $+0.5597$ & $\approx 0$ & $\approx 0$ & $\approx 0$ & $\approx 0$ & $+0.0825$ \\
    Yoked & 7 & $+0.4862$ & $-6.3\times10^{-7}$ & $+0.006601$ & $-0.006602$ & $+2.4\times10^{-6}$ & $+0.0482$ \\
    Yoked & 8 & $+0.5368$ & $+0.000022$ & $-0.000614$ & $+0.000636$ & $-0.000010$ & $+0.0290$ \\
    Yoked & 9 & $+0.5359$ & $\approx 0$ & $-0.000006$ & $+0.000006$ & $+3.4\times10^{-7}$ & $+0.0536$ \\
    \midrule
    Broadcast & 0 & $+0.5128$ & $+0.2039$ & $+0.1097$ & $+0.0943$ & $+0.0611$ & $+0.1155$ \\
    Broadcast & 1 & $+0.5204$ & $+0.1792$ & $+0.0194$ & $+0.1599$ & $+0.1305$ & $+0.1146$ \\
    Broadcast & 2 & $+0.5197$ & $+0.2301$ & $+0.0879$ & $+0.1422$ & $+0.1731$ & $+0.1186$ \\
    Broadcast & 3 & $+0.5411$ & $+0.2682$ & $+0.0857$ & $+0.1825$ & $+0.3887$ & $+0.1000$ \\
    Broadcast & 4 & $+0.5164$ & $+0.2942$ & $+0.0490$ & $+0.2452$ & $+0.3927$ & $+0.1104$ \\
    \bottomrule
  \end{tabular}
  }
\end{table}

\paragraph{Base task in all conditions.}
Yoked and broadcast agents learn the task and the protocol.
Under the training channel ($\varepsilon = 0.25$), partner-state accuracy with communication is $0.859$ (yoked, 10 seeds) and $0.857$ (broadcast, 5 seeds), against $0.334$ and $0.335$ without communication.
The AND accuracy is $0.847$ and $0.851$, against $0.324$ and $0.329$.
Both agents are jointly correct in $0.761$ of steps in the yoked condition, against $0.106$ without communication.
On a clean channel, partner-state accuracy is $0.904$ (yoked) and $0.904$ (broadcast).
The communication benefit is positive in every seed.

\paragraph{Yoked control.}
The seed-level means with 95\% CIs are as follows.
The self trigger is $+1.68 \times 10^{-6}$ $[-3.54, +6.90] \times 10^{-6}$ (3/10 positive in sign).
The partner response is $+0.000927$ $[-0.000690, +0.002543]$, the source contrast is $-0.000925$ $[-0.002543, +0.000693]$, and the global sender-minus-receiver response is $-0.000805$ $[-0.002590, +0.000980]$.
The both-speak self trigger is $-1.03 \times 10^{-6}$, and the same-token retransmission-probability effect is $-2.93 \times 10^{-6}$.
The ratio of the yoked to the with-echo self trigger is $1.68 \times 10^{-6} / 0.2306 = 7.28 \times 10^{-6}$, far below the 20\% threshold of the decision rule (Appendix~\ref{app:evaluator}).
On a clean channel ($\varepsilon = 0$), the self trigger is $-4.31 \times 10^{-7}$ and the source contrast is $-0.000868$.
At the secondary times, the self trigger is $+2.52 \times 10^{-7}$ ($t_c = 1$) and $-0.000027$ ($t_c = 3$), with source contrasts $-0.000257$ and $-0.002916$.
For the 9 seeds that passed the phase-1 gate, the self trigger is $+1.87 \times 10^{-6}$ and the source contrast is $-0.001028$.
The training logs contain a coarser diagnostic that corrupts the sender's actual transmission and gives a trigger of roughly $+0.0025$.
This diagnostic also changes the partner's input and, through the batch shift, the yoked line of another episode, so it does not isolate the self-feedback observation and is not used as the yoked result.

\paragraph{Hidden-state response.}
The table below shows the $L_2$ change of the focal hidden state between MATCH and MISMATCH for self-copy and partner-copy interventions.
The hidden states of yoked agents respond to a change in the partner message with the same magnitude as in the other conditions.
They respond much less to an isolated change of the yoked feedback slot ($2.8\%$ of the partner-copy change, and $5.1\%$ of the self-copy change in the with-echo condition).
The policy thus largely ignores a line that has no predictive value for the current episode.
\begin{center}
  \small
  \begin{tabular}{@{}lcc@{}}
    \toprule
    Condition & Self-copy hidden $L_2$ & Partner-copy hidden $L_2$ \\
    \midrule
    With-echo & $1.7705$ & $2.8939$ \\
    No-echo & $0.4996$ & $3.2058$ \\
    Yoked & $0.0895$ & $3.1864$ \\
    Broadcast & $2.7509$ & $3.3147$ \\
    \bottomrule
  \end{tabular}
\end{center}

\paragraph{Randomized broadcast.}
The seed-level means with 95\% CIs are as follows, and all are positive in 5/5 seeds.
The self trigger is $+0.2351$ $[0.1772, 0.2931]$, the partner response is $+0.0703$ $[0.0258, 0.1149]$, and the source contrast is $+0.1648$ $[0.0960, 0.2336]$.
The global sender-minus-receiver response is $+0.1650$ $[0.0952, 0.2349]$.
In the both-speak subset, the self trigger is $+0.2292$ $[0.0396, 0.4189]$ and the source contrast is $+0.1943$ $[0.0124, 0.3762]$.
The same-token retransmission-probability effect is $+0.0749$ $[-0.0406, 0.1904]$ (3/5 positive).
The response is thus an increased tendency to speak and not a fixed re-emission of the previous token.
The both-speak subset contains $38.9\%$ of the episodes.
At the direction level, all ten self triggers are positive, and one direction (seed 0, A$\to$B) has a negative both-speak response while the seed average remains positive.
At the secondary times, the self trigger and source contrast are $+0.0610$ and $+0.0385$ at $t_c = 1$ (CIs include zero) and $+0.1331$ and $+0.0813$ at $t_c = 3$ (5/5 positive).
The response thus depends on the intervention time.
For the 4 seeds that passed the phase-1 gate, the self trigger is $+0.2204$ and the source contrast is $+0.1447$ (4/4 positive).

\paragraph{Scope.}
The yoked result is a bounded estimate and not a proof of an exact null, and it does not rule out some representation of the yoked input.
The broadcast result does not establish compositional identity encoding, strict permutation invariance, or scaling beyond two agents.
The repair value is also positive in the yoked and no-echo conditions ($+0.0610$ and $+0.0581$).
This reflects the general value of subsequent communication and is not evidence of self-monitoring in those conditions.

\subsection{Repair-Value Counterfactual: Full Results}
\label{app:repair-results}

Tables~\ref{tab:repair-full} and \ref{tab:repair-perseed} give the complete results of the counterfactual of Appendix~\ref{app:repair-spec} at the three intervention times.
Positive values mean that the natural post-mismatch message improves the outcome relative to the receiver-side control.

\begin{table}[h]
  \centering
  \caption{Repair value of the natural post-mismatch message (10 with-echo seeds, clean channel apart from the forced mismatch). ``Window'' scores from $t_c{+}2$ to the episode end, and ``first'' scores the first affected step only.}
  \label{tab:repair-full}
  \small
  \resizebox{\linewidth}{!}{
  \begin{tabular}{@{}c l l c c c c@{}}
    \toprule
    $t_c$ & Outcome & Control & Repair value & 95\% CI & $p$ & Positive seeds \\
    \midrule
    1 & Receiver partner-state acc.\ (window) & drop & $+0.1964$ & $[0.1149, 0.2780]$ & $4.1\times10^{-4}$ & 9/10 \\
    1 & Receiver partner-state acc.\ (window) & yoked & $+0.2502$ & $[0.1939, 0.3066]$ & $3.5\times10^{-6}$ & 10/10 \\
    1 & Team AND reward (window) & yoked & $+0.1250$ & $[0.0967, 0.1533]$ & $3.6\times10^{-6}$ & 10/10 \\
    1 & Receiver partner-state acc.\ (first) & yoked & $+0.3024$ & $[0.2567, 0.3481]$ & $1.2\times10^{-7}$ & 10/10 \\
    \midrule
    2 & Receiver partner-state acc.\ (window) & drop & $+0.0754$ & $[0.0250, 0.1258]$ & $0.0081$ & 10/10 \\
    2 & Receiver partner-state acc.\ (window) & yoked & $+0.0790$ & $[0.0292, 0.1288]$ & $0.0059$ & 10/10 \\
    2 & Team AND reward (window) & drop & $+0.0377$ & $[0.0125, 0.0629]$ & $0.0081$ & 10/10 \\
    2 & Team AND reward (window) & yoked & $+0.0395$ & $[0.0146, 0.0644]$ & $0.0059$ & 10/10 \\
    2 & Receiver partner-state acc.\ (first) & yoked & $+0.0542$ & $[0.0225, 0.0858]$ & $0.0038$ & 10/10 \\
    \midrule
    3 & Receiver partner-state acc.\ (window) & drop & $+0.0118$ & $[-0.0069, 0.0304]$ & $0.19$ & 6/10 \\
    3 & Receiver partner-state acc.\ (window) & yoked & $+0.0107$ & $[-0.0046, 0.0260]$ & $0.15$ & 6/10 \\
    3 & Team AND reward (window) & yoked & $+0.0053$ & $[-0.0023, 0.0130]$ & $0.15$ & 6/10 \\
    \bottomrule
  \end{tabular}
  }
\end{table}

\begin{table}[h]
  \centering
  \caption{Per-seed repair value (receiver partner-state accuracy, window, yoked control, clean channel apart from the forced mismatch) and the sender's forced-mismatch trigger (sampled speak rate at $t_c{+}1$, mismatch minus clean) at $t_c = 2$. Across seeds, trigger and repair value correlate at $r = 0.886$ ($p = 0.0006$; Spearman $\rho = 0.830$, $p = 0.003$).}
  \label{tab:repair-perseed}
  \small
  \resizebox{\linewidth}{!}{
  \begin{tabular}{@{}l cccccccccc@{}}
    \toprule
    Seed & 0 & 1 & 2 & 3 & 4 & 5 & 6 & 7 & 8 & 9 \\
    \midrule
    Trigger ($t_c = 2$) & $0.406$ & $0.058$ & $0.170$ & $0.113$ & $0.221$ & $0.063$ & $0.098$ & $0.110$ & $0.340$ & $0.040$ \\
    Repair value, yoked & $0.1887$ & $0.0035$ & $0.0842$ & $0.0063$ & $0.1650$ & $0.0037$ & $0.0512$ & $0.0918$ & $0.1518$ & $0.0435$ \\
    Repair value, drop & $0.2035$ & $0.0001$ & $0.1288$ & $0.0005$ & $0.1418$ & $0.0002$ & $0.0399$ & $0.0953$ & $0.1068$ & $0.0375$ \\
    \bottomrule
  \end{tabular}
  }
\end{table}

\paragraph{Interpretation.}
This test estimates the causal value of the learned post-mismatch message.
Self-reference is supported by other results: the sender/receiver, echo-only/receive-only, lag, and intention-reference dissociations, and the feedback conditions.
The mean trigger across the three intervention times ($+0.3603$, $+0.1618$, $+0.0263$) and the corresponding repair values ($+0.2502$, $+0.0790$, $+0.0107$) decline together, which supports the sensitivity analysis in \S\ref{subsec:repair-value}.
The trigger here is the sampled speak rate at $t_c{+}1$ (mismatch minus clean), averaged over seeds with $5{,}000$ episodes per direction.
The same intervention in Appendix~\ref{app:secondorder-b2}, which uses $20{,}000$ episodes per seed, gives $+0.141$ at $t_c = 2$.

\subsection{State-Change Variant: Full Results}
\label{app:statechange-results}

This appendix reports the full results of the variant specified in Appendix~\ref{app:statechange-spec}.
Intervals are two-sided 95\% $t$ intervals over the five seeds.
Where a parameter has logical bounds (e.g., probabilities), the displayed interval is intersected with those bounds.

\paragraph{Four-cell speaking response.}
The table below gives the sender's speaking probability at $t_{\Delta}$ in the four cells of the design, on a clean channel and at the training noise level, together with the state-change effect and the interaction.
\begin{center}
  \small
  \resizebox{\linewidth}{!}{
  \begin{tabular}{@{}lcc@{}}
    \toprule
    Condition & Clean $\varepsilon = 0$ & Training noise $\varepsilon = 0.25$ \\
    \midrule
    Stable $\times$ Match & $0.383$ $[0.293, 0.474]$ & $0.412$ $[0.340, 0.483]$ \\
    Stable $\times$ Mismatch & $0.535$ $[0.429, 0.642]$ & $0.592$ $[0.501, 0.684]$ \\
    Changed $\times$ Match & $0.907$ $[0.786, 1.000]$ & $0.911$ $[0.792, 1.000]$ \\
    Changed $\times$ Mismatch & $0.914$ $[0.782, 1.000]$ & $0.916$ $[0.784, 1.000]$ \\
    \midrule
    State-change effect (Changed$\times$Match $-$ Stable$\times$Match) & $+0.523$ $[0.385, 0.662]$ & $+0.499$ $[0.376, 0.623]$ \\
    Mismatch $\times$ change interaction & $-0.145$ $[-0.227, -0.063]$ & $-0.176$ $[-0.261, -0.091]$ \\
    \bottomrule
  \end{tabular}
  }
\end{center}
The interaction is negative in 5/5 seeds.
After a state change, a mismatch on the old transmission adds little speaking beyond the increase already caused by the change.
This interaction is limited by a ceiling.
Seeds 1 and 4 are saturated in the Changed$\times$Match cell (headroom $< 10^{-3}$).
A post-hoc association between headroom and the absolute interaction magnitude has the direction predicted by a ceiling on the clean channel (Spearman $\rho = -0.80$, exact two-sided $p = 0.13$, $n = 5$; $\rho = -0.40$, $p = 0.52$ at $\varepsilon = 0.25$).
We thus report the interaction only descriptively, and the interpretation rests on the content and receiver-side results below.

\paragraph{Content of the post-mismatch update.}
The readable subset (old intention, returned token, and current-policy reference all non-SILENCE and pairwise distinct) covers $10{,}854$ of $50{,}000$ clean-channel observations ($21.7\%$; $22.8\%$ at $\varepsilon = 0.25$).
Four seeds have readable observations in both directions.
The A$\to$B direction of seed~3 has none, so the direction-averaged content contrast of this seed is undefined, and the seed is excluded from the content contrasts only.
\begin{center}
  \small
  \begin{tabular}{@{}lcc@{}}
    \toprule
    Measure (readable subset) & Clean & $\varepsilon = 0.25$ \\
    \midrule
    Mean $p(\text{old intention})$ & $0.0000$ & $0.0001$ \\
    Mean $p(\text{returned token})$ & $0.0008$ & $0.0013$ \\
    Mean $p(\text{current-policy reference})$ & $0.9992$ & $0.9972$ \\
    Reference $-$ old (4 estimable seeds) & $+0.9992$ $[0.9972, 1.000]$ & $+0.9971$ $[0.9910, 1.000]$ \\
    Reference $-$ returned token (4 estimable seeds) & $+0.9985$ $[0.9945, 1.000]$ & $+0.9959$ $[0.9881, 1.000]$ \\
    \bottomrule
  \end{tabular}
\end{center}
Both contrasts are positive in 4/4 estimable seeds under both channel conditions.
The reference is sampled from the same sender in the paired changed$\times$match branch.
The contrasts thus show that a mismatch induces neither a replay of the old intention nor a copy of the returned token, and that the sender's choice under mismatch matches its choice without mismatch.
They do not test whether that choice is correct.
Large values are expected if the state-conditioned message head is nearly deterministic and the echo has little effect on it.

\paragraph{Receiver-side revision.}
The table below reports how far the receiver's action readout moves from the old state toward the new state at lags 1 and 2 after the update.
\begin{center}
  \small
  \begin{tabular}{@{}lcc@{}}
    \toprule
    Measure & Clean & $\varepsilon = 0.25$ \\
    \midrule
    New $-$ old action readout, lag 1 & $-0.065$ $[-0.260, +0.130]$ & $-0.240$ $[-0.397, -0.083]$ \\
    New $-$ old action readout, lag 2 & $+0.867$ $[+0.735, +0.999]$ & $+0.590$ $[+0.476, +0.705]$ \\
    Fraction revised by lag 2 & $0.931$ $[0.852, 1.000]$ & $0.793$ $[0.722, 0.864]$ \\
    \bottomrule
  \end{tabular}
\end{center}
Lag-2 revision is positive in 5/5 seeds under both channel conditions.
The lag-1 value reflects the one-step information delay at the change step, where the receiver acts on the old message before the update arrives.
The readout is measured on the receiver's actions and not with a hidden-state decoder.

\paragraph{Downstream causal value of the update.}
The table below compares the receiver's partner-state accuracy when the natural update is delivered and when it is replaced by one of four alternatives.
The last two rows give the changed-versus-stable difference-in-differences.
\begin{center}
  \small
  \resizebox{\linewidth}{!}{
  \begin{tabular}{@{}lcc@{}}
    \toprule
    Contrast (receiver partner-state accuracy) & Clean & $\varepsilon = 0.25$ \\
    \midrule
    Natural update $-$ SILENCE & $+0.387$ $[0.256, 0.519]$ & $+0.353$ $[0.227, 0.480]$ \\
    Natural update $-$ yoked token & $+0.251$ $[0.198, 0.303]$ & $+0.229$ $[0.182, 0.276]$ \\
    Natural update $-$ old intention & $+0.137$ $[0.011, 0.263]$ & $+0.102$ $[-0.022, 0.225]$ \\
    Natural update $-$ returned erroneous token & $+0.410$ $[0.350, 0.471]$ & $+0.381$ $[0.307, 0.454]$ \\
    \midrule
    Changed-vs-stable difference-in-differences, SILENCE control & $+0.272$ $[-0.048, 0.592]$ & $+0.205$ $[-0.087, 0.497]$ \\
    Changed-vs-stable difference-in-differences, yoked control & $+0.051$ $[-0.121, 0.222]$ & $-0.003$ $[-0.153, 0.147]$ \\
    \bottomrule
  \end{tabular}
  }
\end{center}
The contrasts against SILENCE, a yoked token, and the returned token are positive under both channel conditions.
The contrast against the old intention is positive on the clean channel, and its interval includes zero on the noisy channel.
In seeds 1--4, the old-intention contrast is positive in 3/4 seeds and the returned-token contrast in 4/4 seeds under both conditions.
The changed-versus-stable differences-in-differences are inconclusive.
The data thus do not establish that the update has greater incremental value than ordinary communication in the stable condition.

\paragraph{Time-only alternative.}
At $\varepsilon = 0.25$, speaking after a recent unsignalled change exceeds speaking after a recent valid history by $+0.389$ $[0.312, 0.467]$ (5/5 seeds).
A long valid history reduces speaking relative to a recent valid history by $-0.468$ $[-0.529, -0.406]$ (5/5 seeds).
A long unsignalled change raises speaking relative to a long valid history by $+0.627$ in the three seeds with estimable cells.
Speaking is thus not explained by a policy that becomes more likely to speak as time elapses, because the state change raises speaking within a fixed recent-history stratum.
This result does not address whether the sender represents the receiver's belief.

\paragraph{Stale-message dependence in the stationary environment.}
Old messages also have a task cost in the stationary environment of the ten with-echo seeds.
The mean completed receipt-to-receipt gap for non-SILENCE tokens is $1.97$ steps $[1.87, 2.07]$, with at least two intervening decision steps in $24.7\%$ of cases.
Replacing a received token with a yoked token reduces the receiver's correct-action probability by $0.45$ $[0.39, 0.52]$ two decisions later when no newer message has arrived.
Old messages thus continue to drive receiver behavior across the interval in which a state change would make them obsolete.

\paragraph{Summary.}
The variant gives five results.
\begin{compactenum}[(1)]
  \item After a state change, the marginal mismatch response is attenuated in every seed, although this attenuation is confounded with the speaking ceiling.
  \item In 4/4 estimable seeds, a mismatch induces neither a replay of the old intention nor a copy of the returned token, and the sender follows its current-policy reference.
  \item The receiver revises its action readout in 5/5 seeds.
  \item The update outperforms SILENCE, a yoked token, and the returned token in receiver value, while the changed-versus-stable difference-in-differences is inconclusive.
  \item The content, revision, and receiver-value patterns persist at the training corruption rate.
\end{compactenum}
These results reject the reading of P3 as a fixed reflex that detects a mismatch and copies the returned token.
Because speaking is near ceiling in some seeds, the evidence against the broader description ``detect a mismatch and speak'' is weaker.
The state-change pressure and the feedback channel remain part of the task, so the result is learned adaptive control that depends on temporal context.
The variant does not test whether the retained state persists across an interval with no task demand.

\section{Extended Discussion}
\label{app:discussion}
\subsection{Why Communication, Not Just Probing?}
\label{app:why-communication}

A natural objection is that the structural properties we study (self-state encoding, temporal persistence, mismatch detection) could be investigated by probing internal representations directly, without requiring a communication channel.
Three considerations motivate the EL framework over pure internal probing.

\paragraph{An externally observable, manipulable interface.}
Communication provides an output that is both externally observable and causally manipulable.
Probing hidden states requires the researcher to decide what to probe for. The choice of probe target already encodes assumptions about which representations matter.
An emergent communication channel reverses this dependency: the agent decides what information to transmit, and the researcher observes the result.
This is why P1 is informative: the finding that messages encode $s_{\text{self}}$ rather than $s_{\text{other}}$ is a discovery about the agent's communicative priorities, not an artifact of the experimenter's probe design.

\paragraph{P3 depends on communication, not architecture.}
The core finding, P3, relies on the echo channel as an \emph{environmental affordance}: a feedback copy of the agent's own communicative output that may be corrupted by the channel.
This affordance is external to the agent's architecture. It is a property of the communication environment.
The key evidence (sender-receiver dissociation, echo causal split, no-echo ablation) all exploit the fact that the echo is a manipulable environmental variable, not an internal architectural component.
A pure probing approach would have no analogue: there is no ``echo of a hidden state'' to corrupt and observe behavioral responses to.
The distinction matters because it grounds P3 in the agent-environment interface rather than in researcher-imposed interventions on internal states.

\paragraph{Grounding in interaction.}
The symbols agents produce are grounded in coordinated action: they carry information because they enable task success, not because they match a researcher-specified labeling scheme.
This grounding ensures that the structures we detect are functional (they contribute to coordination) rather than epiphenomenal (statistically present in hidden states but causally inert).
The communication necessity criterion ($\Delta_{\text{comm}} \gg 0$) makes this explicit: removing the channel degrades performance, confirming that the encoded information is used by the receiver.

\subsection{The Prior-Leakage Problem in Detail}
\label{app:prior-leakage}

\citet{li2024langground} propose LangGround, which aligns MARL agent communication vectors with LLM-generated reference communications via a cosine-similarity loss $L_{\text{sup}} = \sum_{t,i} [1 - \cos(\mathbf{c}_i^t, D(o_i^t, a_i^t))]$, where $D$ maps observation--action pairs to LLM-generated natural language embeddings.
This supervised alignment directly transfers human linguistic structures, including the first-person pronoun ``I'' and its associated referential patterns, into the agent's communication space.
Any indexical structure observed in such a system cannot be attributed to emergent self-reference; it may be inherited from the human language prior embedded in the LLM.

The prior-leakage problem extends beyond explicit alignment losses.
Any approach that uses LLM outputs as training signal, reward shaping, or communication initialization risks importing human linguistic priors implicitly.
An LLM agent that produces ``I see the target'' is not demonstrating emergent self-reference. It is performing next-token prediction over a distribution trained on human text in which ``I'' is the most common subject pronoun.
This is the philosophical zombie problem in computational form: the system produces first-person language not because it has developed a self-referential structure, but because its training distribution contains such language as a statistical regularity.
The distinction between emergent self-reference and inherited self-reference is precisely the distinction between spontaneous behavior and next-token prediction over a human-language prior.

\subsection{Positioning LLM Priors in Our Methodology}
\label{app:llm-environment}

The long-term motivation for this methodology can be traced to the simulated-civilization agenda sketched in our previous work on LLM-based simulation \citep{wu2023smart}, where multimodal agents with rudimentary communicative abilities were envisioned as interacting in rich environments, forming communities, developing shared practices, and gradually inventing or enriching their language. A central question was whether such agents could independently develop consciousness-relevant vocabulary without those terms being supplied by human designers. This question was motivated in part by \citet{moody1994conversations}'s zombie civilization argument and by Gadamer's thesis that language is fundamental to our being-in-the-world \citep{gadamer1977philosophical}.

Our methodology differs from LLM-based simulation in two respects: the agent and the interaction medium. First, regarding the agent, LLM-based agents carry a rich human language prior that dominates their behavior, so any observed self-referential structure may be inherited from the training corpus rather than emergent from the current task. By contrast, the agents in our methodology are prior-minimal, ensuring causal attributability to task pressure. Second, regarding the interaction medium, LLM-based agents interact through natural language, which itself carries accumulated conceptual structure, whereas our agents interact through a designed discrete channel. The causal evidence for P3 rests on echo corruption, echo silencing, and channel ablation, interventions with no counterpart in natural-language interaction.

This contrast raises a natural question: how does our methodology relate to the vast body of work using LLMs as agents in complex environments?
We distinguish two roles that LLMs can play, with different implications for the study of consciousness-relevant structures.

\paragraph{LLM as an interpreter.}
At intermediate and large scales of environment complexity (\S\ref{subsec:interpretation}), LLMs can serve as interpreters of emergent protocols: given examples of agent behavior and environment states, an LLM can describe the structure of the protocol in human-readable terms.
In this role, the LLM does not participate in the language game; it translates between the emergent protocol and human language after the fact.
This use is methodologically unproblematic, because the emergent structure itself remains causally attributable to task demands rather than to the LLM's priors.

\paragraph{LLM as an agent.}
When an LLM is used as the agent itself, producing actions and communication in a multi-agent environment, its outputs are shaped by the statistical regularities of human language, including first-person reference, mental-state attribution, and narrative perspective.
Any consciousness-relevant structure observed in such a system faces the prior-leakage problem. 
This does not mean LLM agent approaches are uninformative (they may reveal which human-language structures are robust under novel task pressure), but it does mean that the causal provenance of observed structure is ambiguous.

\paragraph{Towards convergence.}
As we mentioned in \S\ref{subsec:method-limits}, an important open question is whether the structures that emerge via prior-minimal design and those that emerge via next-token prediction on human text can be shown to converge to equivalent functional organizations.
If the consciousness-relevant structures identified by our methodology (P1--P3) also appear, in functionally equivalent form, in LLMs deployed in similar multi-agent settings, this would provide evidence that the structures are driven by task demands rather than by substrate or training regime.
Conversely, if the structures diverge, this would illuminate which aspects the structures are universal (driven by the information-theoretic structure of communication under constraints) and which are specific to the training regime.
This convergence question is one of the most important open problems for the field, as it would provide a bridge between the addition and subtraction approaches described in Appendix \ref{app:ai-debate}.



\subsection{The Comparative Cognition Continuum}
\label{app:continuum}

The comparative cognition literature documents a range of self-referential capacities that vary with organism complexity and ecological demand.
\citet{gallup1970chimpanzees} demonstrated mirror self-recognition in chimpanzees, the capacity to recognize one's own body as distinct from others', which has since served as a canonical marker of self-other distinction.
\citet{tulving1985memory} introduced the concept of autonoetic consciousness: the capacity to mentally travel in time and re-experience one's own past, grounded in episodic memory that maintains a temporally extended self-representation.
\citet{hampton2001rhesus} showed that rhesus monkeys can monitor the strength of their own memories, declining to answer when uncertain, a form of metacognitive monitoring that goes beyond first-order performance.
\citet{smith2003comparative} extended this to a broader comparative framework, documenting uncertainty monitoring across species and arguing that metacognitive capacity scales with the complexity of the organism's decision environment.
\citet{premack1978chimpanzee} framed theory of mind as the capacity to attribute mental states to others, a social-cognitive capacity that presupposes the self-other distinction.
\citet{dennett1991consciousness} proposed that the human self is a ``center of narrative gravity'': not a metaphysical entity but a useful abstraction generated by the brain's narrative-constructing processes.

Our three preconditions map onto the lower end of this continuum:
P1 provides the self-other distinction that Gallup Jr's mirror test assesses.
P2 is a minimal form of temporal self-continuity, a precursor to the autonoetic consciousness Tulving describes.
P3 instantiates the simplest form of the metacognitive monitoring documented by \citet{hampton2001rhesus} and \citet{smith2003comparative}: not confidence about memories, but detection of one's own communicative errors and functionally beneficial adjustment of subsequent output.

\subsection{Wittgenstein and Public Language Games}
\label{app:wittgenstein}

\citet{wittgenstein1953philosophical} argued that meaning is constituted by use within a language game: a pattern of rule-governed activity embedded in a form of life.
Two aspects of this argument bear directly on our results.

\paragraph{Private reference, public criteria.}
The private language argument (\S\S243--315) contends that a language whose terms refer exclusively to the speaker's private sensations is incoherent, because there is no criterion of correctness independent of the speaker's impression of correctness.
Our agents' tokens are superficially ``private'' in the sense that they encode the sender's own state (P1), yet they are publicly verifiable: the receiver's action accuracy provides an external criterion that disciplines the mapping between token and state.
This is precisely Wittgenstein's point: what makes a sign meaningful is not its association with an inner state per se, but the existence of a practice within which correct and incorrect uses can be distinguished.
Task reward supplies this practice.
The beetle-in-the-box analogy (\S\S293) reinforces the point: even if each agent's internal representation is ``private,'' the behavioral consequences (consistent, decodable communication) are public and predictable.

\paragraph{Language games without pre-existing language.}
Wittgenstein's examples presuppose an existing linguistic community: the builder's language game (\S\S2) begins with ``slab,'' ``beam,'' and so on.
Our setup removes this presupposition entirely.
Agents begin with no tokens, no conventions, and no shared history.
The language game is not inherited but constituted through interaction under task pressure.
This provides an empirical demonstration of how the rule-governed regularities Wittgenstein describes can arise from scratch: the agents' emergent protocol satisfies the criteria for a language game (regular use, public criteria of correctness, embedded in coordinated activity) without any prior linguistic endowment.

\paragraph{Rule-following and training.}
Wittgenstein's discussion of rule-following (\S\S185--242) raises the question of what constitutes following a rule rather than merely behaving in accord with one.
Our agents' behavior is consistent with a rule (``encode $s_{\text{self}}$ using token $k$''), but this rule was never stated or represented as such.
Whether the agents ``follow'' the rule or merely ``accord with'' it depends on one's philosophical commitments; the structural facts are the same either way.
What our results show is that the behavioral regularity that Wittgenstein identifies as constitutive of meaning (consistent, publicly checkable use embedded in a coordinated practice) can emerge from RL without presupposing the rule.

\subsection{Compositional Self-Reference}
\label{app:compositional}

Our agents develop indexical encoding: each message $m$ carries information about the sender's state, with $I(m; s_{\text{self}}) \gg I(m; s_{\text{other}})$.
However, the current protocol is \emph{holistic}: each token maps to a single state value as an unanalyzed unit.
The receiver can decode this mapping because it knows who sent the message, since in the two-agent case there is only one partner.
This subsection discusses the conditions under which a \emph{compositional} self-referential protocol would emerge, where messages decompose into independently meaningful components analogous to the pronoun--predicate structure of natural language (``I'' + ``am in state $k$'').

\paragraph{Why compositionality matters.}
Natural language self-reference is compositional: ``I am hungry'' separates the self-referential indexical (``I'') from the predicated content (``am hungry''), and each component generalizes independently: a new listener understands ``I'' without having to learn a new holistic mapping for each speaker.
In the emergent communication literature, compositionality is the structural property that enables systematic generalization: the ability to interpret novel combinations of familiar components \citep{kottur2017natural,chaabouni2020compositionality}.
Applied to self-reference, a compositional protocol would allow a receiver to decode a message from a previously unseen sender, because the sender-identity component and the state-content component are structurally separable.

\paragraph{When compositionality is unnecessary in our experiments.}
In our current setting, compositionality is not required and does not emerge, for a precise reason.
Each agent receives messages from a single known partner.
Because the sender's identity is unambiguous from the channel structure, the receiver can learn a holistic token--state mapping specific to that partner.
There is no pressure to encode sender identity within the message, because it is already available outside the message.
Formally, let $\text{id}_j$ denote the identity of the sender.
When $H(\text{id}_j \mid \text{channel position}) = 0$ (i.e., the receiver always knows who is speaking), a holistic protocol $m \mapsto s_j$ suffices and is more bandwidth-efficient than a compositional one.

\paragraph{Channel-position disambiguation: a simplifying assumption, not a self-prior.}
In our current setting, the receiver knows who sent each message because channel positions are fixed (in the two-agent case, there is only one possible sender).
This provides the receiver with free \emph{other-identification}, but importantly, it does not provide the sender with any prior about \emph{self}.
P1 is a property of the sender's encoding: the sender encodes $s_{\text{self}}$ rather than $s_{\text{other}}$ because, under bandwidth constraints, the locally observed state is the most informative signal available.
This encoding choice is driven by the information-theoretic structure of partial observability (Appendix~\ref{app:structural-necessity}), not by any concept of self-identity.
The sender's privileged access to $s_{\text{self}}$ is a consequence of embodiment (any spatially situated agent observes its own local state), not a human linguistic prior.
Channel-position disambiguation is thus a simplifying assumption that makes the receiver's task easier, but it does not make the sender's indexical encoding trivial or prior-dependent.

\paragraph{Conditions for compositional pressure.}
Removing this simplifying assumption creates the pressure for compositional structures.
Consider a three-agent setting where each agent receives messages from two others on a shared broadcast channel with shuffled or unlabeled positions.
If the receiver cannot distinguish which incoming token was sent by which partner, it faces an identification problem: it must decode both \emph{who sent the message} and \emph{what state it encodes} from the token itself.
A holistic protocol collapses under this ambiguity, because the same token $m$ sent by agent $j$ (encoding $s_j$) and agent $k$ (encoding $s_k$) would be indistinguishable.

The minimal condition for compositional pressure is thus:
\begin{equation}
  H(\text{id}_{\text{sender}} \mid m_{\text{received}}, \text{channel structure}) > 0
  \label{eq:comp-condition}
\end{equation}
i.e., the receiver has residual uncertainty about sender identity after observing the message and the channel.
Under this condition, an optimal protocol must encode sender identity as a separable component of the message.
If the message space admits factorization $m = (m_{\text{who}}, m_{\text{what}})$, the compositional solution satisfies:
\begin{equation}
  I(m_{\text{who}}; \text{id}_{\text{sender}}) \gg 0, \quad I(m_{\text{what}}; s_{\text{sender}}) \gg 0, \quad I(m_{\text{who}}; s_{\text{sender}}) \approx 0
\end{equation}
where the last condition ensures that the identity component does not redundantly encode state content.
The $m_{\text{who}}$ component would function as an emergent \emph{self-word}: a token whose referent is the sender itself, analogous to the first-person pronoun.

\paragraph{Relation to P1--P3.}
Compositional self-reference is not a fourth precondition for first-person structure; it is an orthogonal property of the communication protocol.
P1 establishes that messages are indexical (dominated by sender state). Compositionality would further require that this indexical content is \emph{structurally decomposable}.
An agent can satisfy P1--P3 with a holistic protocol, as our results demonstrate.
Conversely, a compositional protocol could exist without P2 or P3, since compositionality concerns message structure, not temporal persistence or self-monitoring.
The interest of compositionality lies in generalizability: a compositional self-referential protocol would transfer across partners without retraining, a capacity our current agents lack (as the cross-pair control results confirm).

\paragraph{Towards future work.}
Testing whether compositional self-reference emerges under the conditions described above requires a multi-sender environment with sender-ambiguous channels, a setting we leave to future work.
The key experimental criterion is zero-shot cross-partner transfer: an agent trained with partners $j$ and $k$ should decode messages from a novel partner $l$ without additional training, provided $l$'s messages follow the same compositional structure.
This would establish that the self-word generalizes, paralleling the universality of ``I'' across speakers of a natural language.

\section{Limitations of Experimental Findings}
\label{app:limitations}
We discuss the limitations of the experimental findings in this paper.

\paragraph{Scale.}
Two agents, seven tokens, ten-step episodes.
This minimality strengthens the structural argument but limits direct claims about richer domains.

\paragraph{Prior-minimal design.}
Our agents learn from scratch, ensuring emergence, but this makes learning inefficient.

\paragraph{Simplified partial observability.}
The environment masks $s_i$ entirely after $t=0$, sharper than natural settings where self-state degrades gradually.

\paragraph{Self-as-object versus self-as-subject.}
In the classical distinction \citep{james1890principles,mead1934mind}, our agents demonstrate the self as known (\emph{me}), not the self as knower (\emph{I}).
Whether any system can demonstrate the latter is explicitly bracketed. 

\paragraph{Frozen weights.}
After training, parameters are frozen.
The richer capacities discussed in Appendix~\ref{app:continuum} (metacognitive confidence monitoring, theory of mind, narrative self-continuity) would plausibly require continual learning during deployment \citep{pan2025survey,malenfant2026reinforcing}.

\paragraph{Disembodied agents.}
In our environment, agents are signal receivers: they observe discrete state variables and produce discrete tokens.
In the physical world, cognition is deeply integrated with the body and the environment. Embodied cognition \citep{bisk2020experience} emphasizes that perception, action, and thought are not separable modules but aspects of a single organism--environment coupling.
Our agents lack sensorimotor grounding: they do not act on a physical environment, do not have bodies that constrain their actions, and do not experience the consequences of their actions through continuous sensory feedback.
This simplification means that the self-referential structures we observe are informational rather than embodied, and the degree to which embodied self-reference requires additional mechanisms beyond P1--P3 remains an open question.

\paragraph{Scope.}
We identify structural preconditions, not sufficient conditions for consciousness.
The framework is compatible with the possibility that these structures are necessary-but-not-sufficient for access consciousness and entirely silent on phenomenal consciousness \citep{hoel2026disproof,mcclelland2025agnosticism}.
\section{Broader Implications and Future Work}
\label{app:future-work}

\paragraph{Substrate independence.}
The three preconditions are substrate-independent functional scaffolds defined in terms of information flow, not biological mechanisms.
This is by design: our framework is a purely logical construction that makes no reference to neurons, neurotransmitters, or any specific physical medium.
The substrate independence of P1--P3 exposes a limitation of consciousness theories that are implicitly or explicitly tied to biological hardware: if a functional structure identical to P1--P3 can emerge on silicon, but the theory requires carbon, then the theory's scope is narrower than it claims \citep[cf.][who defend substrate independence for digital minds]{bostrom2025propositions}.
The question of whether the functional scaffold ``carries'' phenomenal experience on a new substrate is precisely the hard problem. Our framework identifies what transfers (functional structure) and what remains unknown (experiential accompaniment).

\paragraph{AI ethics.}
If self-referential structures emerge from multi-agent task pressure, then sufficiently complex multi-agent AI systems may develop self-monitoring and self-referential communication without explicit design.
Understanding the conditions under which this arises, and the architectural features it requires, is important for predicting system behavior in multi-agent deployment and for the ethical governance of systems that may develop consciousness-relevant properties.

\paragraph{Applications.}
The structural preconditions we identify have potential applications beyond consciousness science.
\citet{silver2025era} envision agents that learn from grounded experience rather than human-curated data. Self-referential structures may be a prerequisite for the kind of self-design and self-correction that such agents would require.
In a related vein, \citet{hinton2026maternal} has argued that AI systems need ``maternal instincts,'' a genuine caring about human welfare, not merely hardwired constraints.
Whether such capacities require something like self-referential structures is an open question that our framework helps to operationalize.
More speculatively, understanding the functional architecture of self-reference may eventually inform the design of brain--computer interfaces or neural prosthetics that must bridge artificial and biological self-models, though such applications remain far beyond the scope of the present work.

\paragraph{Bridging prior-minimal and prior-rich approaches.}
The present work operates at one end of a methodological spectrum: agents start from minimal prior, and the resulting causal clarity is precisely what makes the approach scientifically valuable.
At the other end, language models trained on human text possess rich world knowledge and sophisticated linguistic capabilities, but because their training corpora are saturated with self-referential and mental-state language, it is difficult to determine whether any given internal structure was learned from environmental contingencies or inherited from the statistical regularities of the corpus.
Prior-rich approaches provide scale, but cannot by themselves identify whether self-monitoring formed independently of inherited human descriptions; the prior-minimal setting isolates this formation question and offers cleaner causal attribution. A future bridge would require demonstrating equivalence in causal organization rather than similarity in report space.

An important future direction is to bridge this gap: developing principled methods for integrating grounded emergent structures with the knowledge encoded in pre-trained models, so that the strengths of each can be leveraged without sacrificing the other.
Naive approaches, such as directly optimizing alignment between emergent and pre-trained representations (e.g., embedding cosine similarity as a training objective), risk reintroducing prior leakage through the loss function, undermining the causal attributability that makes the grounded approach distinctive.

One possible direction, which we have not yet explored, is staged or modular architectures in which grounded agents develop foundational structures under task pressure, while pre-trained models contribute to environment design, curriculum shaping, or post-hoc interpretation, with the agent's internal representations remaining causally insulated from the pre-trained prior.
Whether such a separation can be maintained in practice is an open question.

If independently developed grounded structures and LLM internal representations (such as the emotion concepts of \citet{sofroniew2026emotion} or the self-referential features of \citet{berg2025llm}) can be shown to converge structurally, this would provide grounding evidence for those LLM representations that the prior-rich setting alone cannot deliver.
Developing such a bridge is, in our view, among the most important open problems for the field.


\end{document}